\documentclass[a4paper,10pt]{article}
\usepackage[utf8x]{inputenc}

\usepackage{amssymb}
\usepackage{amsmath}
\usepackage{epsfig}
\usepackage{graphicx}
\usepackage{subfigure}
\usepackage[T1]{fontenc}
\usepackage{courier}

\title{Penalized K-Nearest-Neighbor-Graph Based Metrics for Clustering}
\author{Ariel E. Bay\'a and Pablo M. Granitto\\
CIFASIS\\ French Argentine International Center for Information and Systems Sciences\\ UPCAM (France) / UNR--CONICET (Argentina),\\
Bv. 27 de Febrero 210 Bis, 2000 Rosario, Argentina\\
{baya, granitto}@cifasis-conicet.gov.ar
}
\date{}
\begin{document}

\maketitle

\begin{abstract}
A difficult problem in clustering is how to handle data with a manifold structure, i.e. data that is not shaped in the form of compact clouds of points, forming arbitrary shapes or paths embedded in a high-dimensional space. In this work we introduce the Penalized k-Nearest-Neighbor-Graph (PKNNG) based metric, a new tool for evaluating distances in such cases. The new metric can be used in combination with most clustering algorithms. The PKNNG metric is based on a two-step procedure: first it constructs the k-Nearest-Neighbor-Graph of the dataset of interest using a low k-value and then it adds edges with an exponentially penalized weight for connecting the sub-graphs produced by the first step. We discuss several possible schemes for connecting the different sub-graphs. We use three artificial datasets in four different embedding situations to evaluate the behavior of the new metric, including a comparison among different clustering methods. We also evaluate the new metric in a real world application, clustering the MNIST digits dataset. In all cases the PKNNG metric shows promising clustering results.
\end{abstract}

\section{Introduction}
Clustering is a key component of pattern analysis methods, aiming at finding hidden structures in a set of data. It is a very active field of research\cite{tpami1,tpami2,tpami3}, with applications that cover diverse problems, from grouping star systems based on astronomical observations to selecting genes and proteins that have similar functionality or characterizing customer groups based on purchasing patterns.

The problem of finding clusters in a dataset can be divided into three stages: i) measuring the similarity of the objects under analysis, ii) grouping the objects according to these similarities, and iii) evaluating the "goodness" of the clustering solution. The last stage has received little attention until recent years, when a growing interest in the problem can be noticed \cite{benhur,gap_stats}. The second stage (finding clusters efficiently given a set of similarities between objects) has been widely studied in the literature \cite{review1} and several clustering algorithms have been introduced. They are usually divided into hierarchical and partitional methods \cite{review2}.
Hierarchical clustering (HC) algorithms find successive clusters using previously defined ones, in an agglomerative ("bottom-up") or divisive ("top-down") way. The result of this process is a binary tree, called a dendrogram\cite{king}. The number of clusters obtained depends on the level at which the user cuts the dendrogram. Partitional algorithms, on the other hand, pre--determine a fixed number of clusters. One of the most widely used approaches is the K-Means \cite{kmeans} algorithm (or its improved version PAM\cite{pam}) that, starting from $k$ random clusters, searches iteratively for a locally optimal solution of the clustering problem. Recently, Frey and Dueck\cite{frey} proposed the innovative and computationally efficient Affinity Propagation (AP) algorithm. According to this method, each data point is viewed as a node in a network. Nodes exchange messages until a set of cluster-centers emerges as a solution. The algorithm shares characteristics with both hierarchical and partitioning methods. 

Despite the success that these methods have shown in several artificial and experimental datasets, they usually fail to handle manifold-structured data, i.e. data that form low-dimensional, arbitrary shapes or paths through a high-dimensional space. There are algorithms that can in principle handle with this situation, like Single Linkage Hierarchical clustering \cite{sneath} or Chamaleon \cite{chamaleon}. Most of these methods are based on graph theory. Unfortunately, they are limited in the range of problems they can solve and usually need thousands of data samples to work properly. Instead of developing new algorithms appropriate for these problems, we propose here to obtain better solutions going back to stage one of clustering, i.e., by finding better ways to measure similarities between datapoints. In some sense, our view shares the spirit of kernel methods \cite{kernelmethods}, looking for solutions to new problems by using appropriate new metrics together with well-known pattern analysis algorithms.

In the last years, several methods for characterizing the non-linear manifold where a dataset may lie were developed, like ISOMAP\cite{isomap}, Locally Linear Embedding\cite{lle} or Laplacian Eigenmaps\cite{laplacian}. Basically, they all look for local neighborhood relations that can be used to produce low dimensional projections of the data at hand. Ham et al.\cite{ham04} showed that they can also be interpreted as particular kernels.

In this paper we discuss new strategies to evaluate similarities in manifold spaces that easily extend the application of any clustering algorithm to these cases. Following ISOMAP, we first create the k-nearest-neighbor-graph (knn-graph) of the data, using a low k-value. If the graph is disconnected, which is expected in clustering problems, we add a number of edges (following different strategies that will be discussed later) in order to create a connected graph. The key point of our method is that the added edges have an exponentially penalized length. We then apply an appropriate algorithm to measure inter-point distances along the connected graph and use these measures as (dis)similarities. We call the method the PKNNG metric (for Penalized K-Nearest-Neighbor-Graph based metric).

In Section II we introduce the method and discuss the different schemes we use to connect the subgraphs. Then, in Section III we introduce three two-dimensional artificial problems, which we embed in higher-dimensional spaces. We use these problems to evaluate the new metrics on different aspects and also show results on a digit recognition problem. Finally, in Section IV we draw some conclusions and discuss future lines of research.

\section{The PKNNG metric}

The evaluation of similarities with the PKNNG metric is a two-step process. First we search the original dataset space for locally dense (connected) structures using the knn-graph. In the ideal case the process will end with a connected subgraph corresponding to each cluster but in the real case, when working with finite noisy samples, there are usually too many separated structures, typically more than one for each real cluster. In the second step we add penalized edges to the graph in order to fully connect it, and use an appropriate algorithm to measure distances in the (now) connected graph.

\begin{figure}
\centering
\subfigure{
\includegraphics[trim = 0mm 15mm 0mm 15mm, width=2.4in]{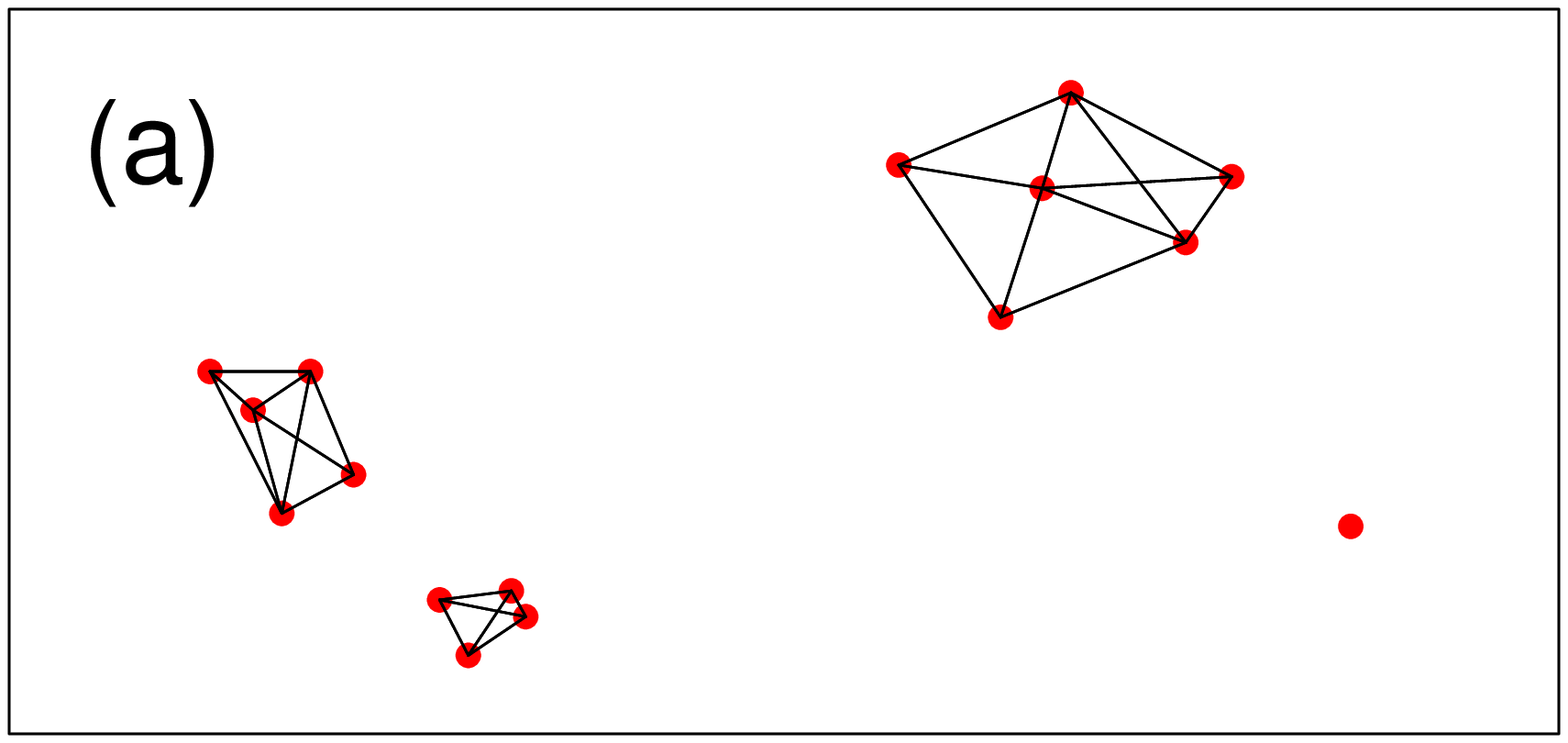}
} \\
\vspace{-.2in}
\subfigure{
\includegraphics[trim = 0mm 15mm 0mm 15mm, width=2.4in]{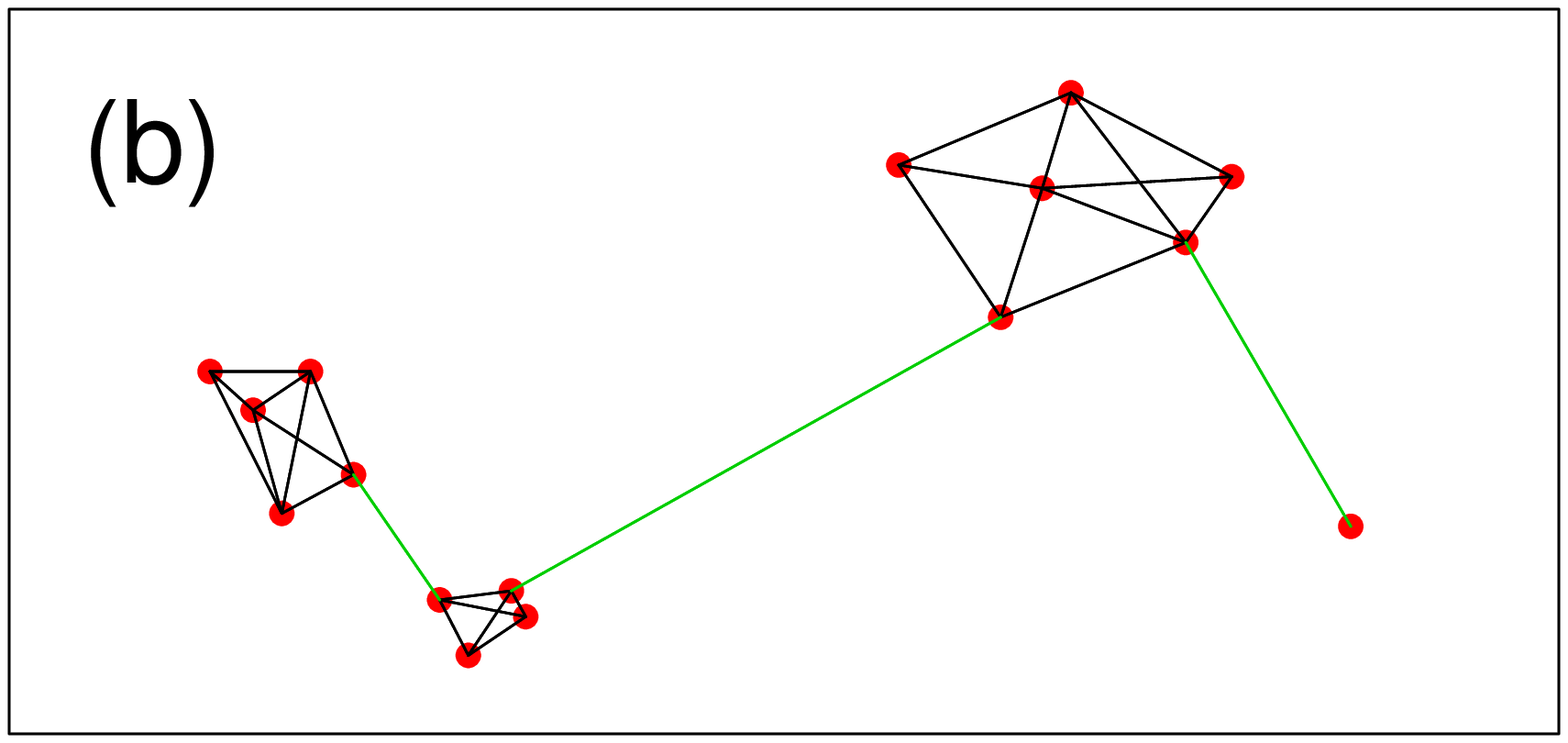}
} \\
\vspace{-.2in}
\subfigure{
\includegraphics[trim = 0mm 15mm 0mm 15mm, width=2.4in]{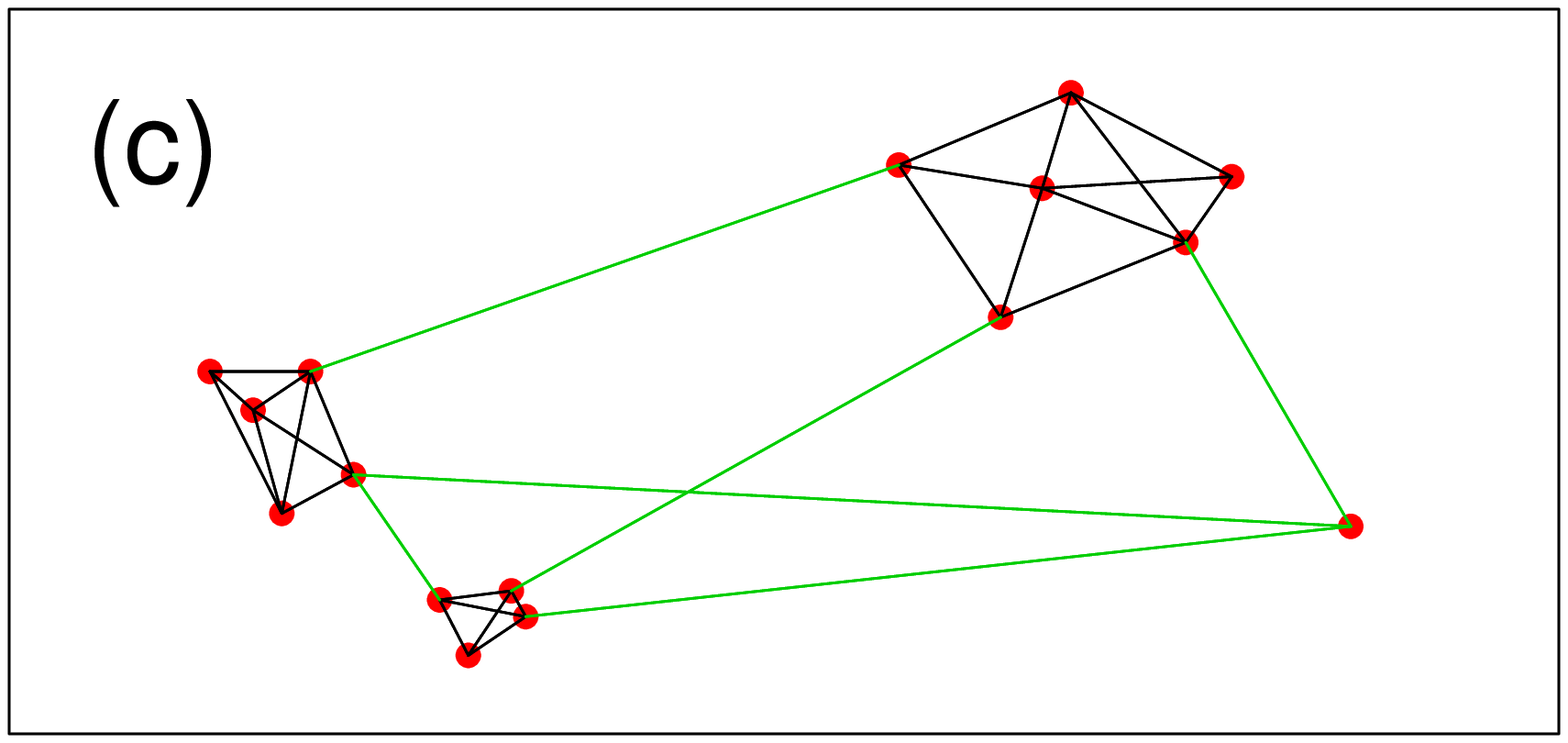}
} \\
\vspace{-.2in}
\subfigure{
\includegraphics[trim = 0mm 15mm 0mm 15mm, width=2.4in]{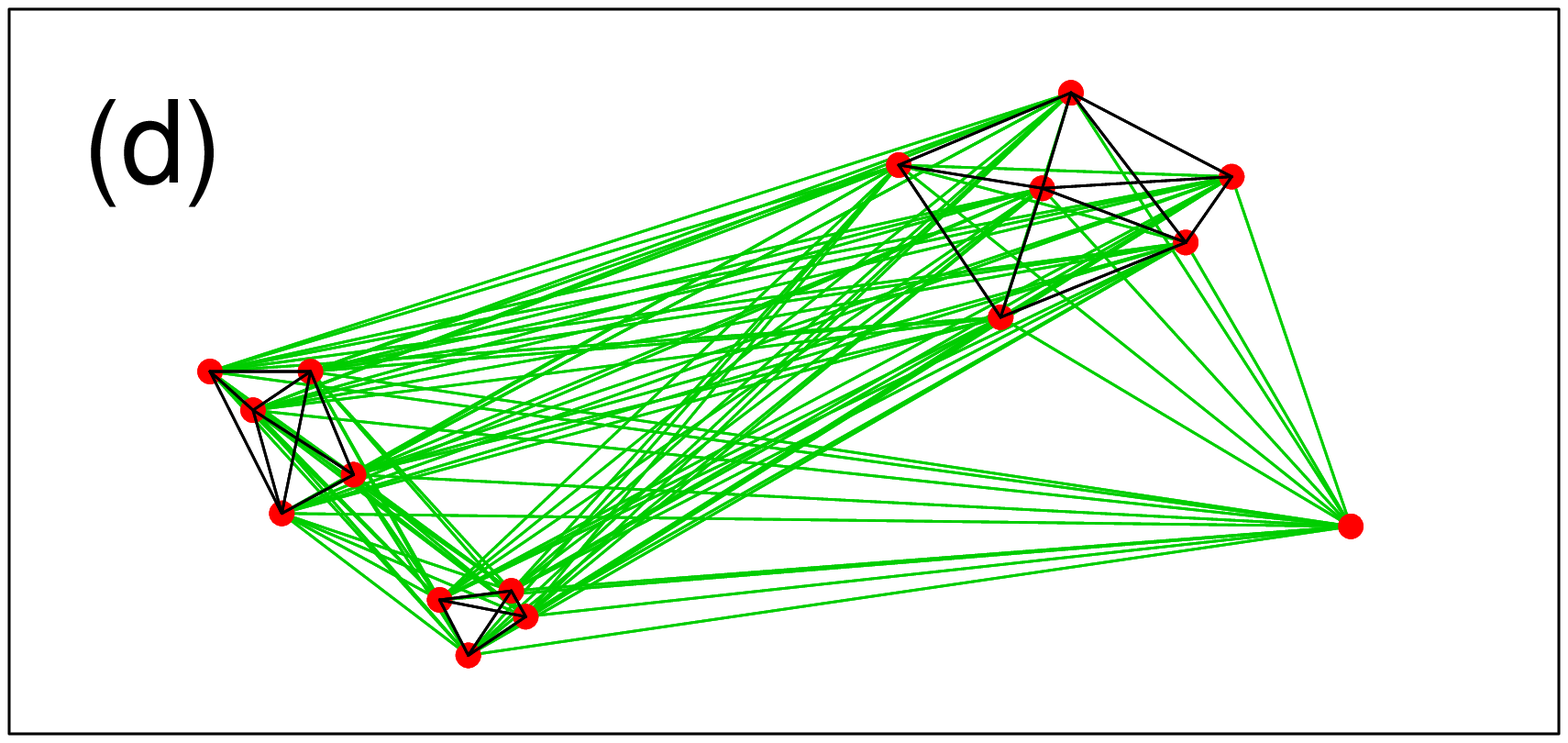}
} \\
\vspace{-.2in}
\subfigure{
\includegraphics[trim = 0mm 15mm 0mm 15mm, width=2.4in]{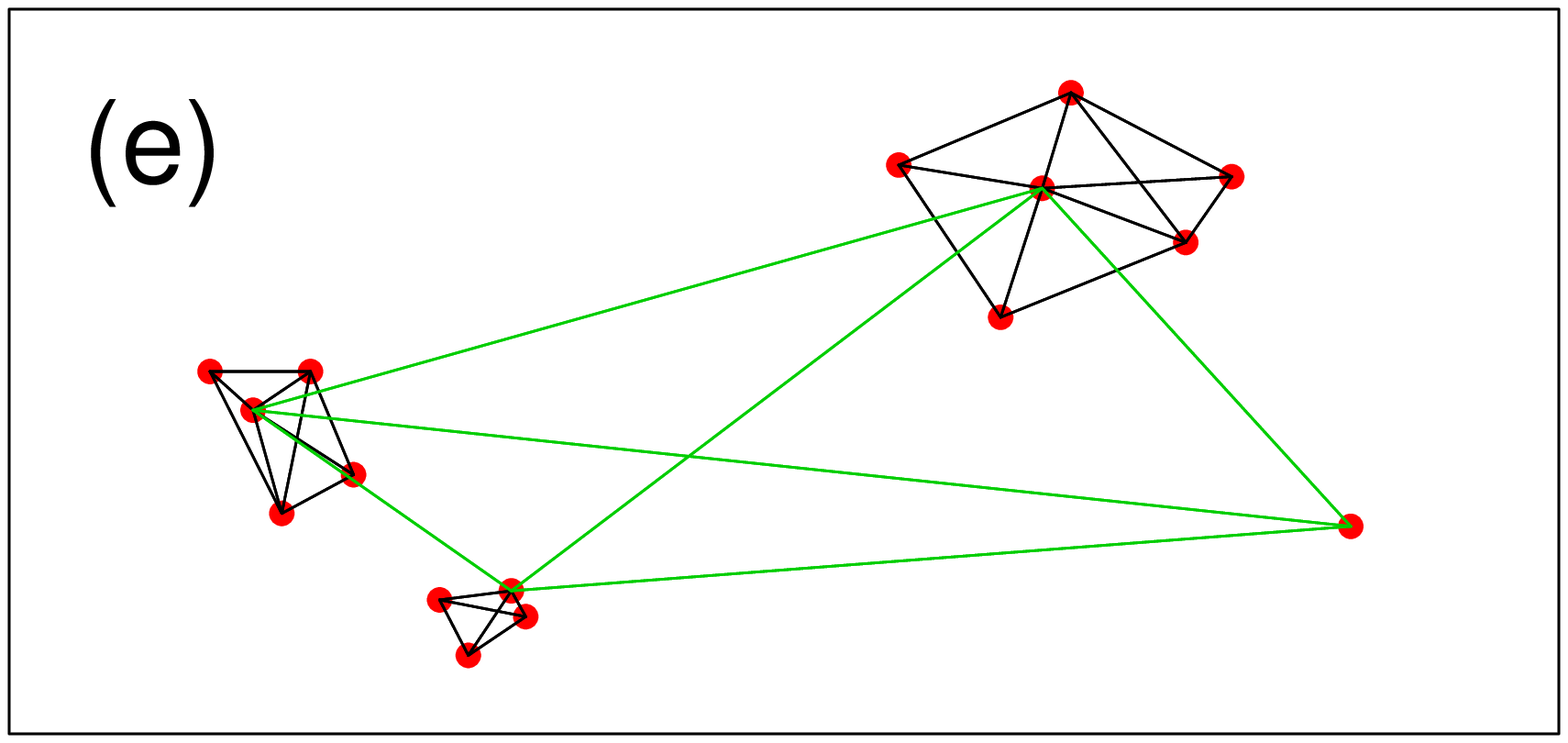}
} \\
\caption{A toy dataset illustrating the different connection schemes evaluated in this work. a) The original data with the knn-graph. Note the disconnected outlier at the bottom-right. b) The MinSpan scheme. c) The AllSubGraphs scheme. Note that all MinSpan added edges are included. d) The AllEdges scheme. e) The Medoids scheme. See text for details on each method.}
\label{fig1}
\end{figure}

\subsection{First step: knn-graphs}
Among the several algorithms for discovering low dimensional manifolds recently introduced, ISOMAP has strong theoretical properties and is also easy to understand. We follow the main ISOMAP idea to search for locally connected structures. As explained by Tenenbaum, de Silva and Langford\cite{isomap}, in a curved manifold the geodesic distance between neighboring points can be approximated by the Euclidean input space distance. For distant points, in contrast, geodesic distances are better approximated as a path of short segments connecting neighboring points. To this end, we construct the knn-graph of the data, i.e. the graph with one vertex per observed example, arcs between each vertex and its $k$ nearest neighbors, and with weights equal to the Euclidean distance between them. As we look for dense subgraphs, at the end of the process we eliminate all outliers from the graphs. We consider that an arc is an outlier if it is not reciprocal (i.e. one of the vertex is not a k-nn of the other) and the length of the arc is an outlier of its distribution (i.e. if it is larger than the 3rd quartile plus $1.5$ times the inter-quartile distance). In Figure \ref{fig1} we show a toy example of this process. Panel (a) shows the original data with the corresponding knn-graph. In general, the use of a low number of neighbors (3 to 7 in all our cases, as in \cite{isomap}) produces graphs that can follow the curved structure of any data without adding "shortcuts" between geodesically distant points\cite{isomap}.

\begin{figure}
\centering
\includegraphics[trim = 0mm 10mm 0mm 10mm, width=2.5in]{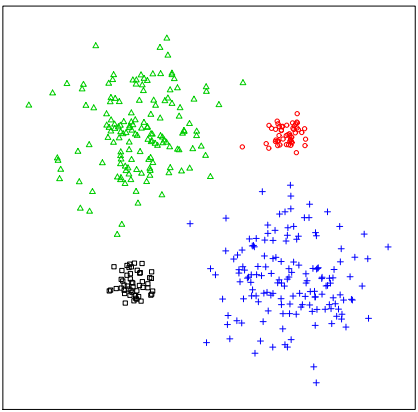}
\caption{Clustering result on an artificial dataset with 4 Gaussians and 2 different densities, using the PAM algorithm with the PKNNG metric (MinSpan scheme). The result is equivalent to using the standard Euclidean metric.}
\label{gaussians}
\end{figure}

\begin{figure*}
\begin{center}

\subfigure{
\includegraphics[width=1.75in]{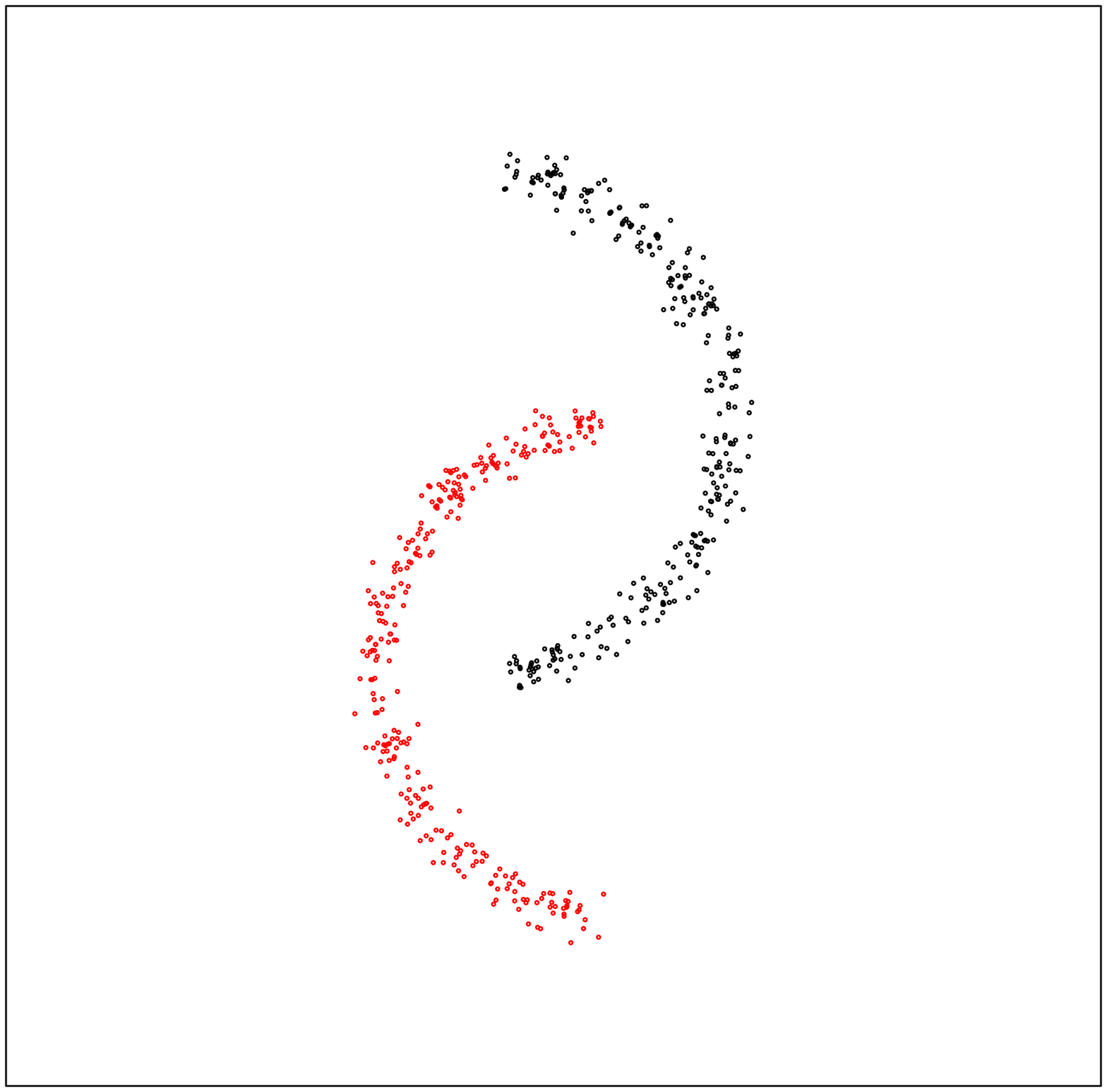}
}
\hspace{-0.25in}
\subfigure{
\includegraphics[width=1.75in]{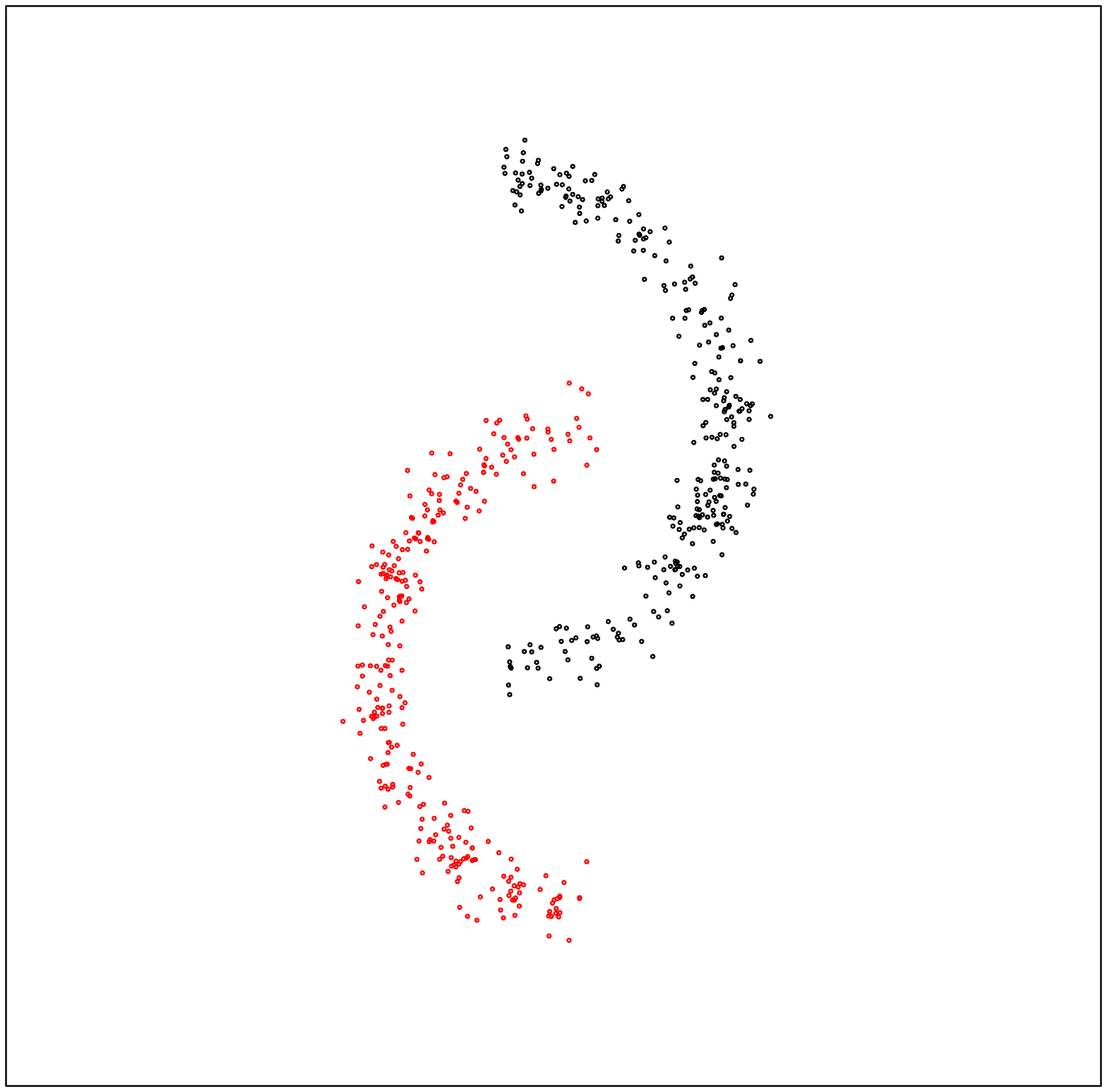}
}
\hspace{-0.25in}
\subfigure{
\includegraphics[width=1.75in]{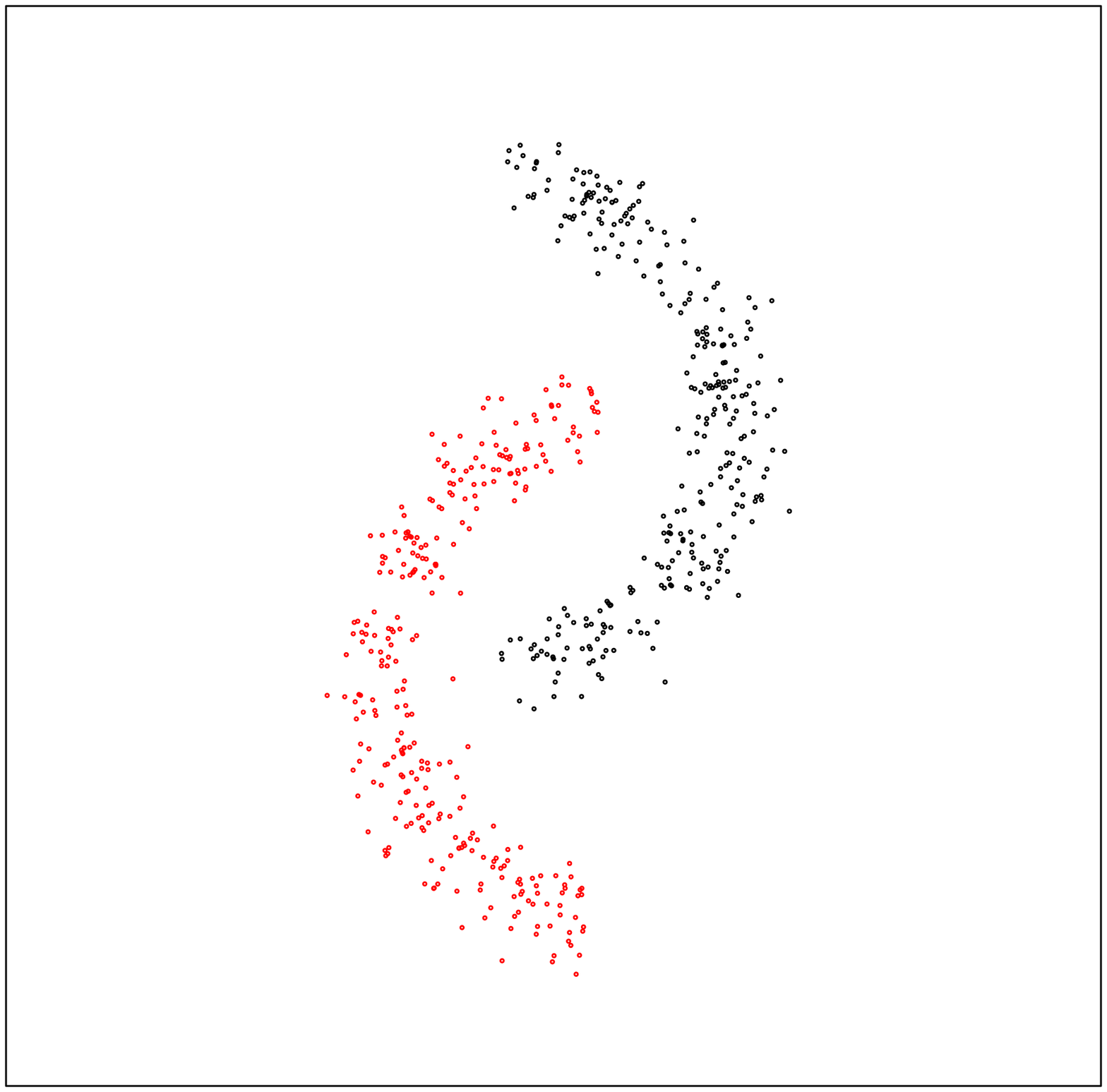}
} \\

\vspace{-.4in}

\subfigure{
\includegraphics[width=1.75in]{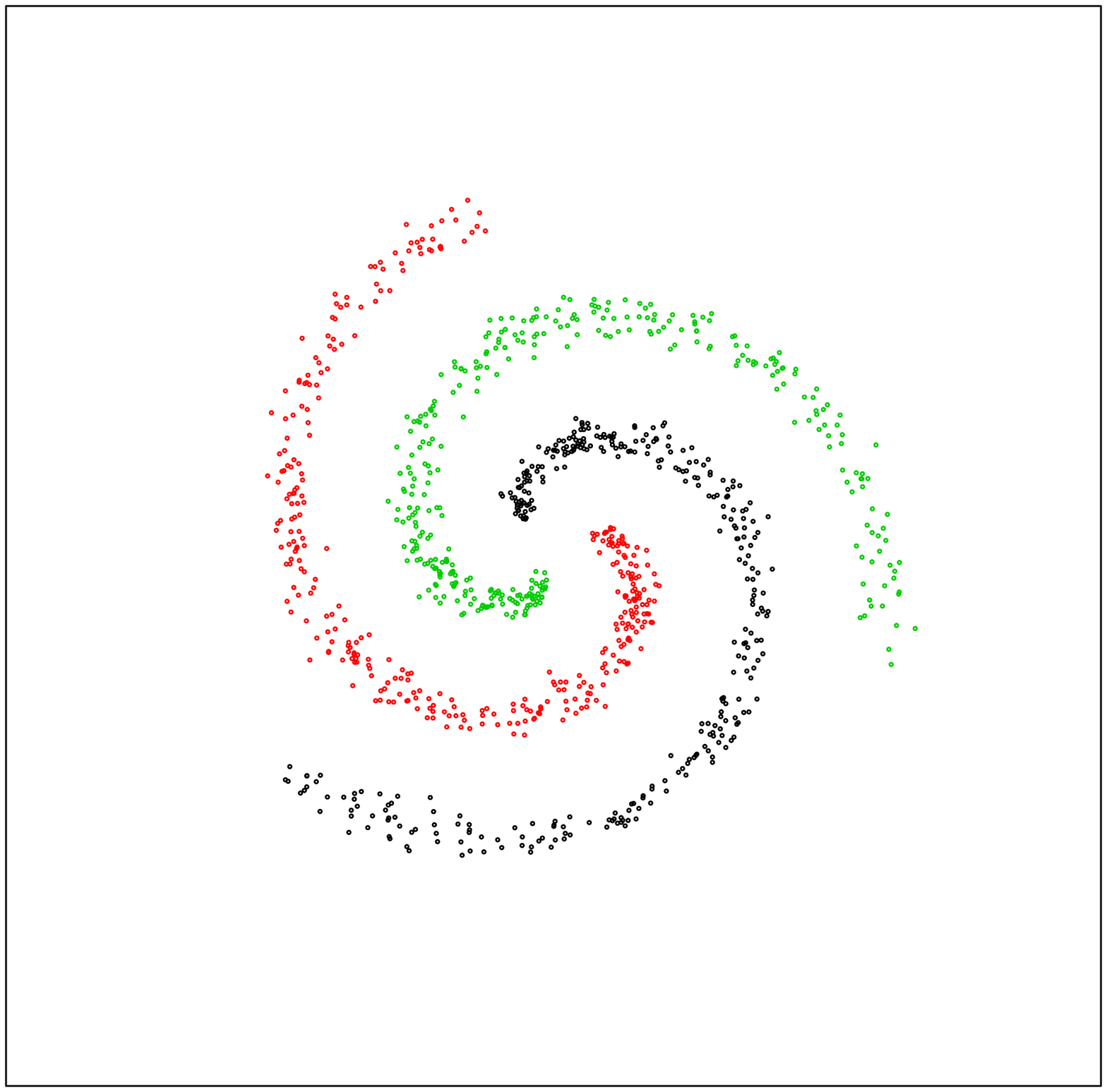}
}
\hspace{-0.25in}
\subfigure{
\includegraphics[width=1.75in]{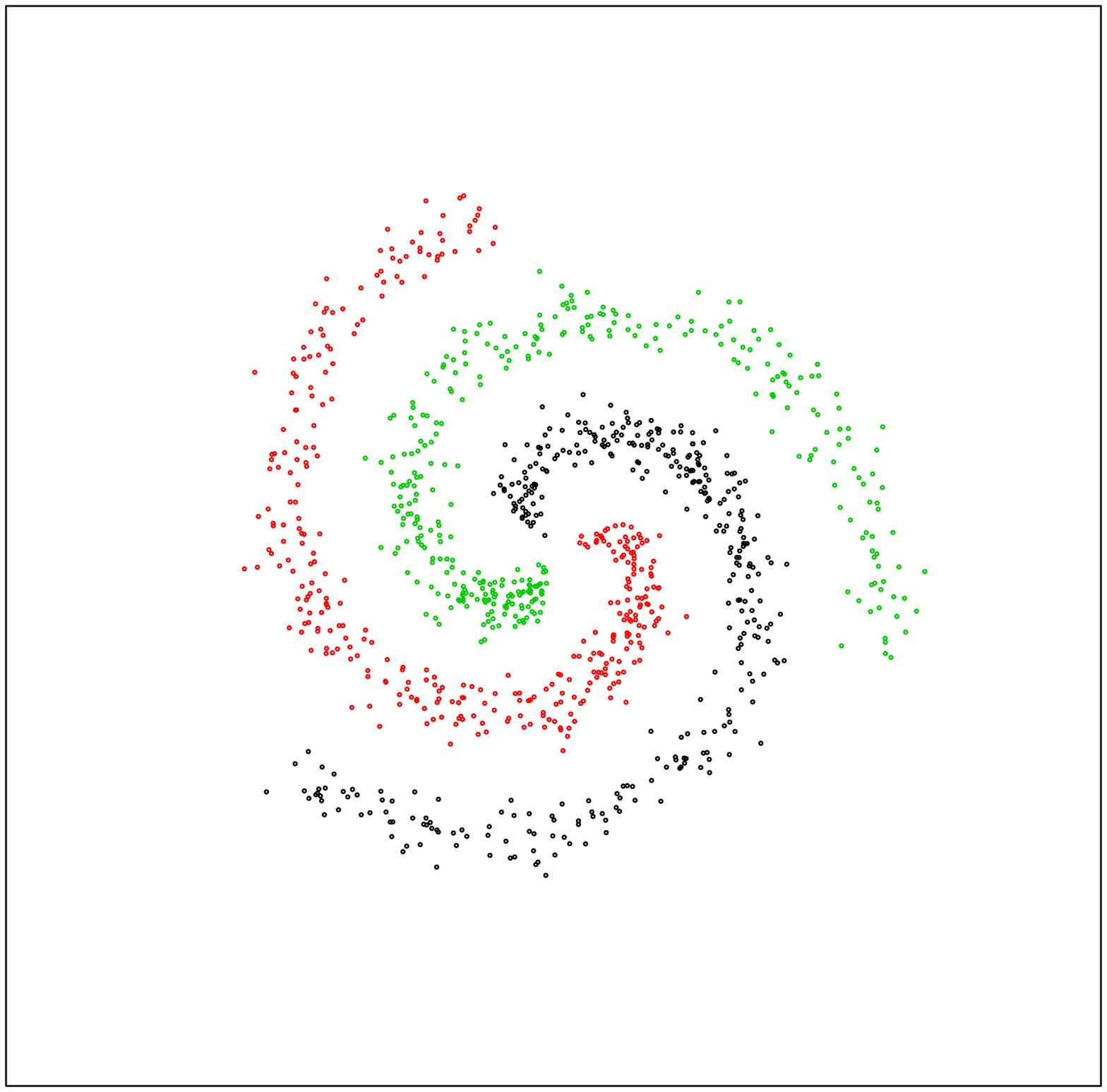}
}
\hspace{-0.25in}
\subfigure{
\includegraphics[width=1.75in]{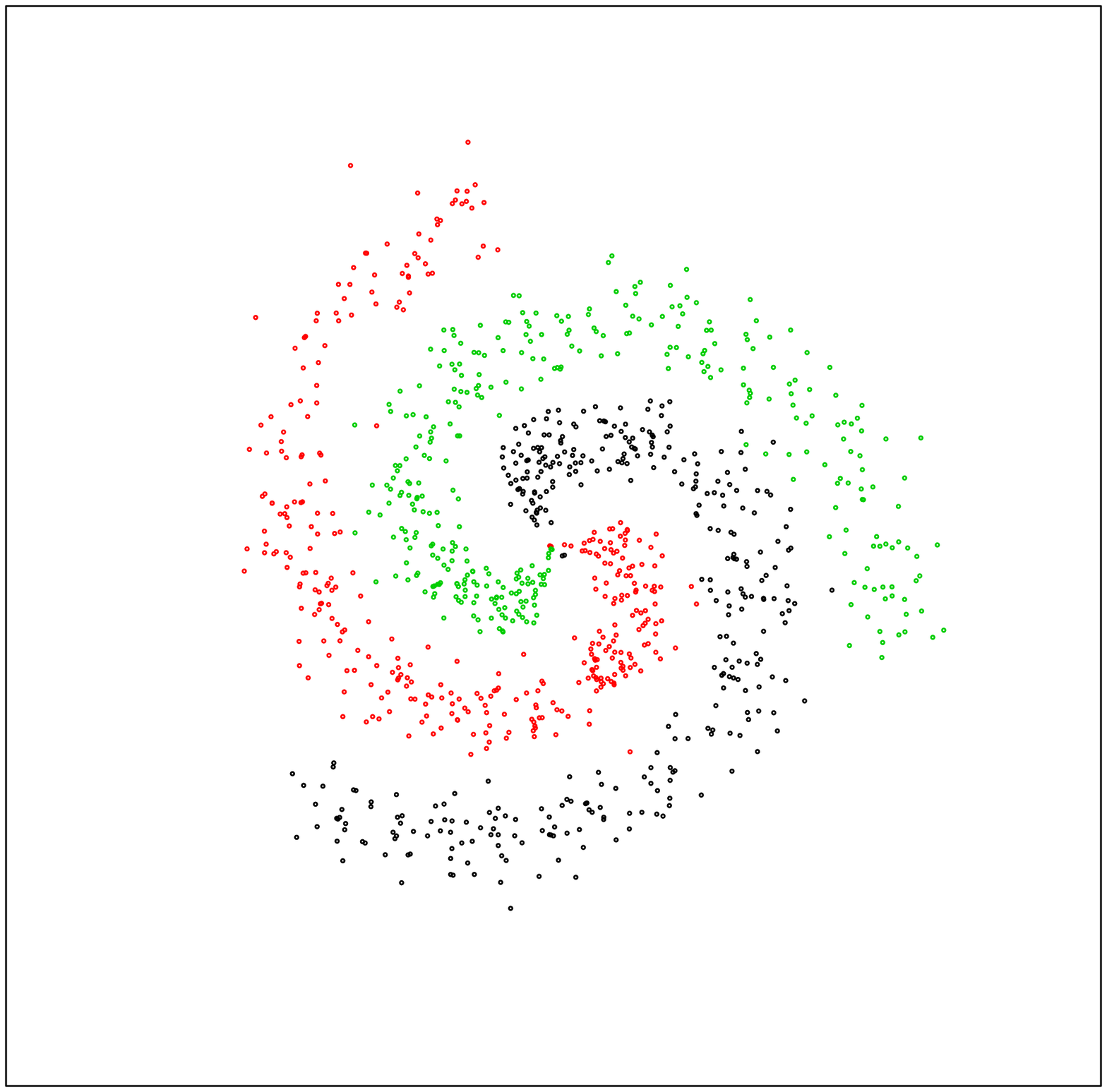}
} \\

\vspace{-.4in}

\subfigure{
\includegraphics[width=1.75in]{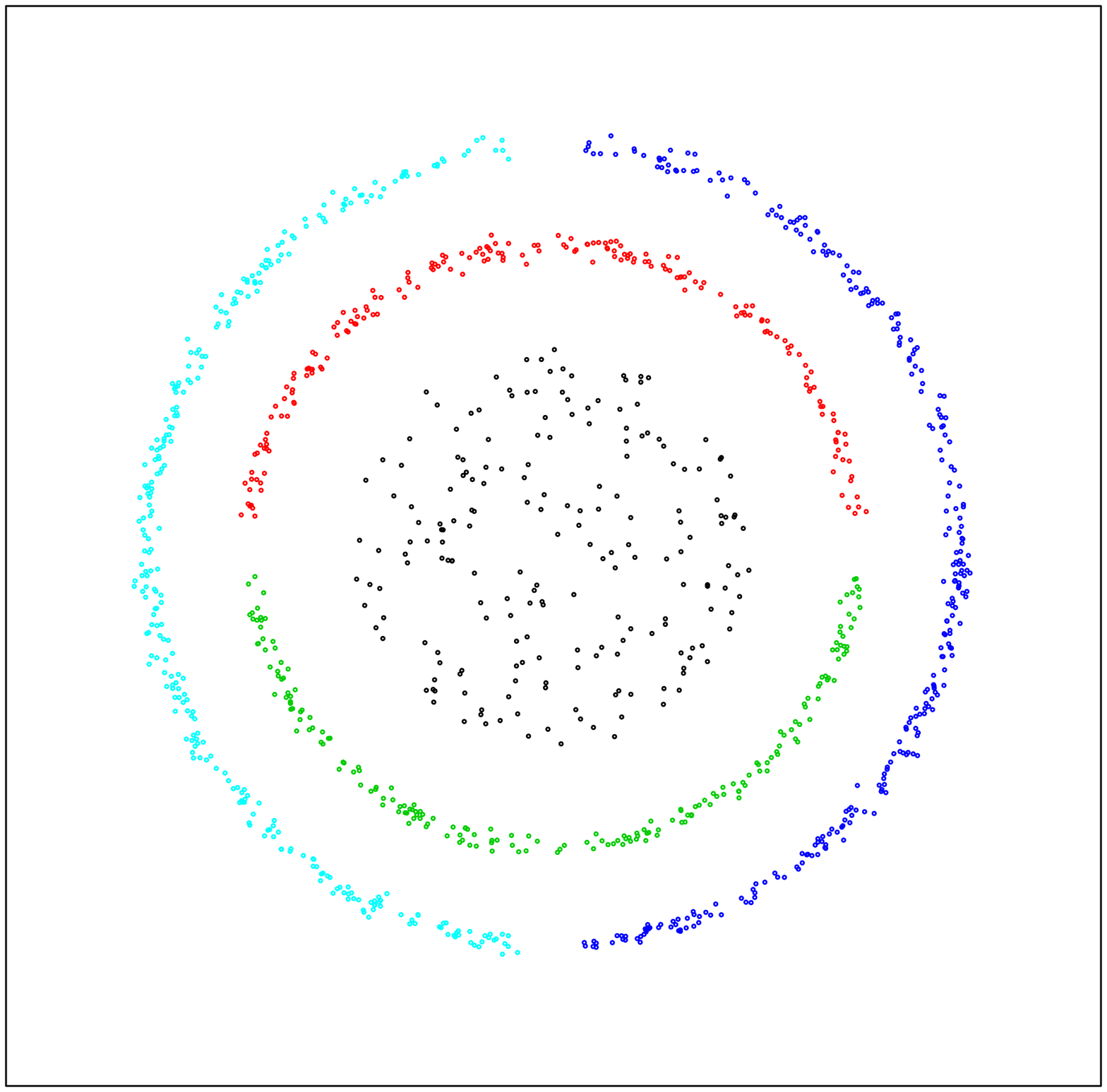}
}
\hspace{-0.25in}
\subfigure{
\includegraphics[width=1.75in]{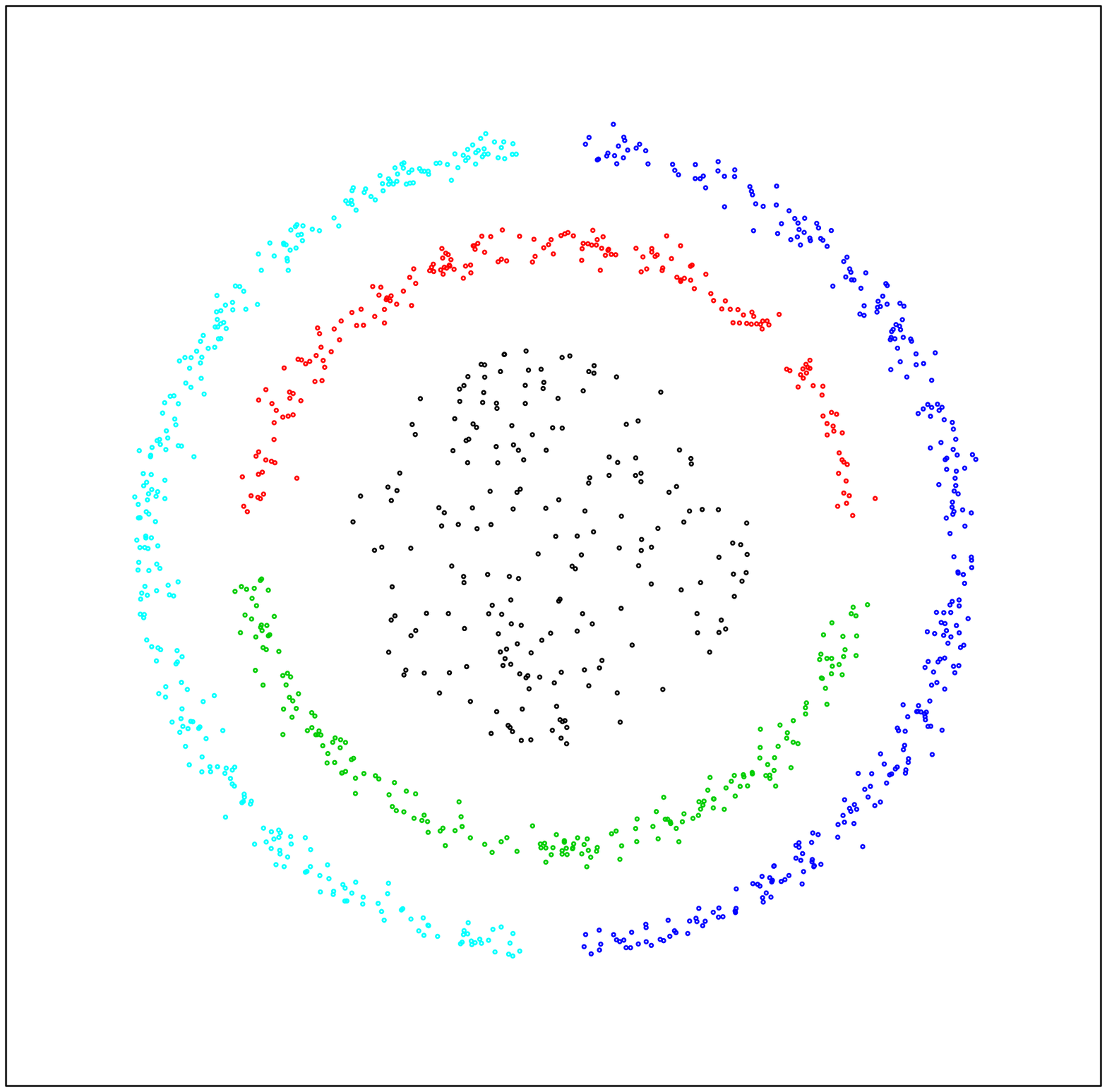}
}
\hspace{-0.25in}
\subfigure{
\includegraphics[width=1.75in]{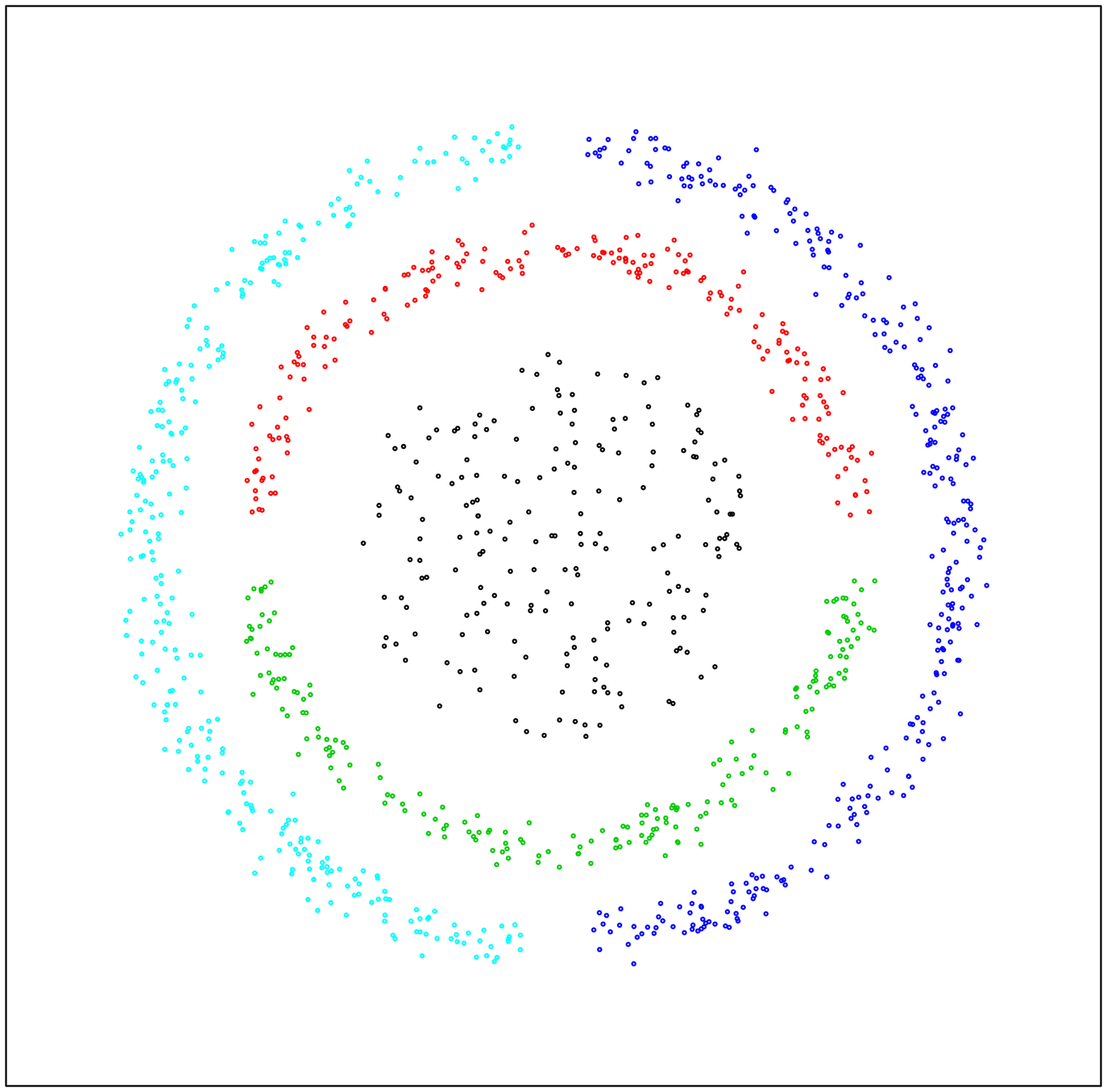}
} \\

\vspace{-.3in}

\end{center}

\caption{Samples of the datasets used in this work. From top to bottom, rows correspond to the Two-arcs (2 clusters), Three-spirals (3 clusters) and Three-rings (5 clusters) datasets. From left to right, columns correspond to low, medium and high noise levels, respectively.}
\label{dataexamples}
\end{figure*}

\subsection{Second step: connecting subgraphs}

As stated previously, when working with real data the first step will usually generate separated structures, typically more than one for each real cluster. In the ideal case, each sub-graph should correspond to a unique cluster and we could leave them disconnected and interpret those distances as equal to infinite. In the real case, however, some sub-graphs correspond to the same cluster. Therefore, we need to evaluate also geodesic distances between points lying in different sub-graphs. With that purpose, we add to the graph a number of edges in order to fully connect it. The key point of the method is that we penalize the weight of these new connecting edges with an exponential factor of the form:
\begin{equation}
w=d \ \ e^{ d / {\mu}}, \label{eq1}
\end{equation}
where $w$ is the graph weight corresponding to the added edge, $d$ is the Euclidean distance between the points being connected by that edge and $\mu$ is the mean edge weight in the original graph. Using this metric we can connect sub-graphs corresponding to the same cluster with a relatively small cost, because connections in the same order of magnitude of $\mu$ will get a low penalization. On the other hand, edges connecting distant sub-graphs will be strongly penalized\footnote{The factor $w=d \ \ e^{ d / {\mu} -1 }$ seems more natural, penalizing only external connections bigger than $\mu$. We choose our formulation in order to penalize in practice all added edges.}.

Which edges to add in order to connect the graph is an interesting problem by itself. We evaluate in this work four different schemes that cover most of the simple possibilities:

\subsubsection*{MinSpan}

In this first scheme we add to the graph the minimum number of edges, each of them of minimum length (the minimum spanning set), which fully connect the graph. On panel (b) of Figure \ref{fig1} we show the result of this connection strategy on our toy example.

\subsubsection*{AllSubGraphs}

Here we connect each sub-graph to all other sub-graphs using minimum length edges. Figure \ref{fig1}, panel (c), shows this strategy on the toy example.

\subsubsection*{AllEdges}

In this simple scheme we add to the graph all remaining edges (of course, with a penalized weight). On panel (d) of Figure \ref{fig1} we show the corresponding graph.

\subsubsection*{Medoids}

In this last strategy we first find the medoid of each sub-graph, and then add edges connecting each medoid to all remaining medoids. Figure \ref{fig1}, panel (e), shows this strategy applied to our toy example.

The idea behind MinSpan is to add the shortest available edges trying to follow as much as possible the structure of the manifold. MinSpan basically produces one-dimensional structures. We showed in a previous work \cite{asai07} that this effect can introduce some instability for distant points, but (as we will confirm later) it does not affect the performance of clustering. AllSubGraphs is an extension of MinSpan and AllEdges can be viewed as an extension of AllSubGraphs. At each step of this chain we add more edges, which increases the connectivity of the graph, reducing the potential instability of MinSpan. On the other hand, by adding edges we increase the probability of introducing "shortcuts" in the manifold, which can reduce the overall performance of the method. The Medoids scheme is basically introduced to have a different scheme to compare with. As it does not include the minimum spanning set, it is always forced to use other paths in the graph and, consequently, it is not expected to be abble to follow a curved manifold. Neighboring points belonging to different sub-graphs can be completely separated by this connection scheme.

Once we have a connected graph, we can compute geodesic distances between all points as minimum-length paths in the graph using computationally efficient algorithms like Floyd or Dijkstra\cite{dijkstra}.

\section{Evaluation}

\subsection{Compact cloud data}

One of the advantages of the PKNNG metric is that it can produce accurate results both for curved/elongated manifolds and for typical compact cloud data. On this last, more typical kind of data, the PKNNG metric produces similar results to the Euclidean metric. As an example, Figure \ref{gaussians} shows the four clusters found by the PAM algorithm using the PKNNG metric (MinSpan scheme) on an artificial dataset created by sampling four Gaussians with different deviations\footnote{We evaluated several artificial datasets of this kind, always obtaining results similar to the Euclidean metric.}.

\subsection{Artificial Datasets}

To evaluate the proposed metric on more challenging problems including curved/elongated clusters, we use three artificial datasets with different characteristics:
\subsubsection*{Two-arcs} The first dataset corresponds to points uniformly sampled from two arcs of circumference (two clusters), with Gaussian noise of a fixed amplitude added to the radial direction.
\subsubsection*{Three-spirals} The second artificial dataset was generated by sampling uniformly from three equally separated spirals, with added Gaussian noise proportional to the radial distance. The result is a three clusters problem, each one of them having a non-uniform density, which can confuse some algorithms.
\subsubsection*{Three-rings} On this problem the central cluster corresponds to a uniform sampling of a circle, which is surrounded by two rings, also sampled uniformly but with constant Gaussian noise added to the radial component. We split both middle- and outer-rings in halves (adding a small gap), to create a more difficult five clusters problem. This third dataset has clusters with different (but uniform) densities.

\begin{figure}
\centering
\includegraphics[trim = 20mm 50mm 20mm 50mm, width=2.3in]{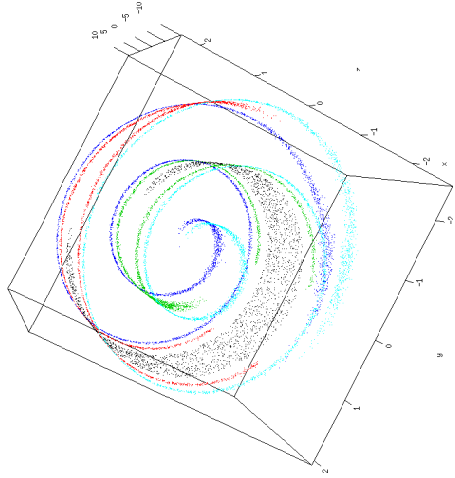}
\caption{Example of the embedding of our 2d artificial clustering problems in a swiss-roll.}
\label{swiss}
\end{figure}

All datasets were generated in a (2-dimensional) plane. For each one of them we used three different noise levels. An example of the clustering problems is shown in Figure \ref{dataexamples}. In order to evaluate the performance of the PKNNG metric at handling low-dimensional manifolds in high-dimensional spaces, we embedded our three datasets in 2d, 3d and 10d spaces, generating the following four settings:
\subsubsection*{2d} In this first case we kept the three datasets in the original 2d space.
\subsubsection*{3d} To start increasing the difficulty of the clustering problems, we coiled the original plane to form a swiss-roll, producing a non-linear embedding of the original clusters into a 3d space. In figure \ref{swiss} we show an example of the resulting problem.
\subsubsection*{3d--noise} In this third setting we added Gaussian noise to the previous 3d embedding in order to drift the points away from the surface of the swiss-roll.
\subsubsection*{10d--noise} As a last and more difficult setting, we took the 3d coiled data and added 7 extra dimensions to the problem. We then applied a random rotation in the 10d space, and finally added Gaussian noise in all 10 dimensions.

\subsection{Experimental settings}

We evaluate different aspects of our new metric using the three artificial datasets combined with the four different embeddings previously described. In all cases, after measuring similarities with a given metric, we used the three clustering algorithms described in the Introduction, namely PAM, HC and AP, to find an appropriate clustering of the data. The quality of the solutions was evaluated in terms of the clustering accuracy, i.e. the percentage of the dataset assigned to the right cluster. For each case under evaluation (dataset + noise level + embedding) we produced 100 different realizations of the dataset and computed the mean clustering accuracy.

\subsection{Connection schemes}

\begin{figure*}
\begin{center}


\subfigure{
\includegraphics[width=1.5in]{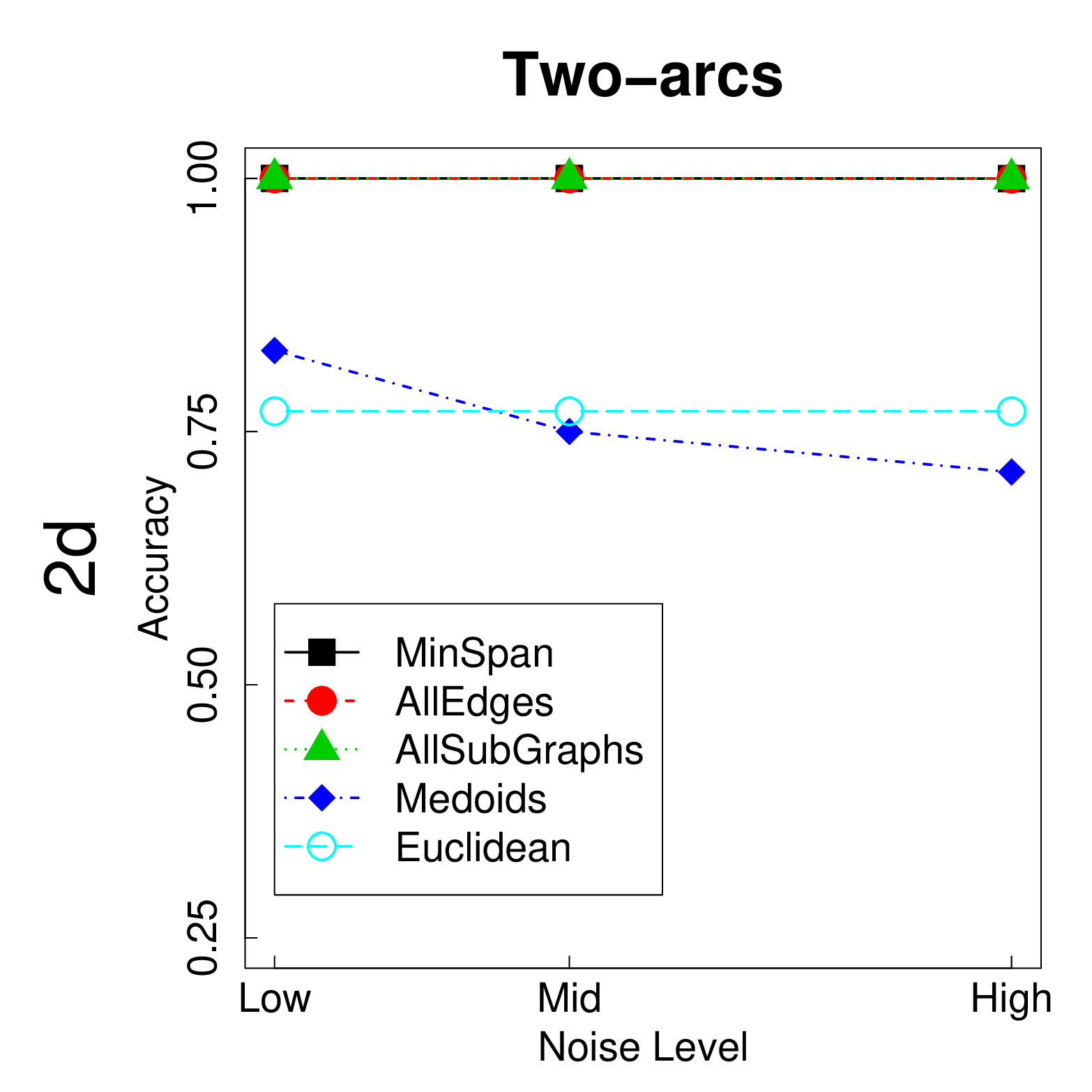}
}
\hspace{-0.3in}
\subfigure{
\includegraphics[width=1.5in]{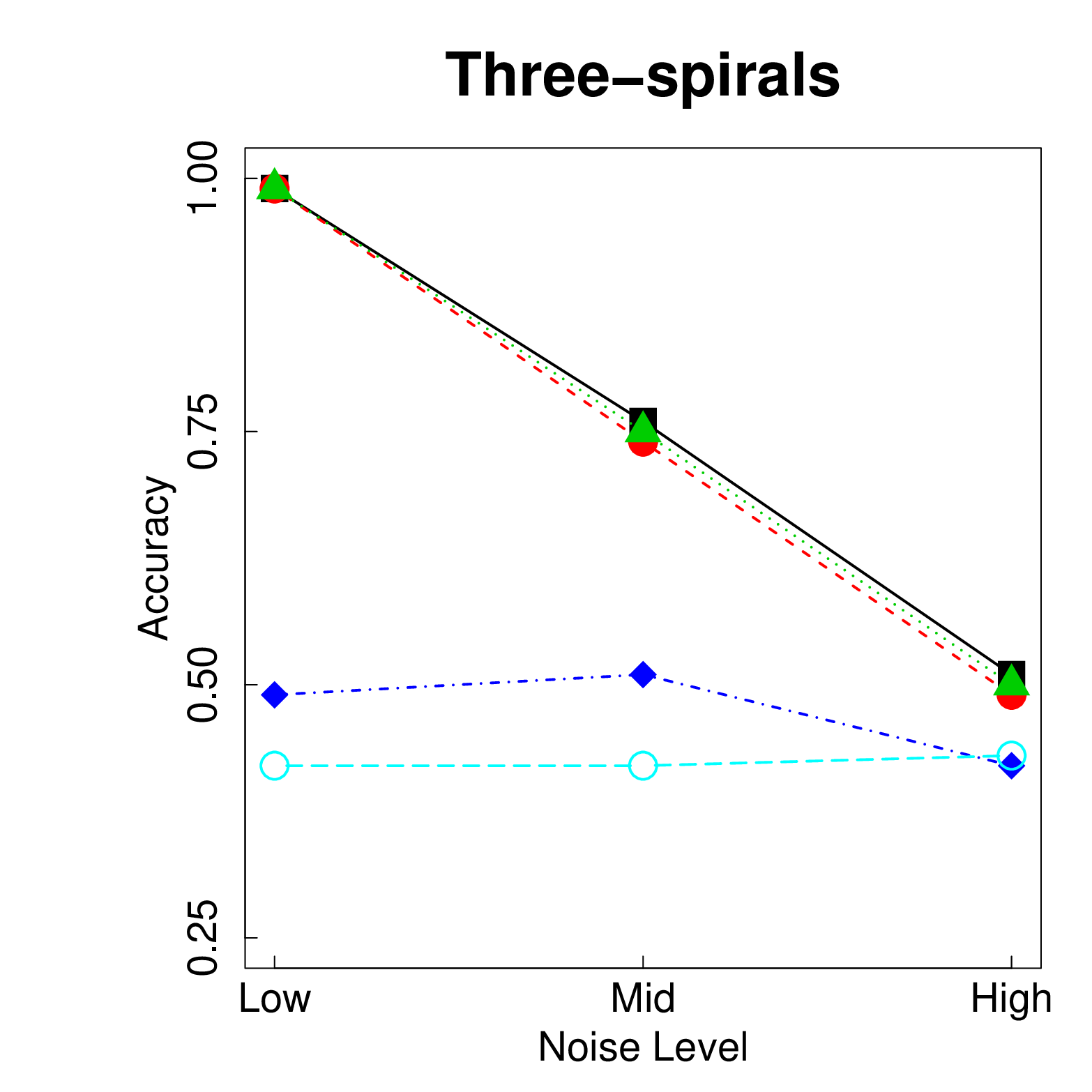}
}
\hspace{-0.3in}
\subfigure{
\includegraphics[width=1.5in]{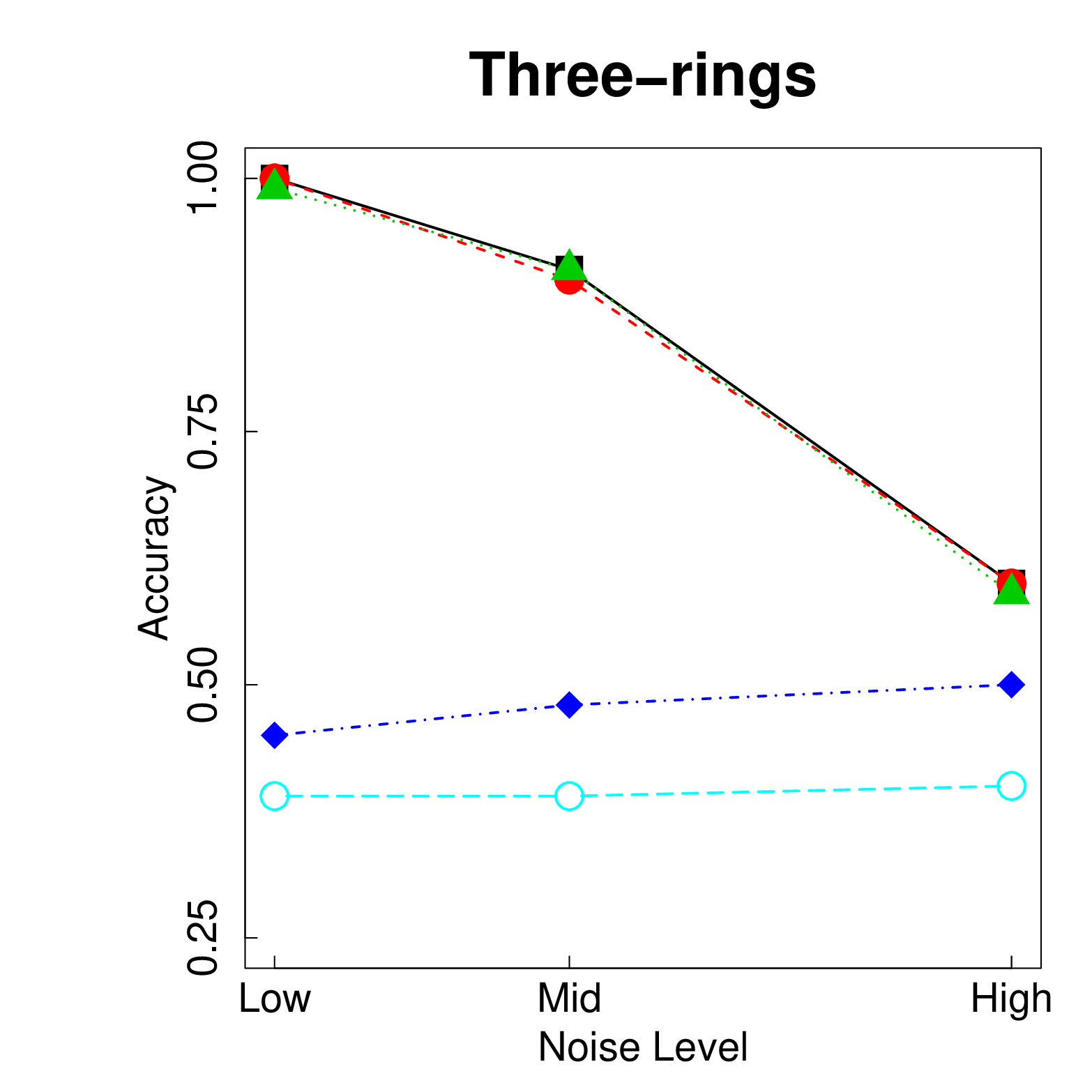}
} \\

\vspace{-.3in}

\subfigure{
\includegraphics[width=1.5in]{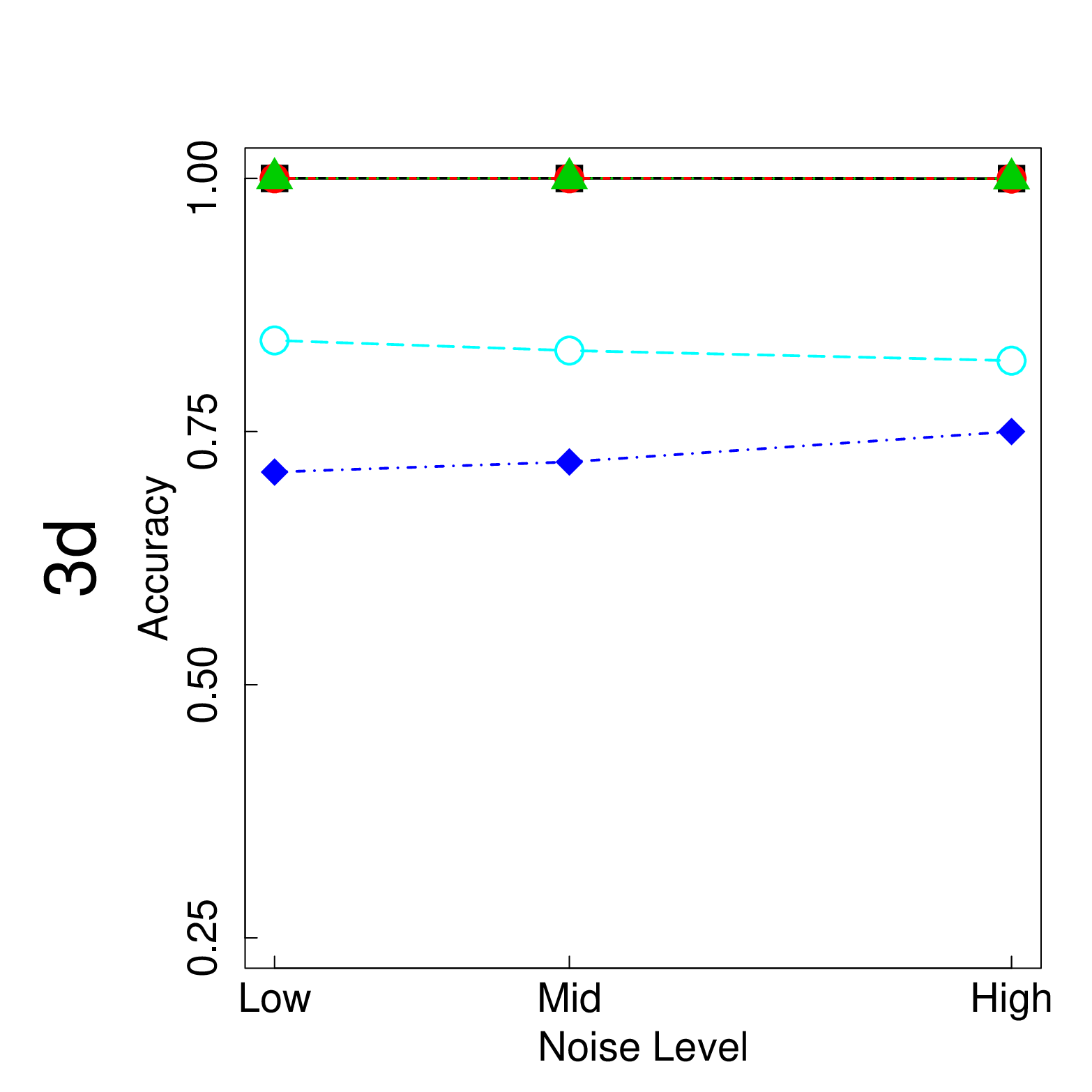}
}
\hspace{-0.3in}
\subfigure{
\includegraphics[width=1.5in]{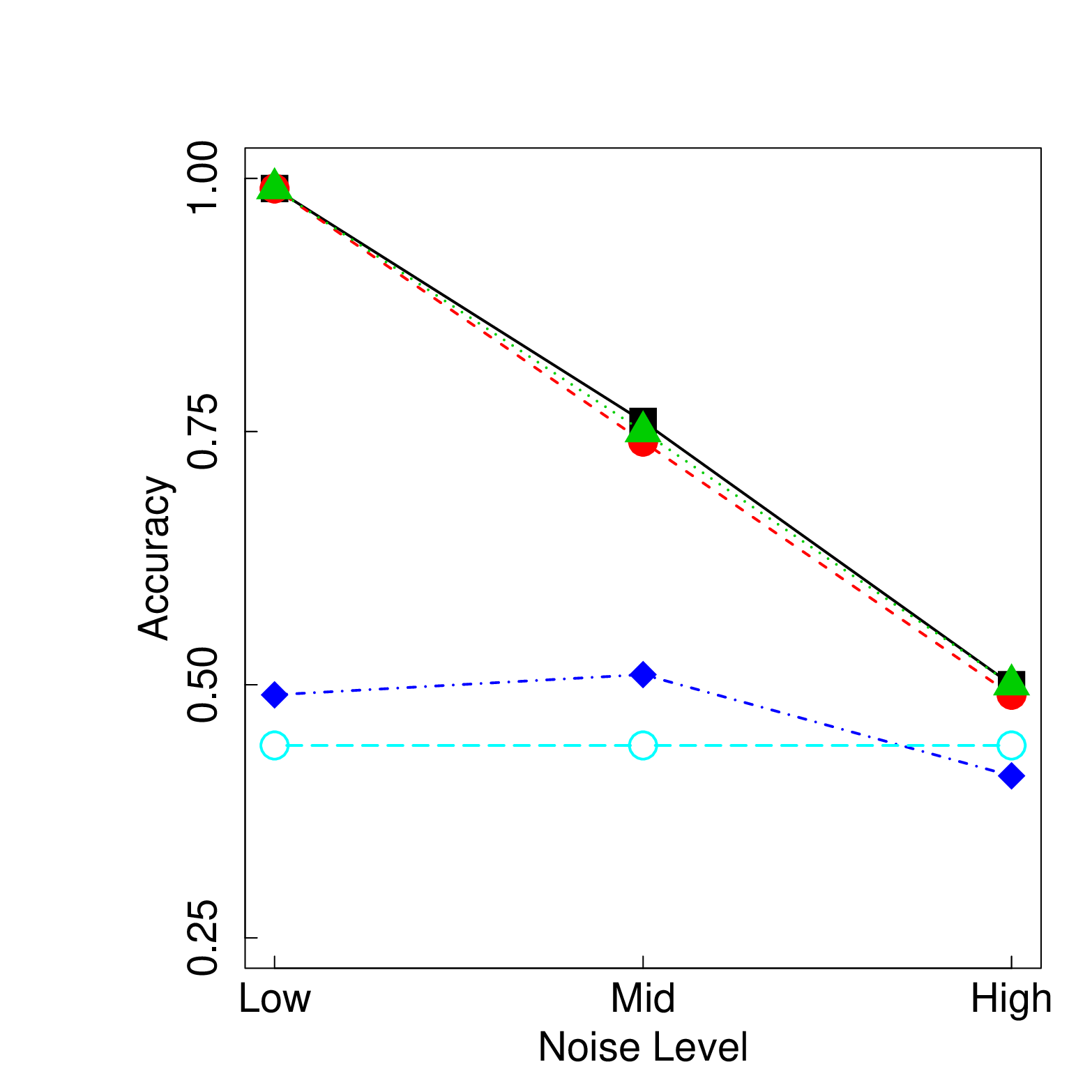}
}
\hspace{-0.3in}
\subfigure{
\includegraphics[width=1.5in]{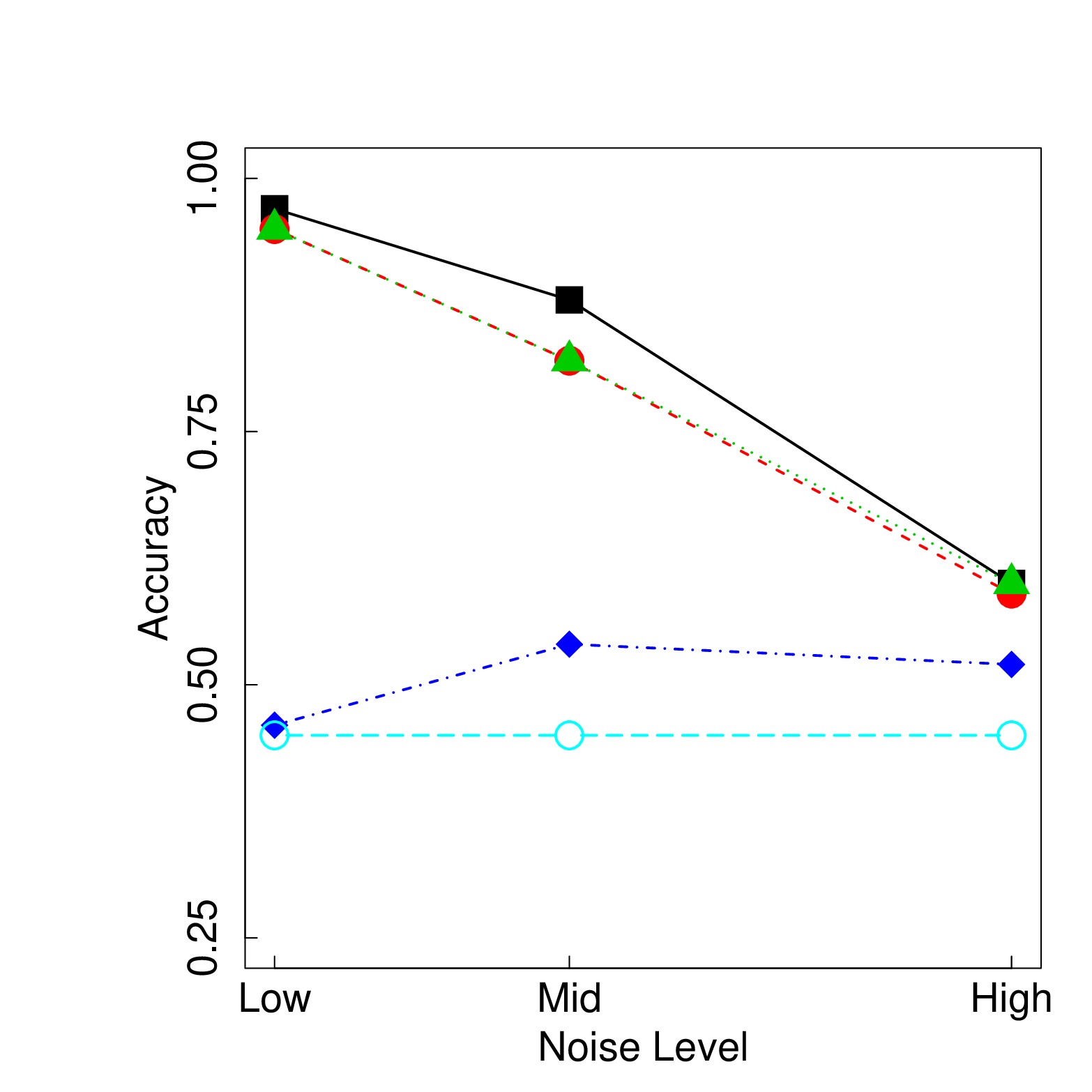}
} \\

\vspace{-.3in}

\subfigure{
\includegraphics[width=1.5in]{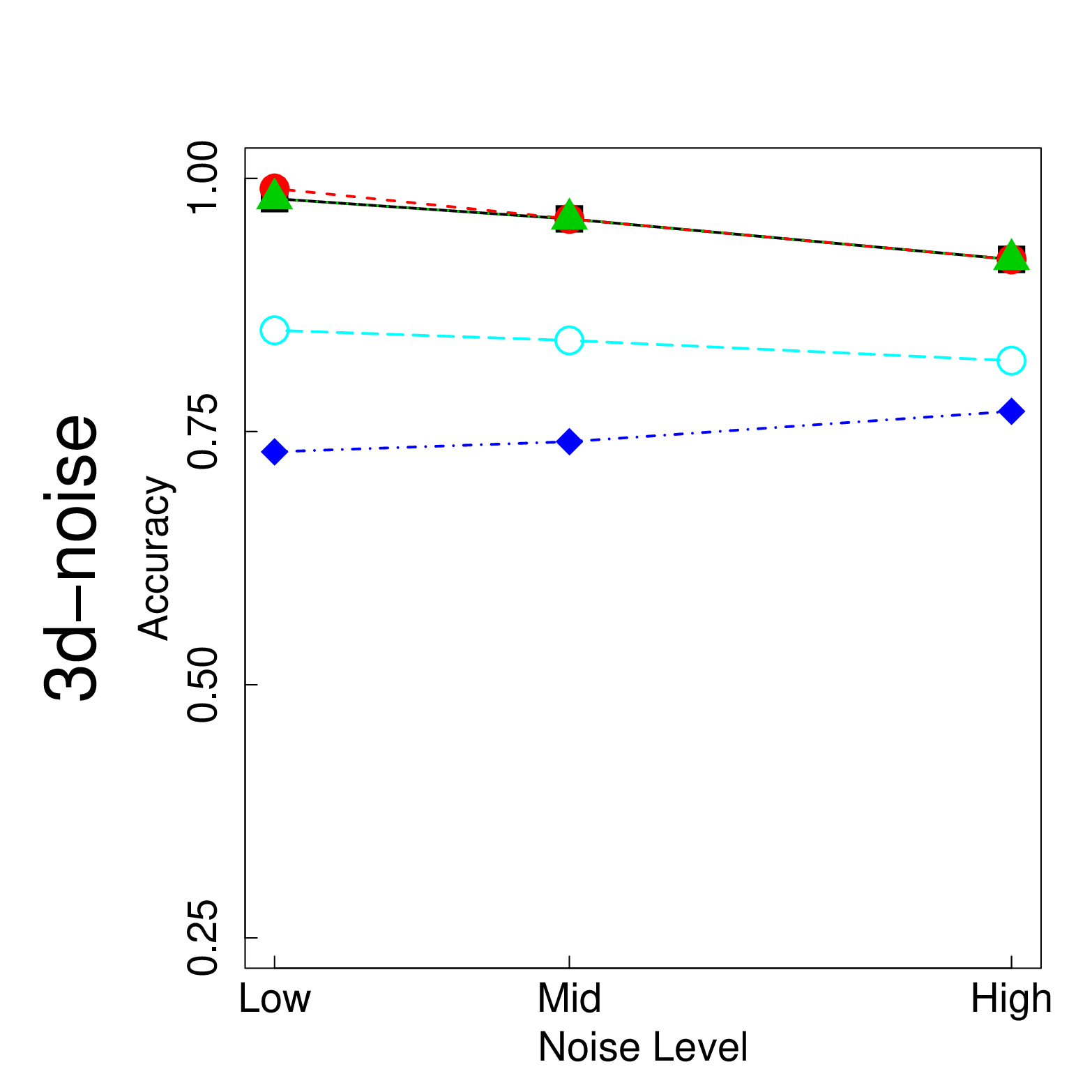}
}
\hspace{-0.3in}
\subfigure{
\includegraphics[width=1.5in]{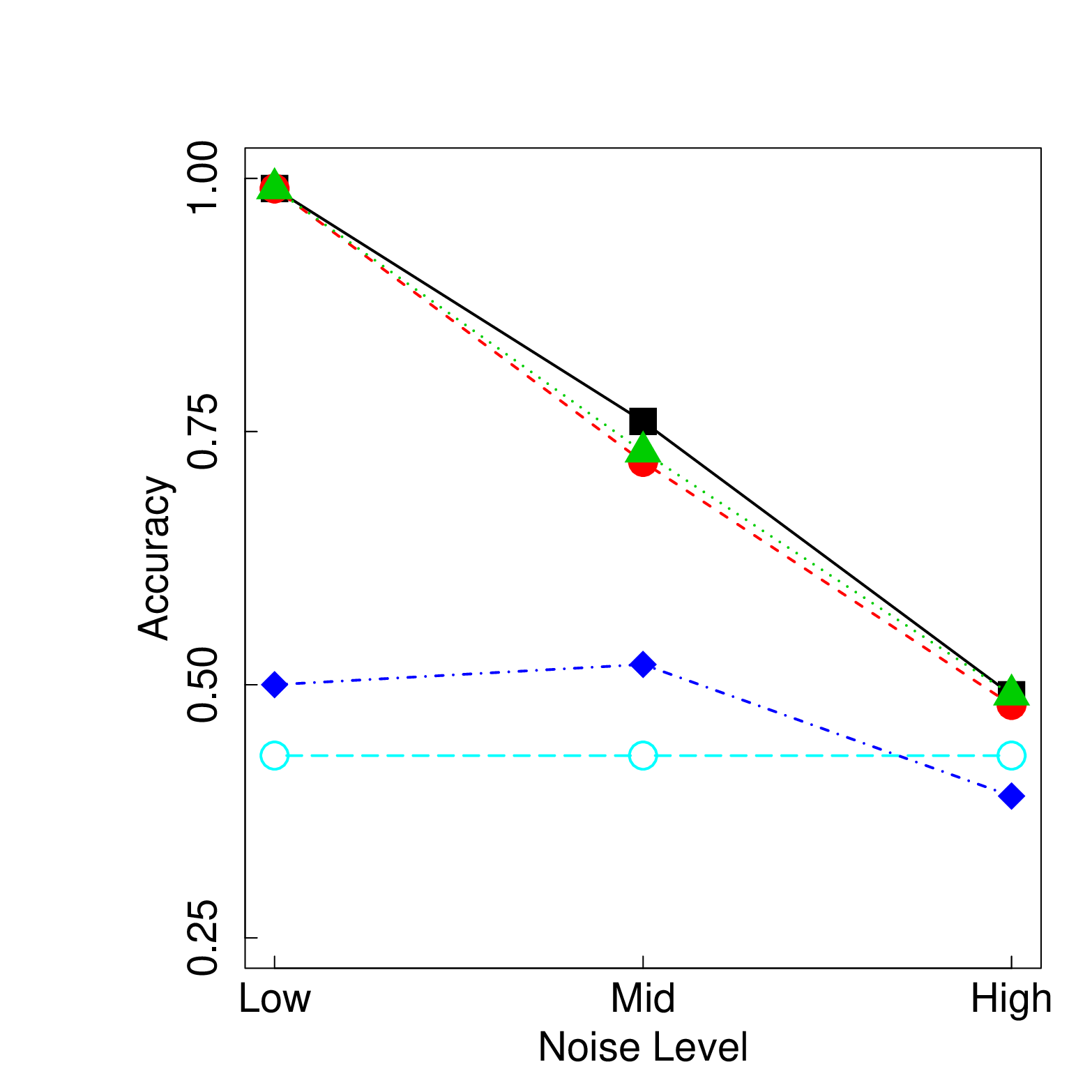}
}
\hspace{-0.3in}
\subfigure{
\includegraphics[width=1.5in]{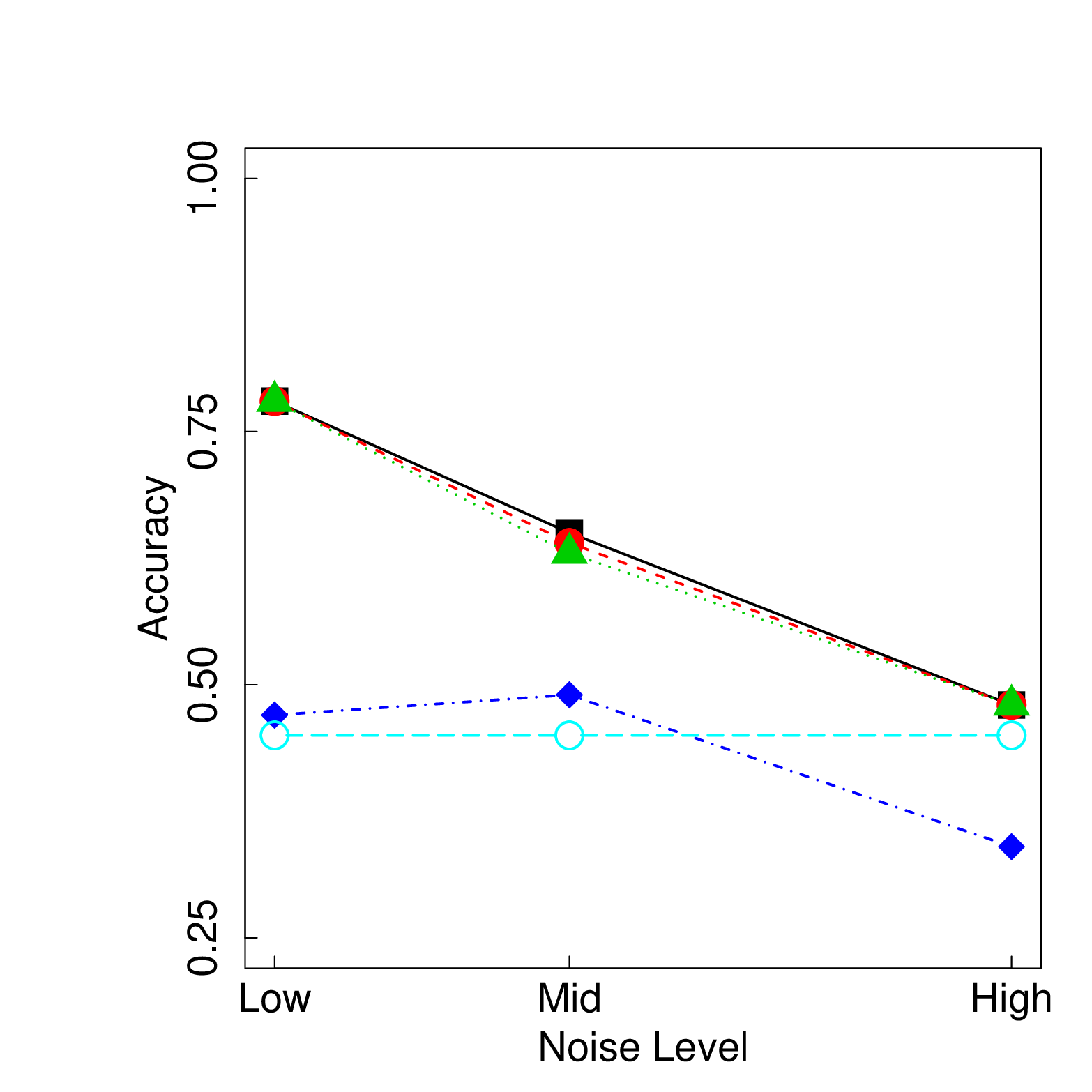}
} \\

\vspace{-.3in}

\subfigure{
\includegraphics[width=1.5in]{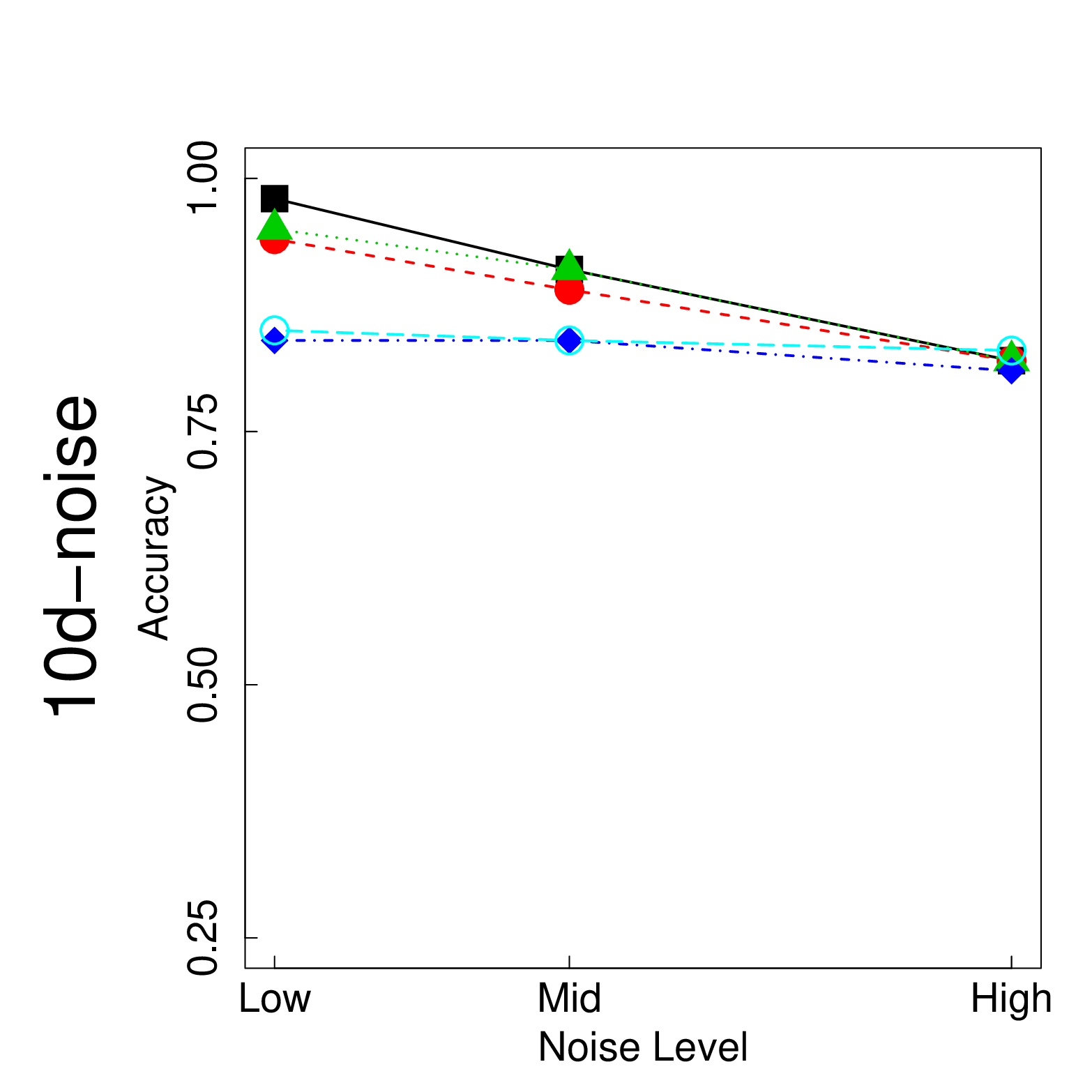}
}
\hspace{-0.3in}
\subfigure{
\includegraphics[width=1.5in]{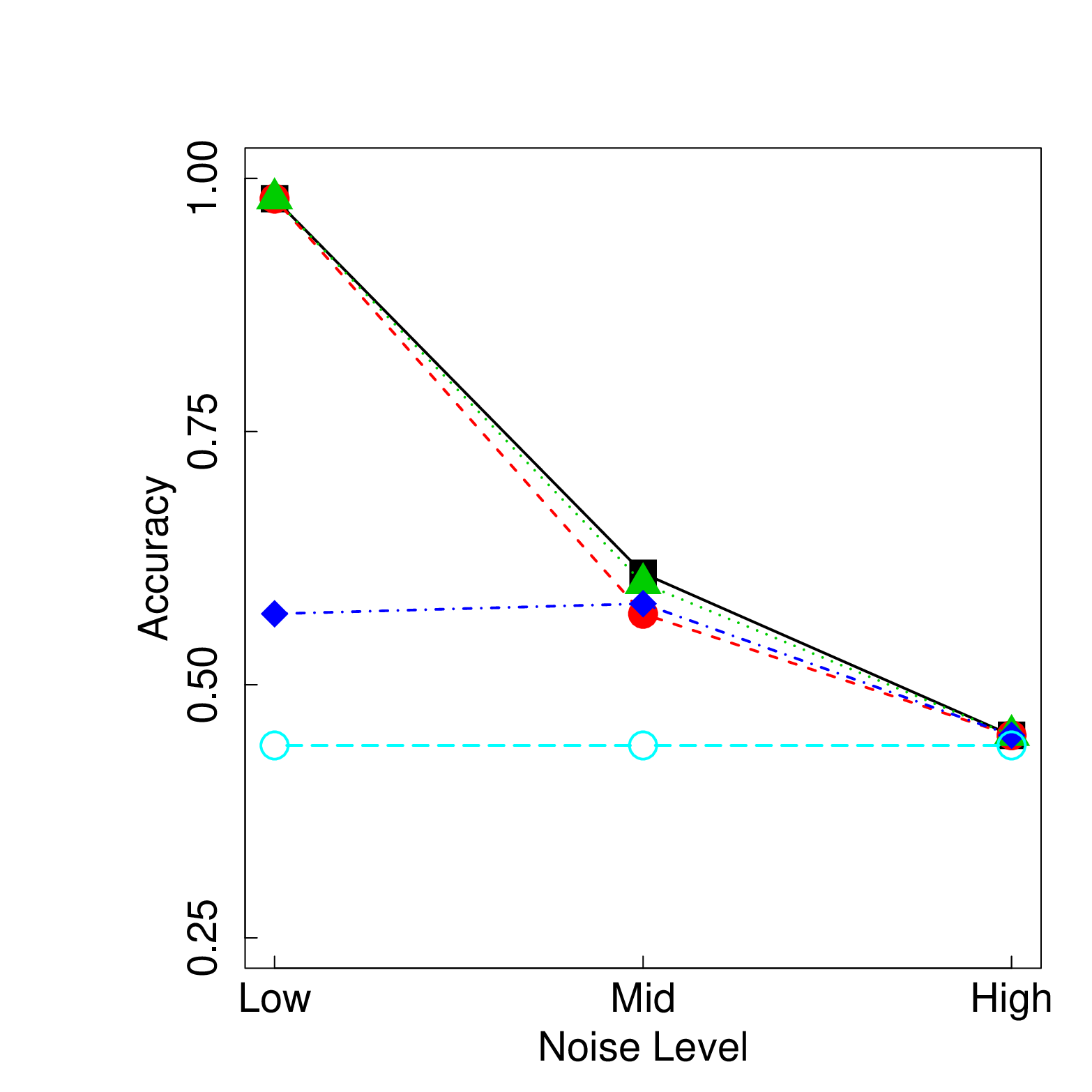}
}
\hspace{-0.3in}
\subfigure{
\includegraphics[width=1.5in]{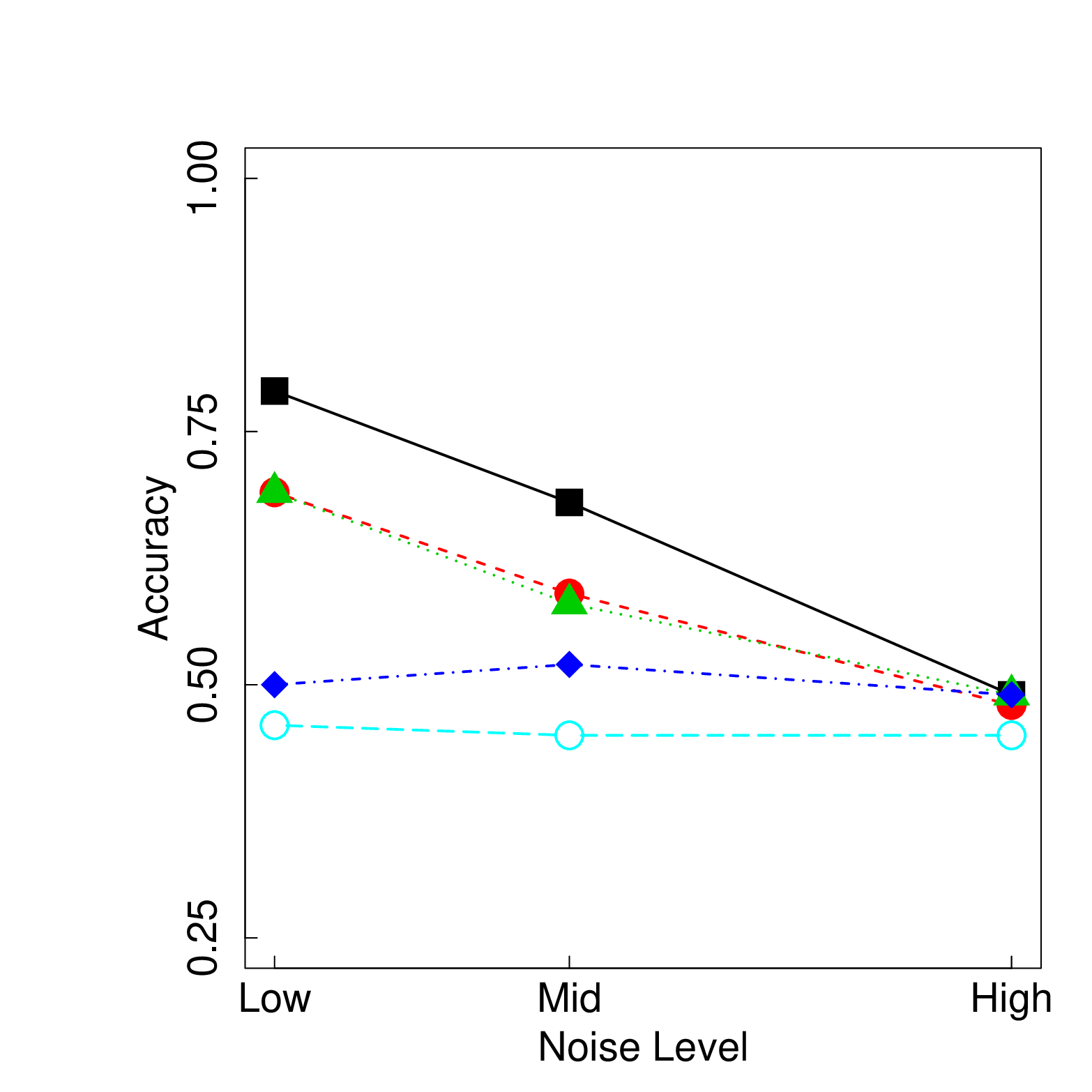}
} \\

\vspace{-.1in}

\end{center}

\caption{Comparison of the different connection schemes. Columns of sub-figures correspond, from left to right, to the Two-arcs, Three-spirals and Three-rings datasets. Rows correspond to the four different embeddings; from top to bottom: 2d, 3d, 3d--noise and 10d--noise. Each sub-figure shows the mean clustering accuracy of the diverse methods at three different noise levels.}
\label{schemes}
\end{figure*}

\begin{figure*}
\begin{center}

\subfigure{
\includegraphics[width=1.75in]{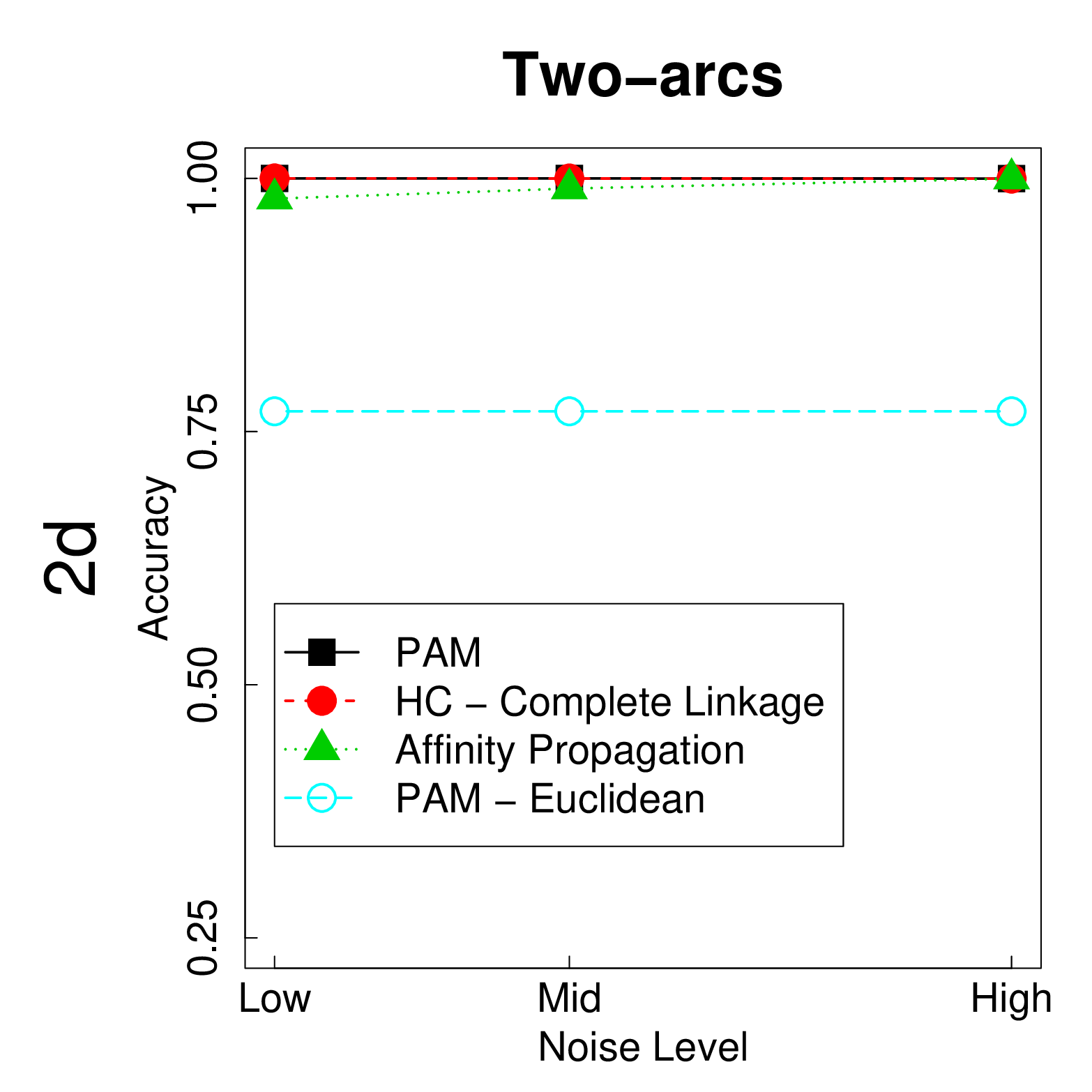}
}
\hspace{-0.3in}
\subfigure{
\includegraphics[width=1.75in]{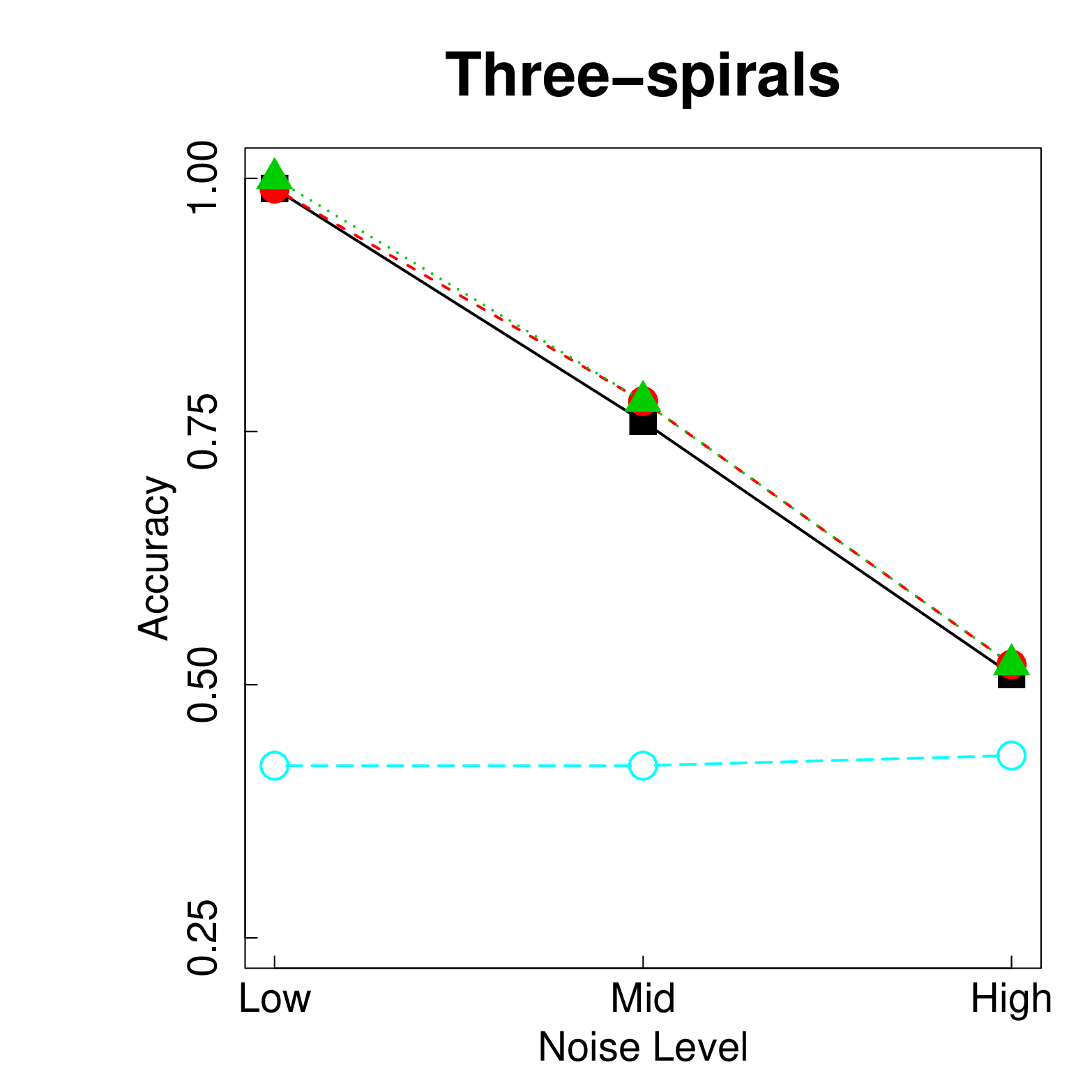}
}
\hspace{-0.3in}
\subfigure{
\includegraphics[width=1.75in]{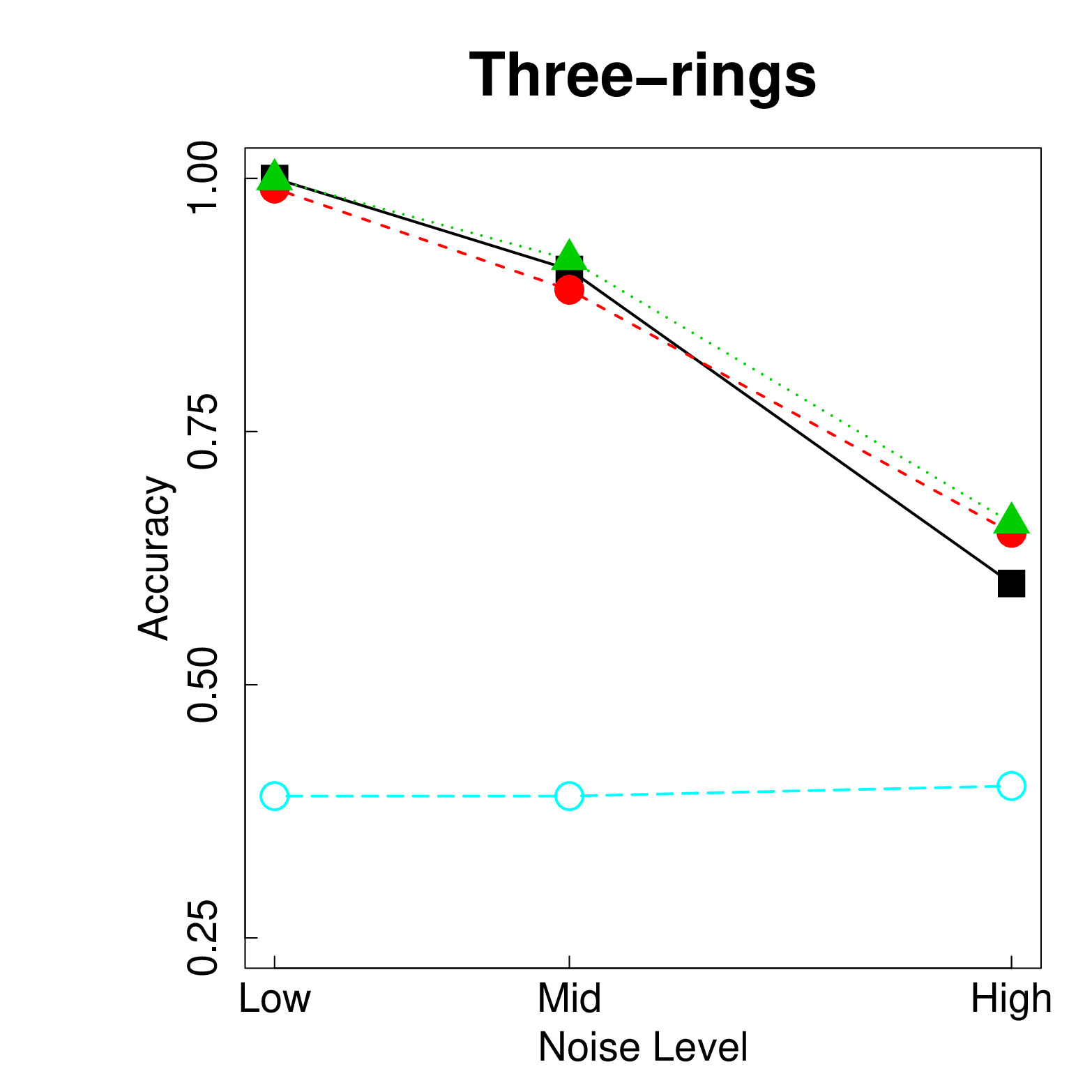}
} \\

\vspace{-.3in}

\subfigure{
\includegraphics[width=1.75in]{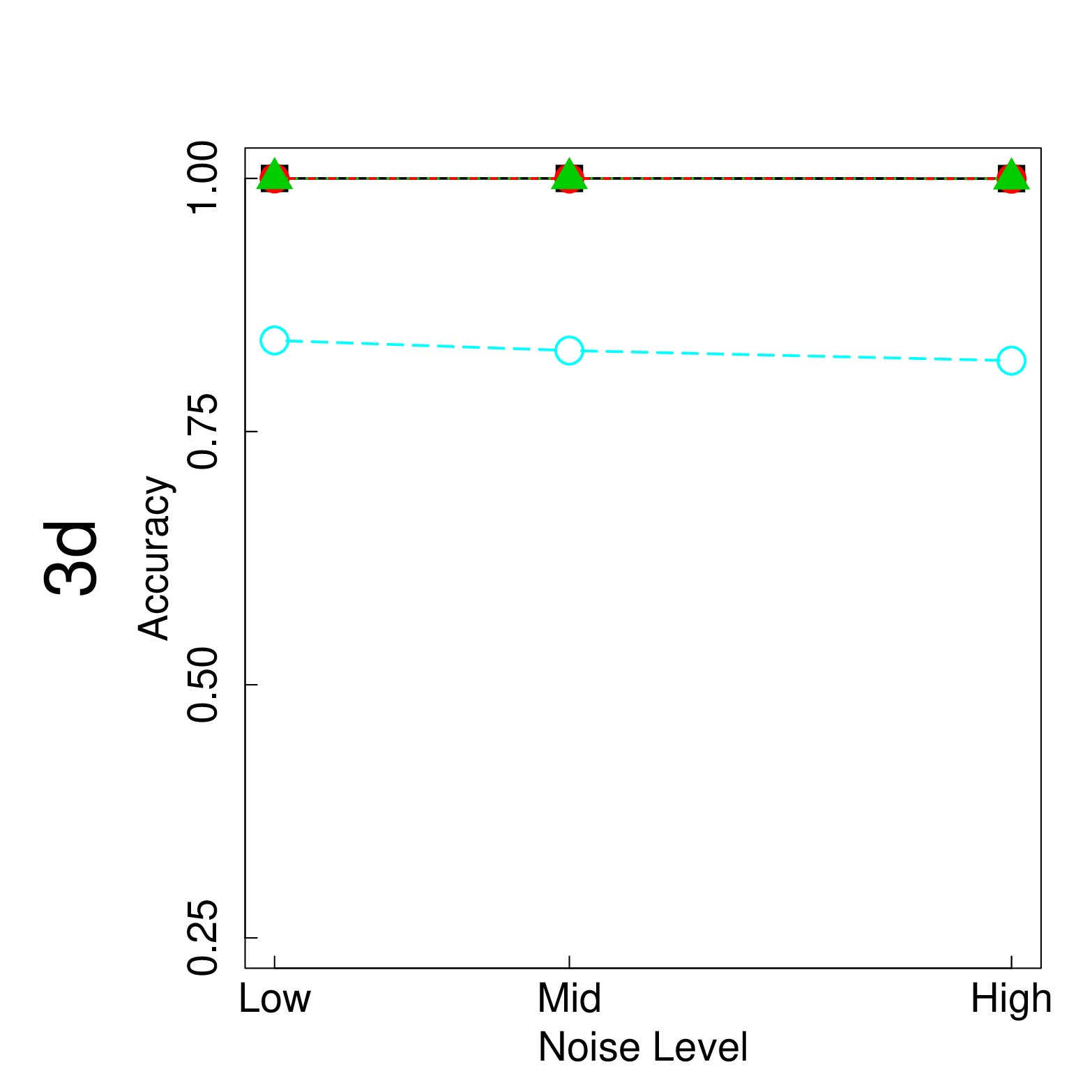}
}
\hspace{-0.3in}
\subfigure{
\includegraphics[width=1.75in]{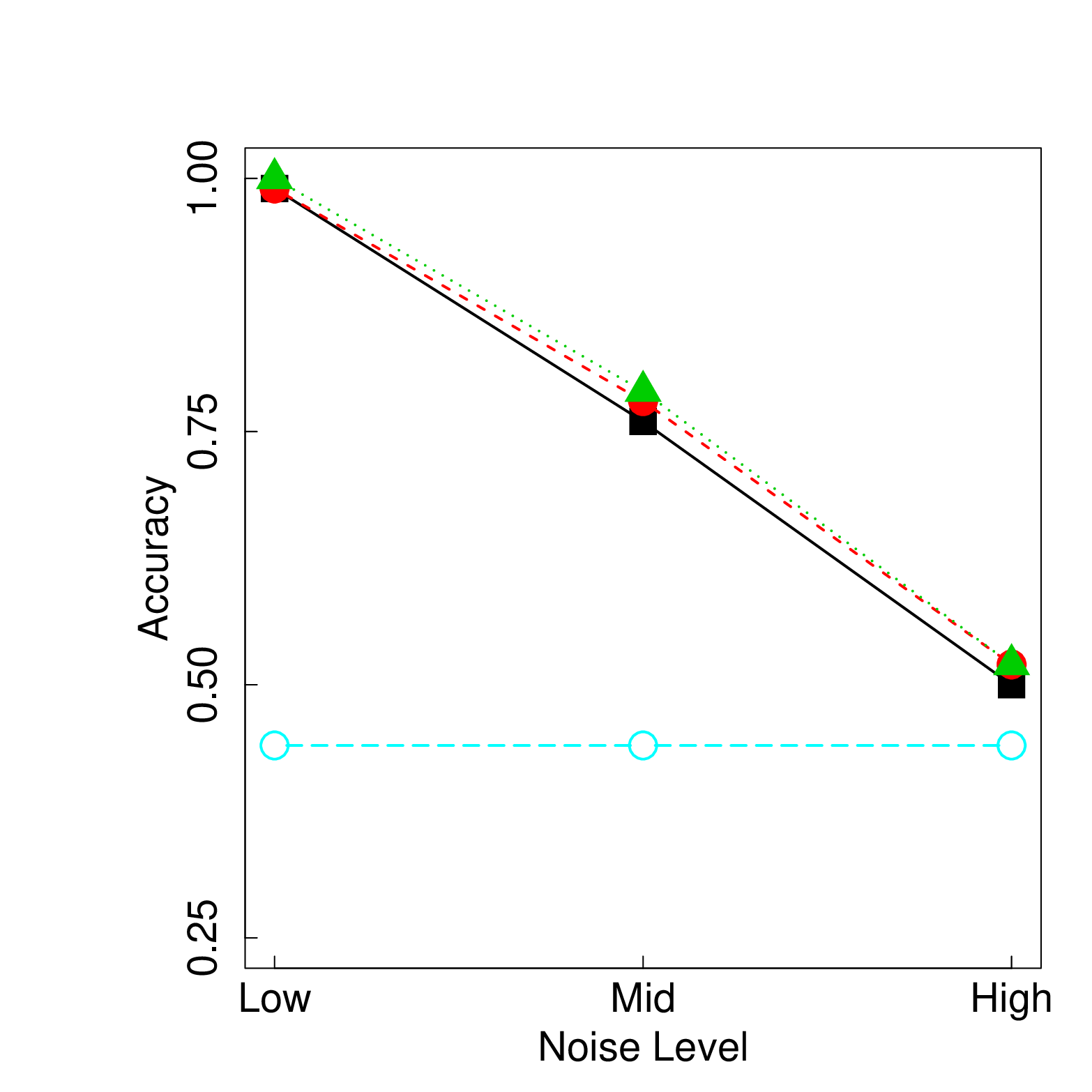}
}
\hspace{-0.3in}
\subfigure{
\includegraphics[width=1.75in]{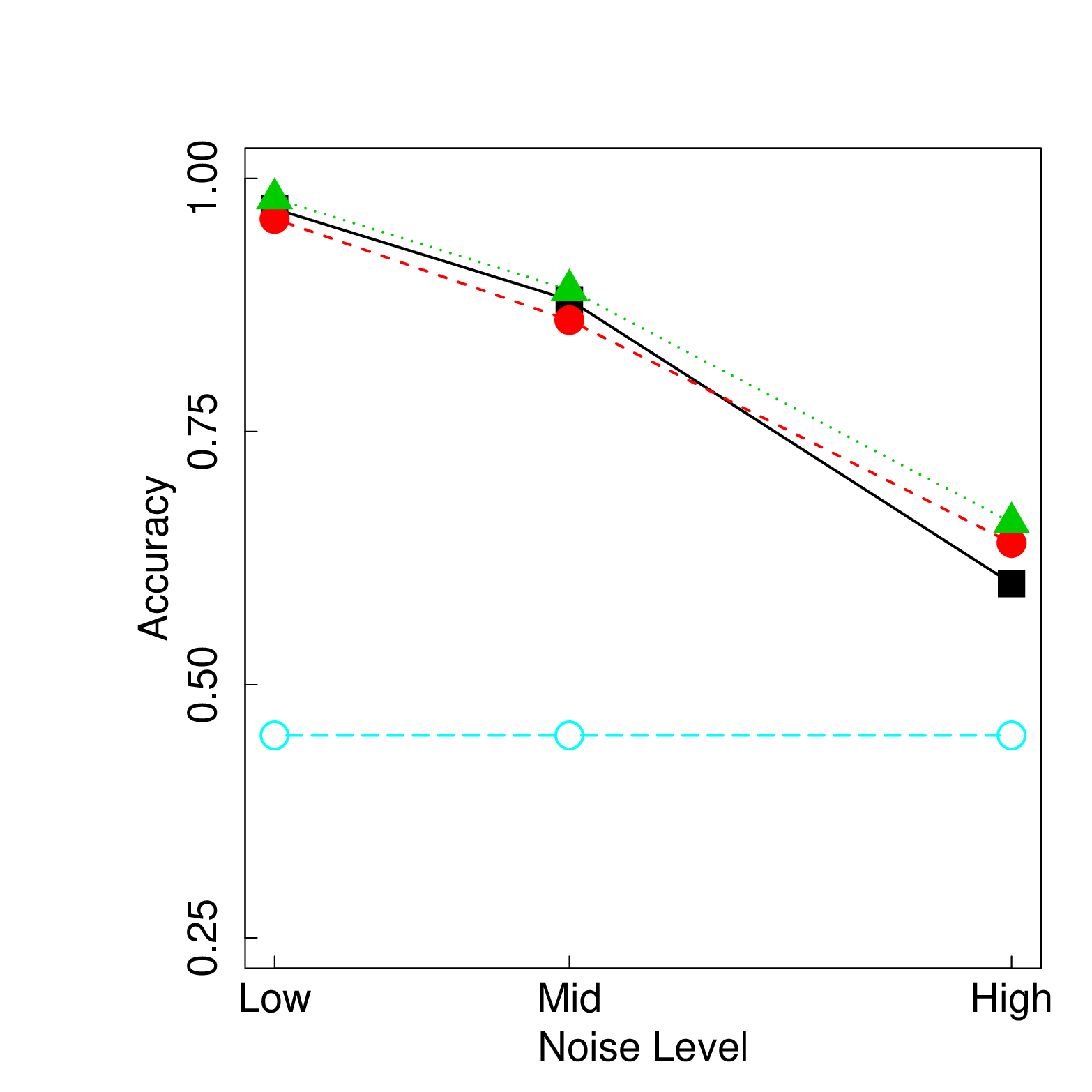}
} \\

\vspace{-.3in}

\subfigure{
\includegraphics[width=1.75in]{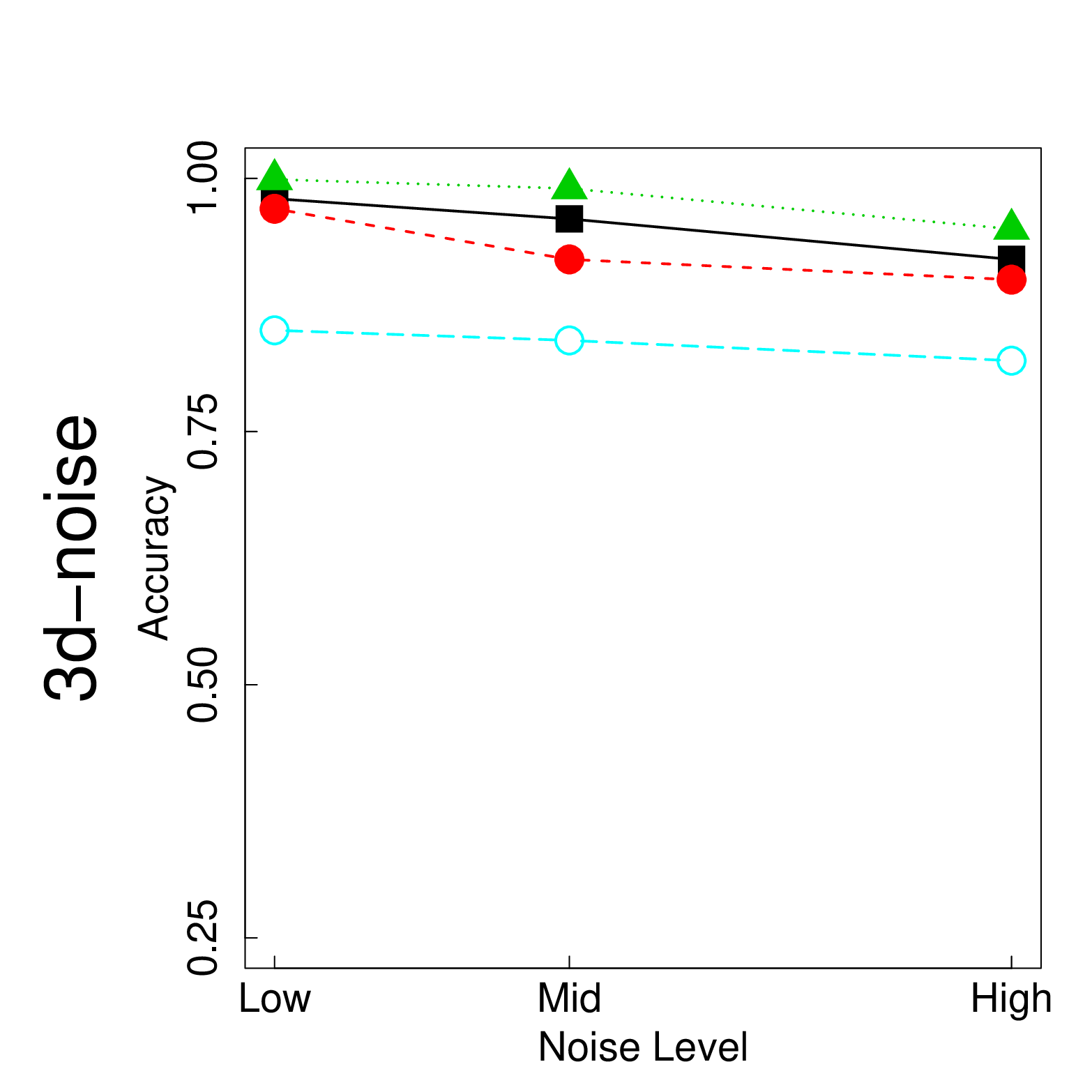}
}
\hspace{-0.3in}
\subfigure{
\includegraphics[width=1.75in]{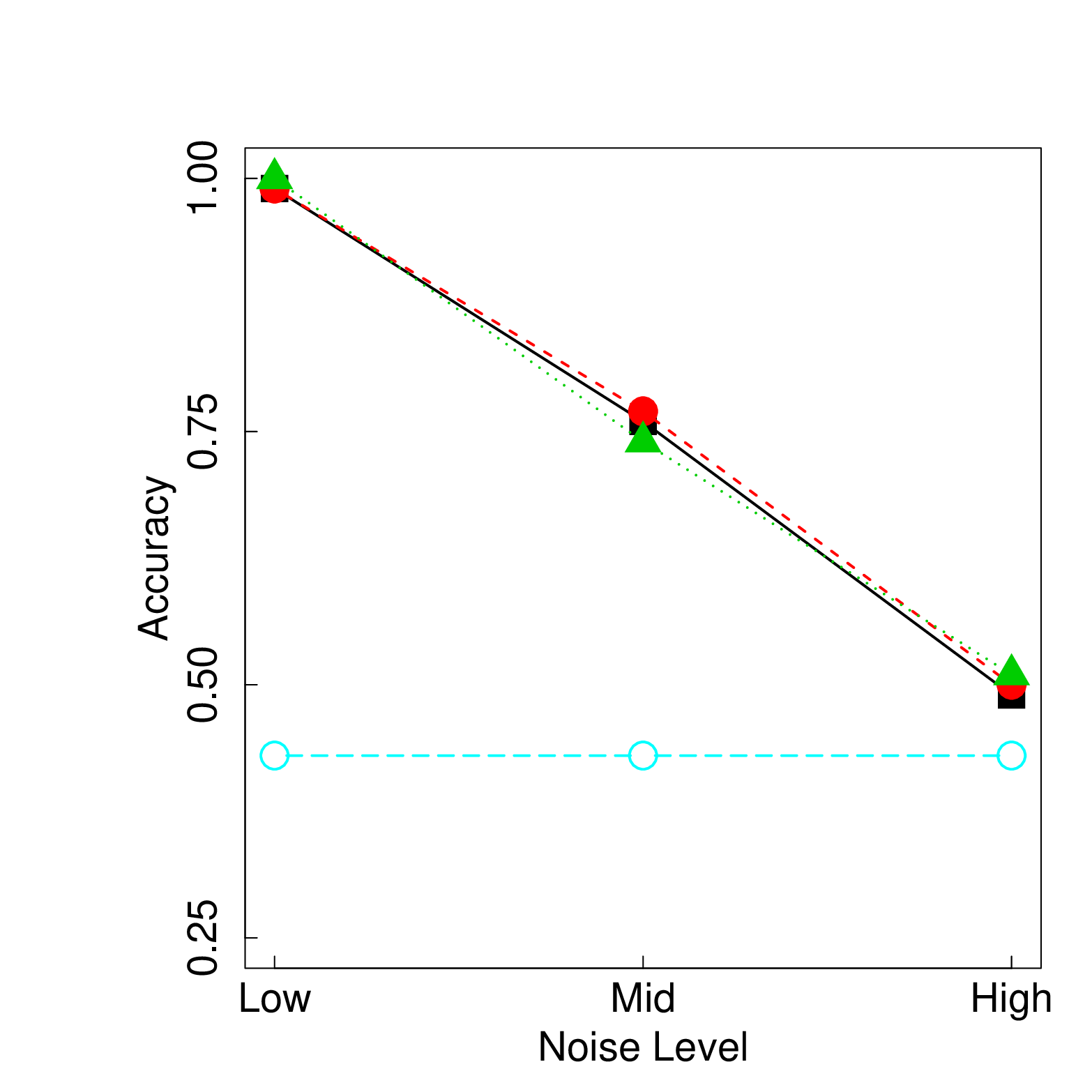}
}
\hspace{-0.3in}
\subfigure{
\includegraphics[width=1.75in]{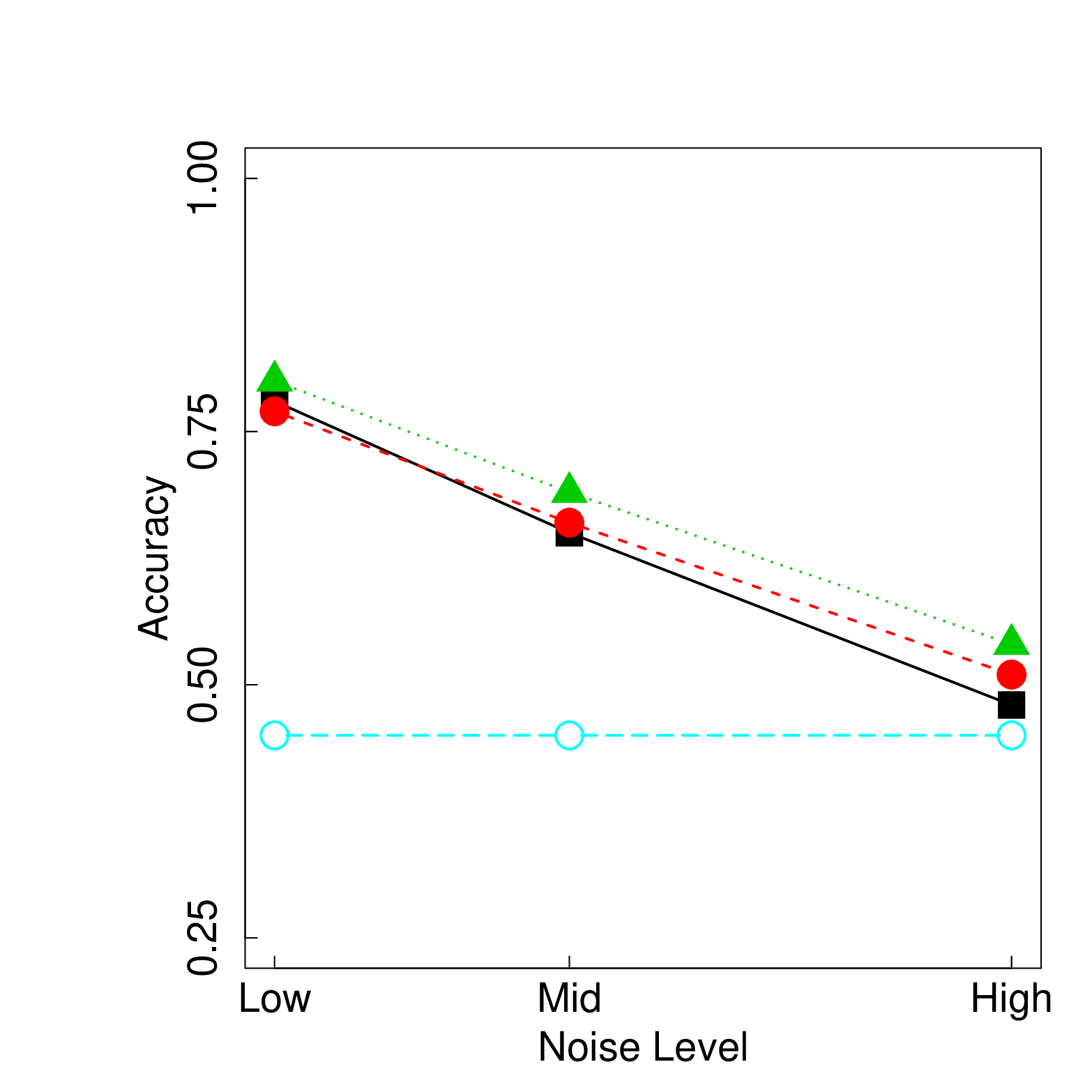}
} \\

\vspace{-.3in}

\subfigure{
\includegraphics[width=1.75in]{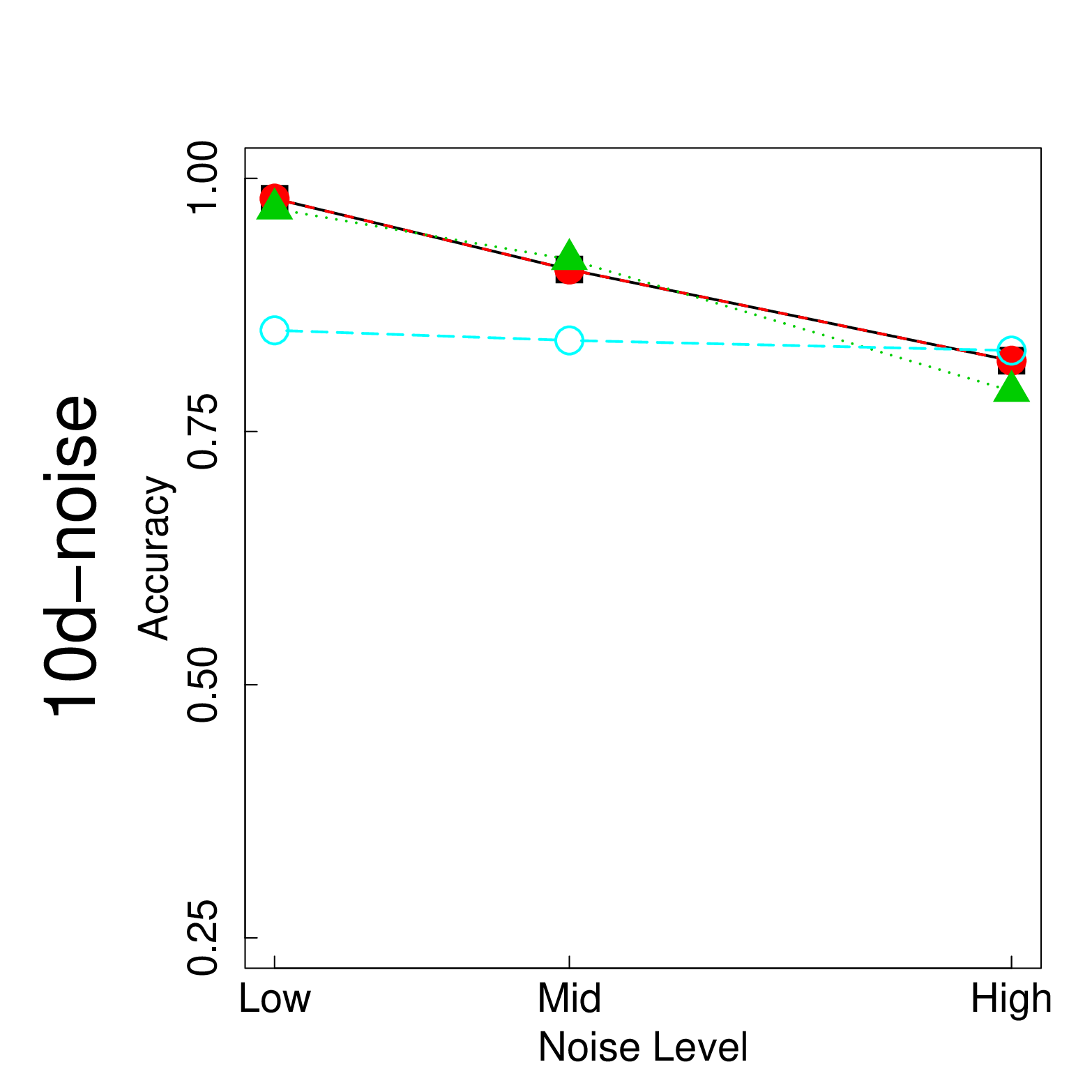}
}
\hspace{-0.3in}
\subfigure{
\includegraphics[width=1.75in]{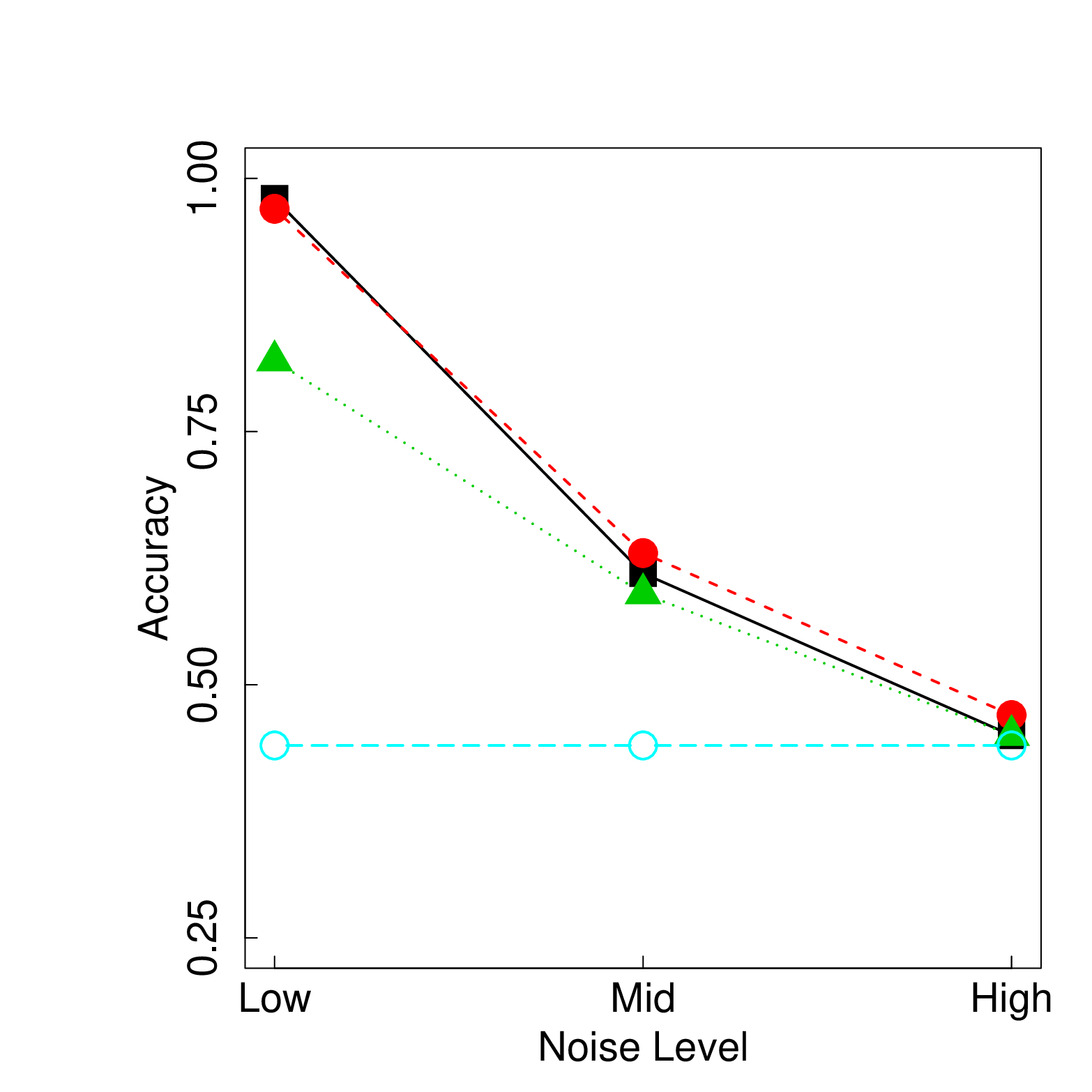}
}
\hspace{-0.3in}
\subfigure{
\includegraphics[width=1.75in]{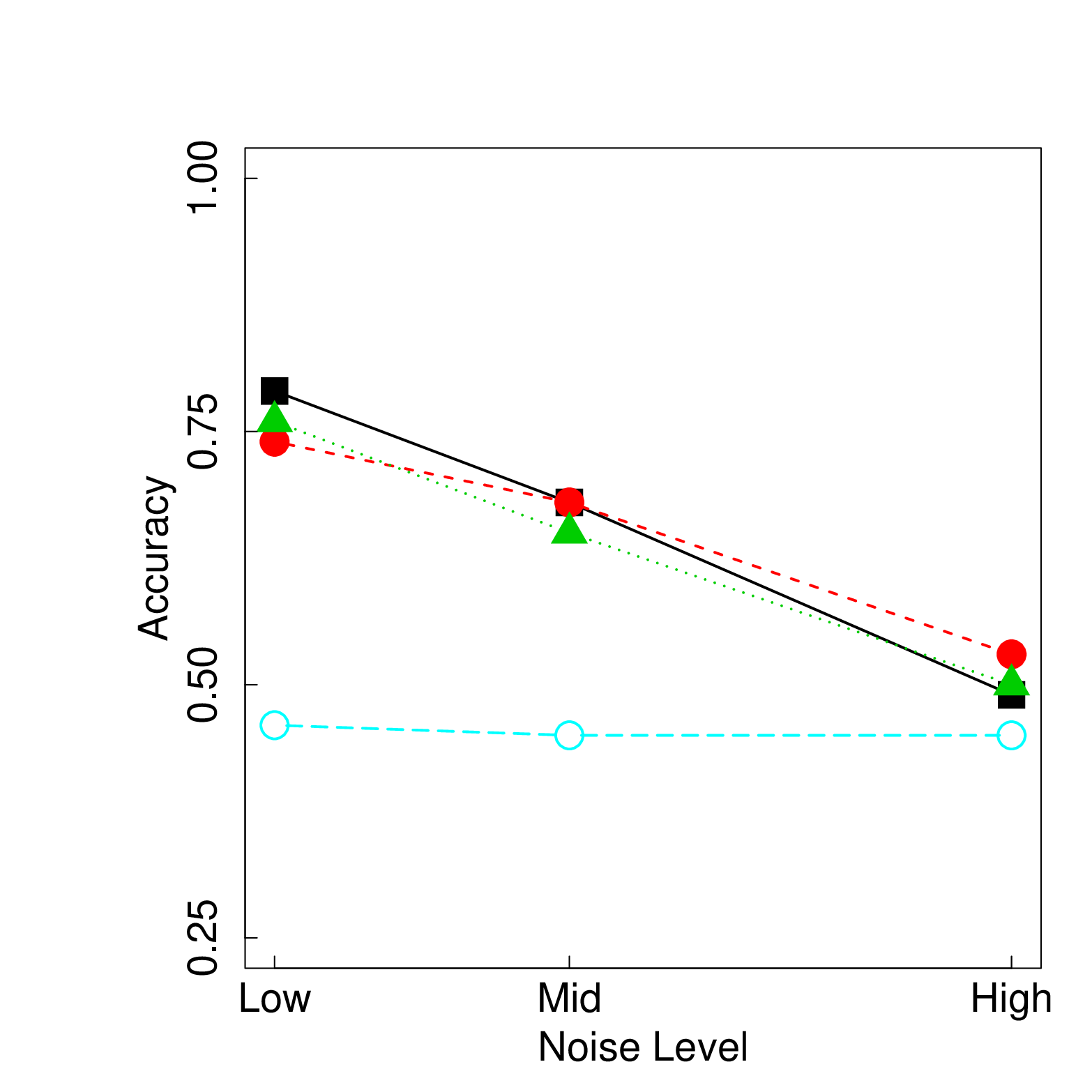}
} \\

\vspace{-.1in}

\end{center}

\caption{Comparison of three different clustering algorithms applied to similarities evaluated with the PKNNG (MinSpan) metric. The arrangement is similar to Figure \ref{schemes}.}
\label{clust_methods}
\end{figure*}

\begin{figure*}
\begin{center}

\subfigure{
\includegraphics[width=1.75in]{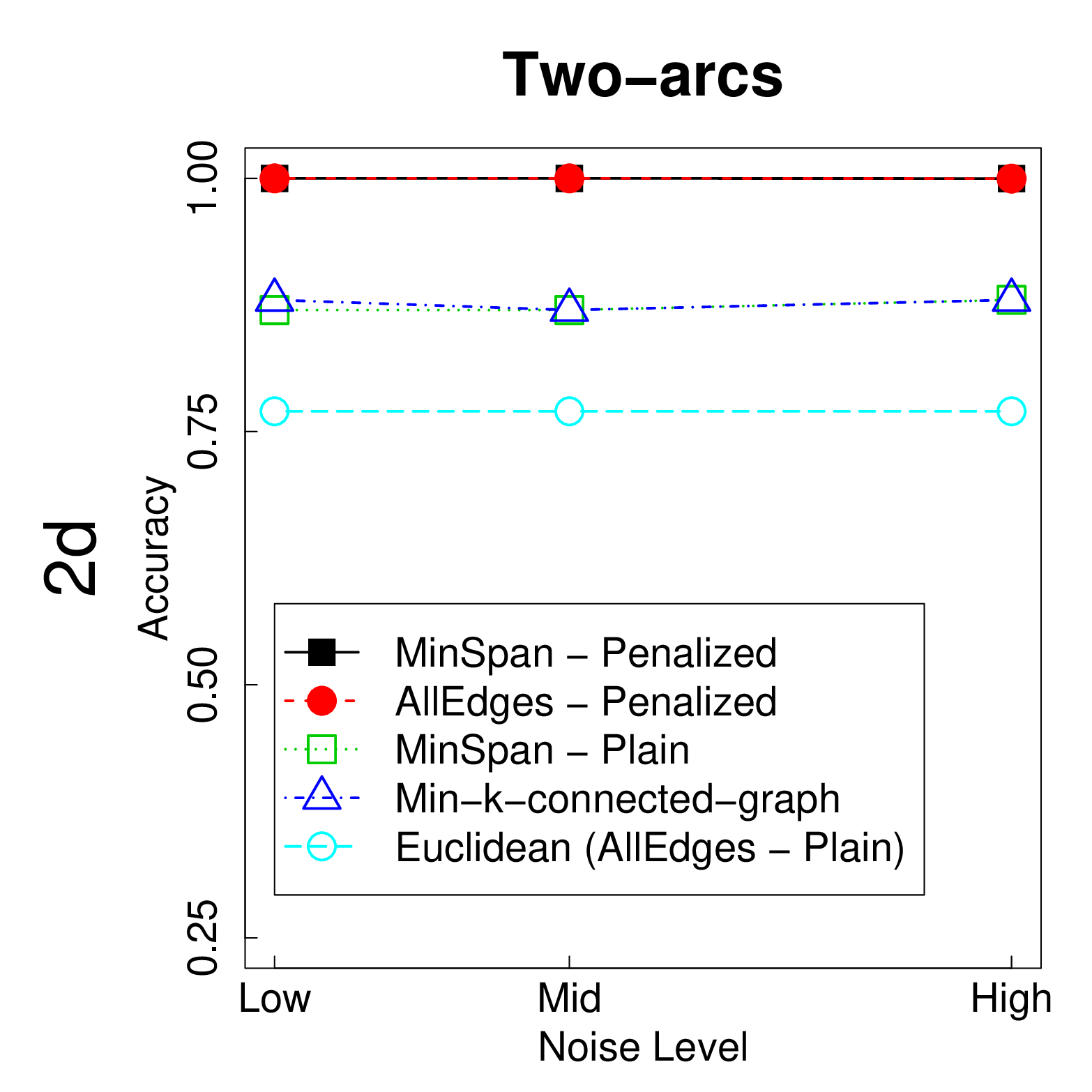}
}
\hspace{-0.3in}
\subfigure{
\includegraphics[width=1.75in]{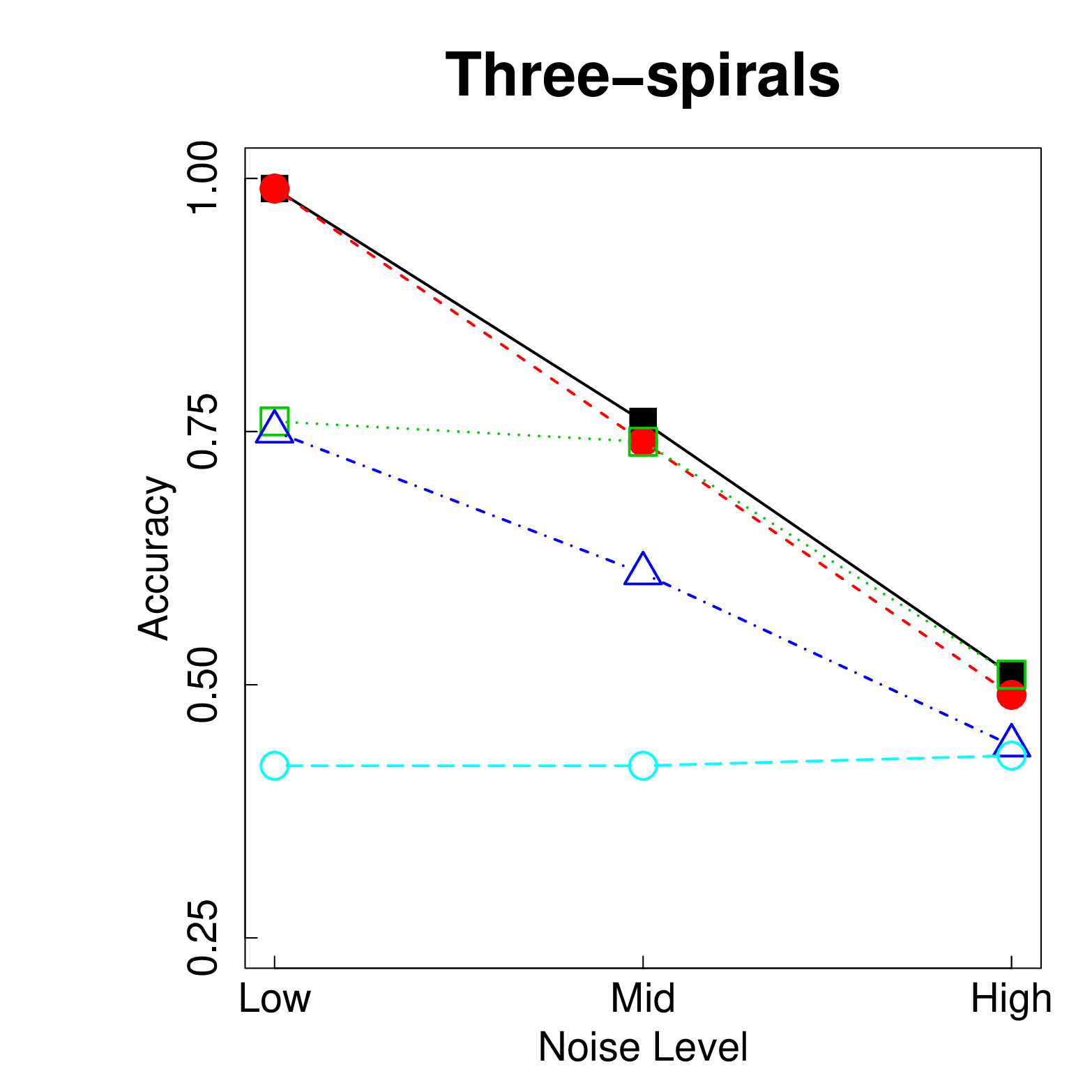}
}
\hspace{-0.3in}
\subfigure{
\includegraphics[width=1.75in]{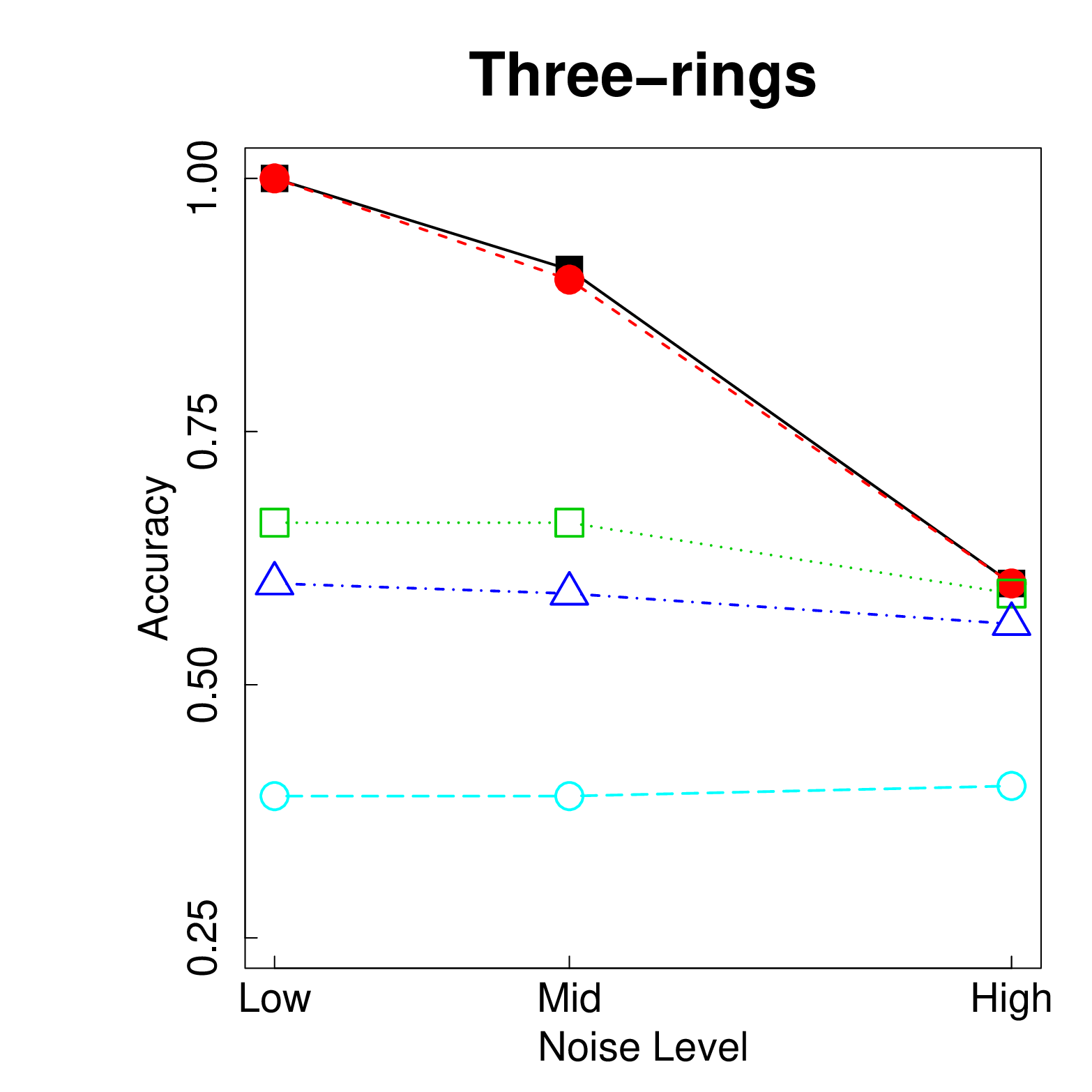}
} \\

\vspace{-.3in}

\subfigure{
\includegraphics[width=1.75in]{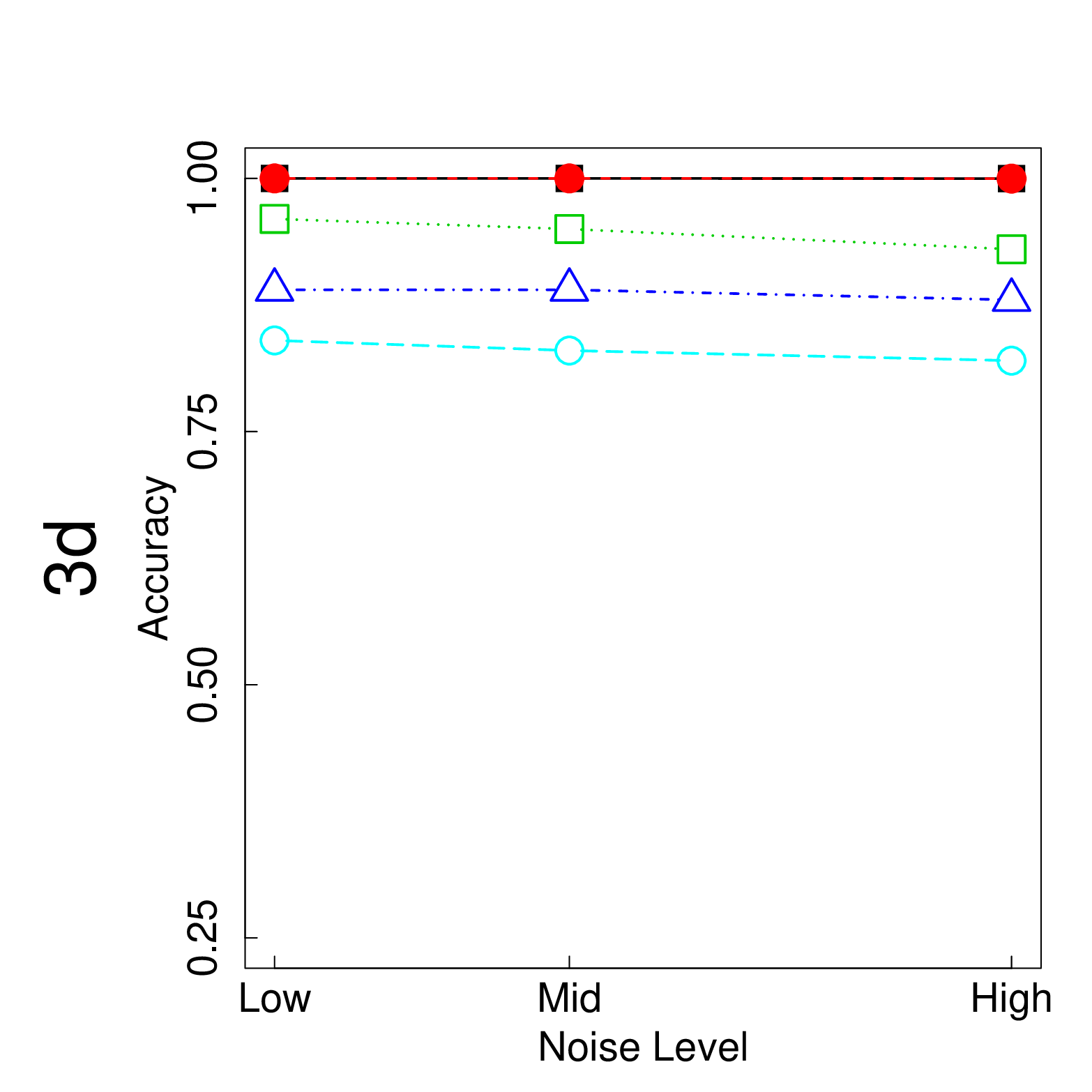}
}
\hspace{-0.3in}
\subfigure{
\includegraphics[width=1.75in]{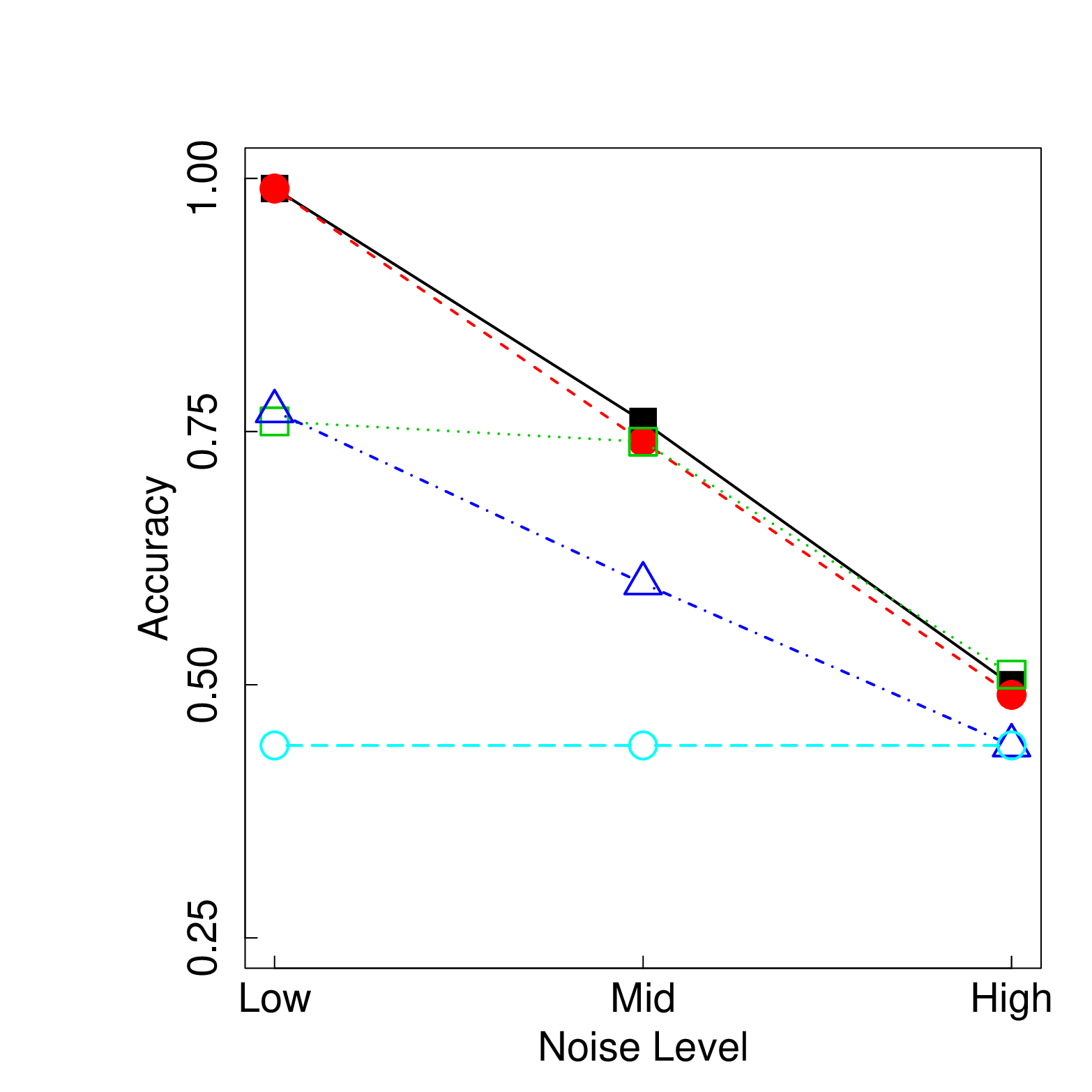}
}
\hspace{-0.3in}
\subfigure{
\includegraphics[width=1.75in]{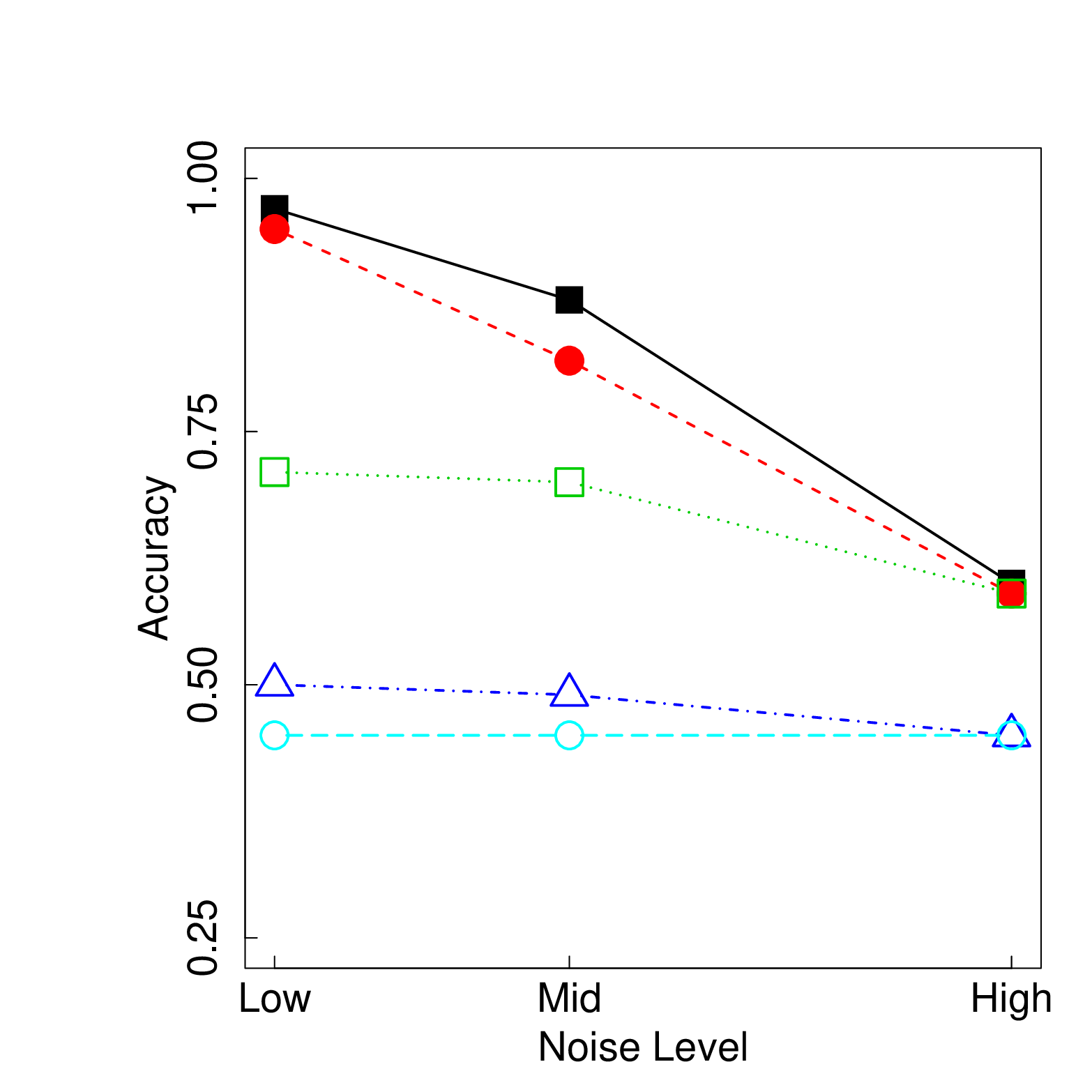}
} \\

\vspace{-.3in}

\subfigure{
\includegraphics[width=1.75in]{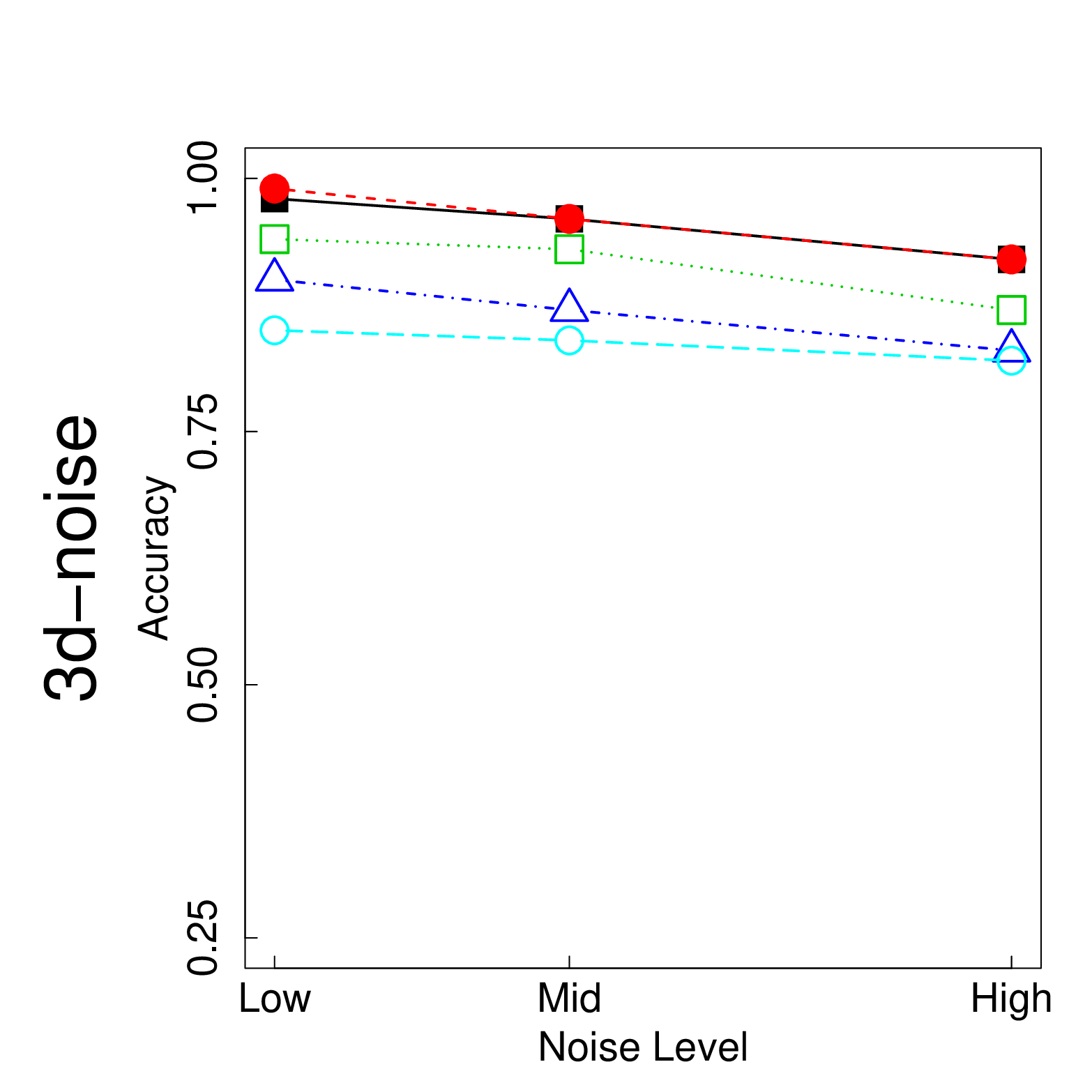}
}
\hspace{-0.3in}
\subfigure{
\includegraphics[width=1.75in]{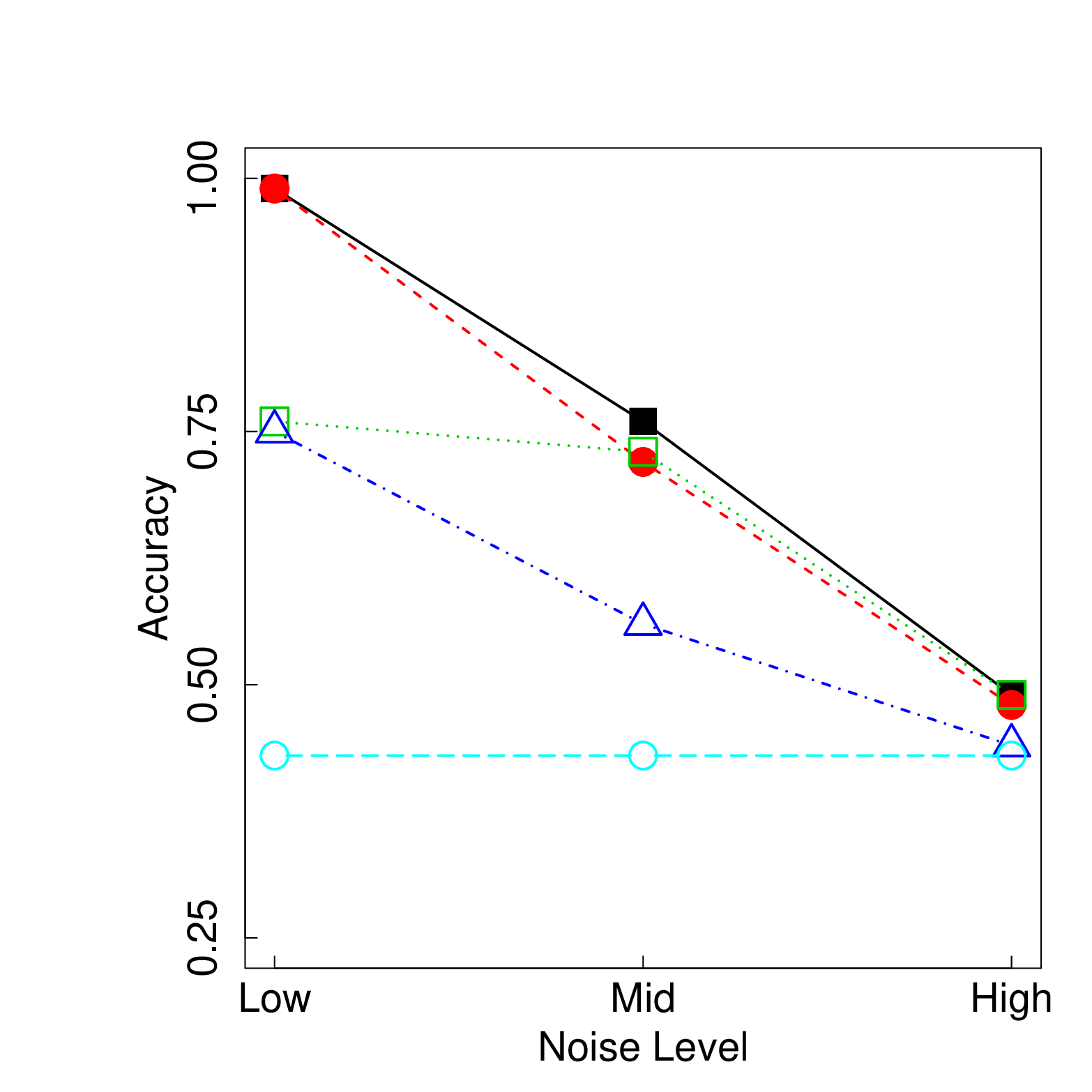}
}
\hspace{-0.3in}
\subfigure{
\includegraphics[width=1.75in]{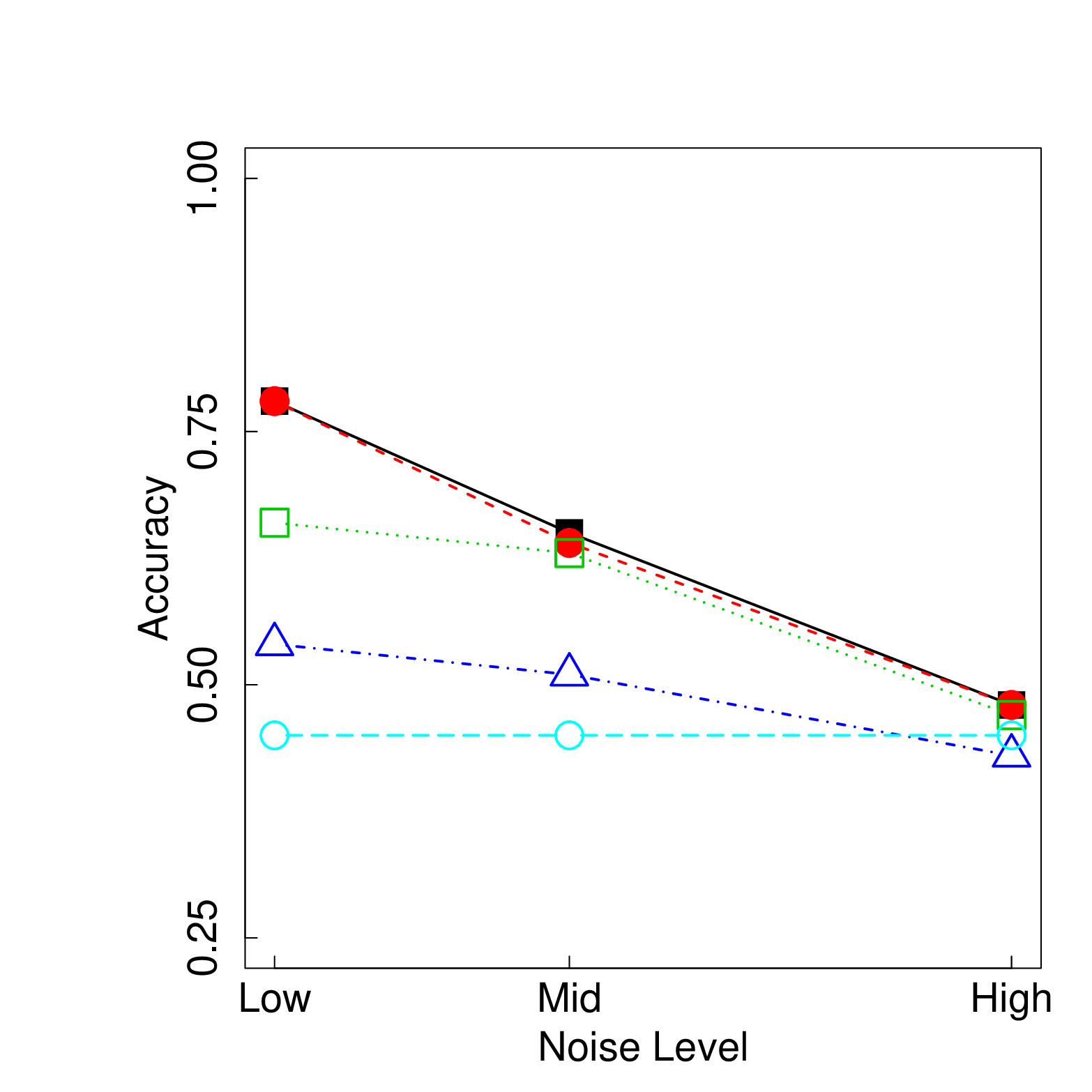}
} \\

\vspace{-.3in}

\subfigure{
\includegraphics[width=1.75in]{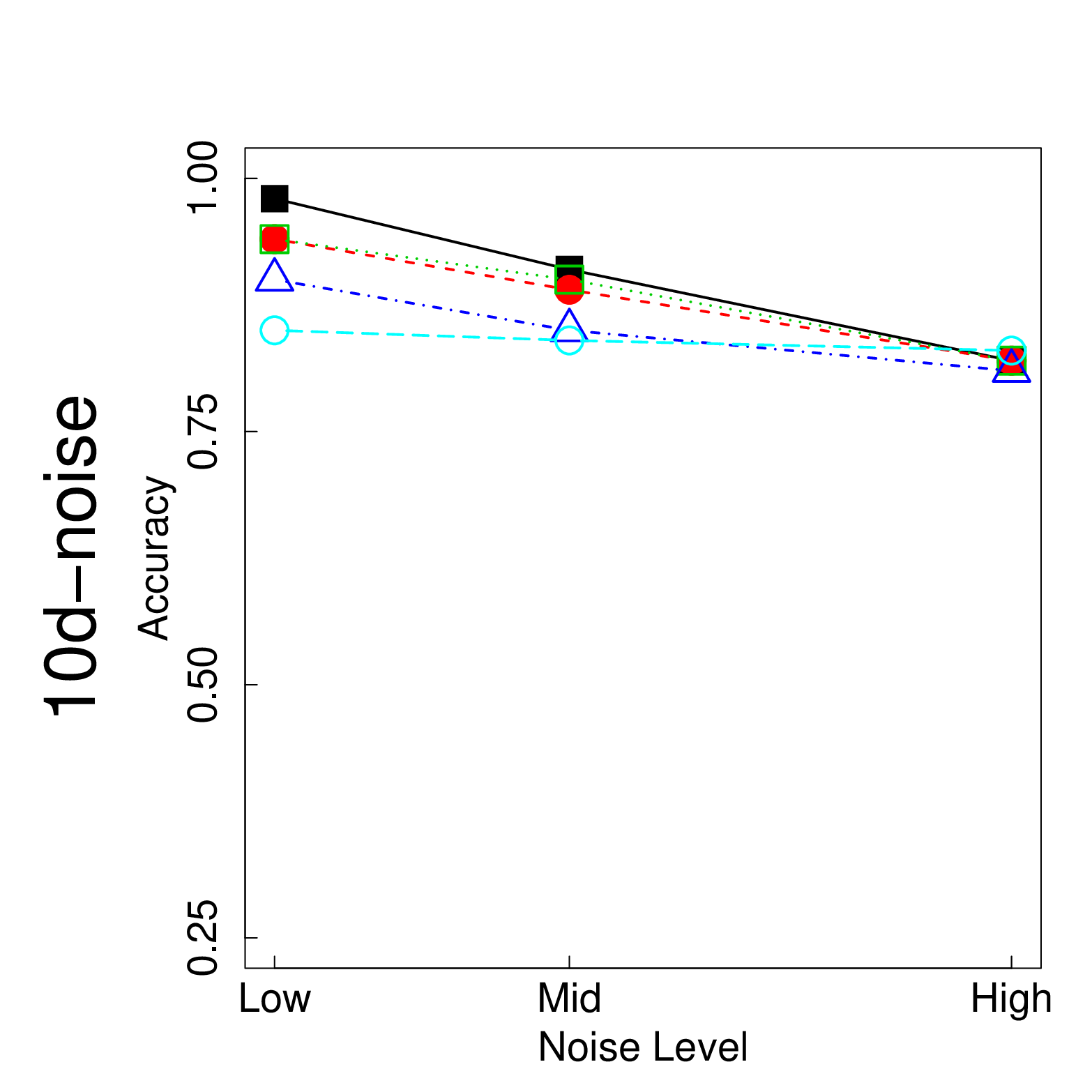}
}
\hspace{-0.3in}
\subfigure{
\includegraphics[width=1.75in]{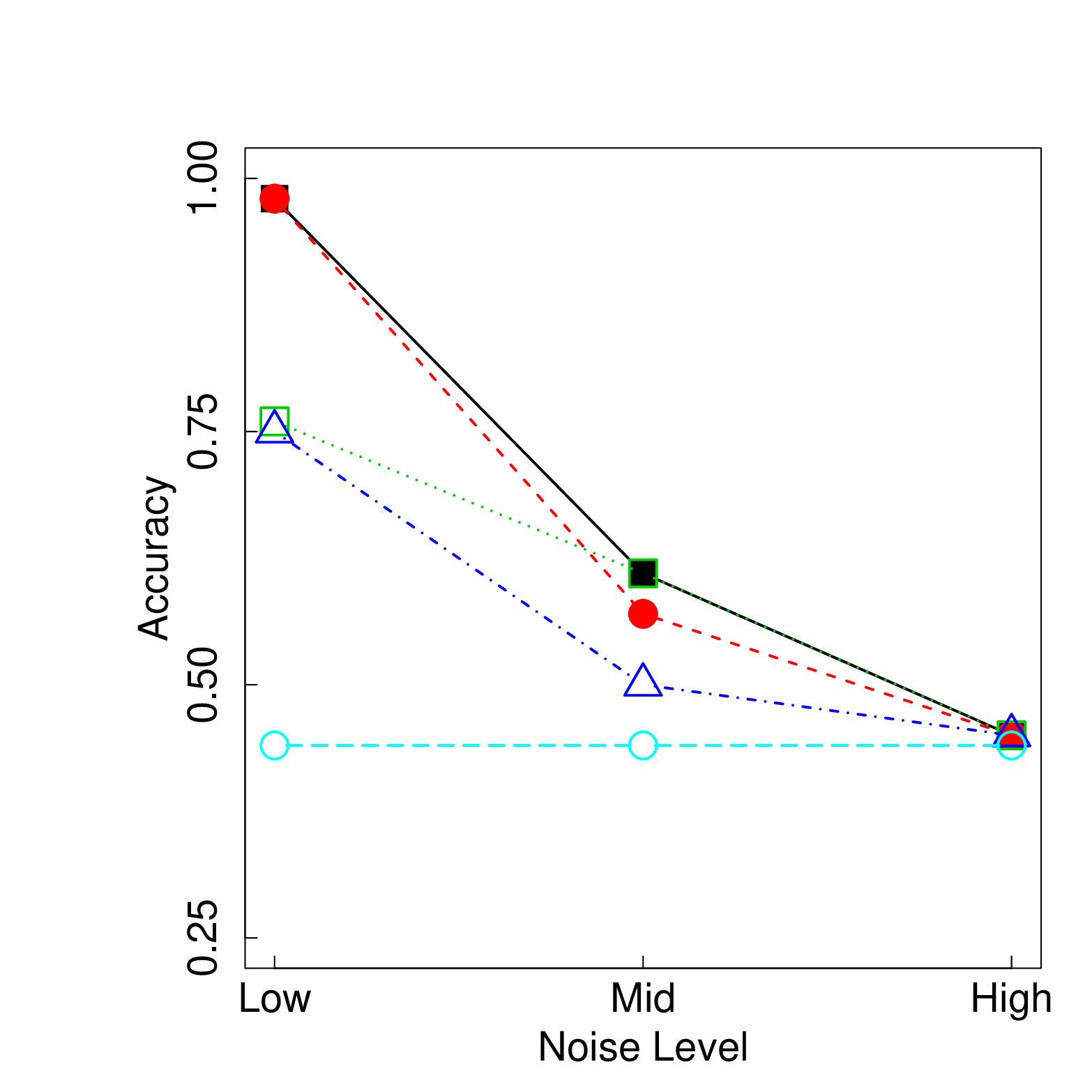}
}
\hspace{-0.3in}
\subfigure{
\includegraphics[width=1.75in]{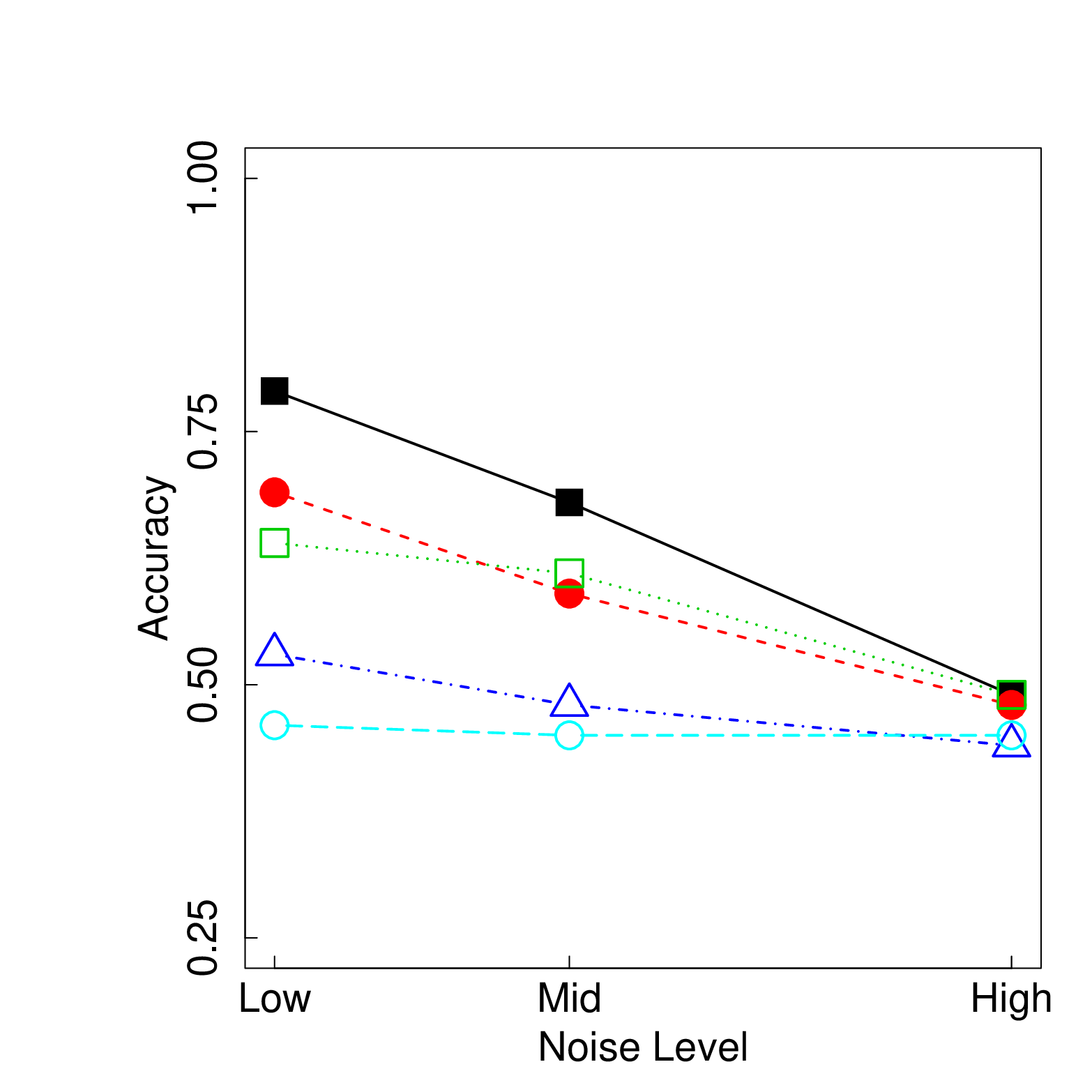}
} \\

\vspace{-.1in}

\end{center}

\caption{Comparison of two connection schemes (MinSpan and AllEdges) in their "plain" and exponentially penalized forms. The arrangement is similar to Figure \ref{schemes}.}
\label{pen_or_no}
\end{figure*}

\begin{figure*}
\begin{center}

\subfigure{
\includegraphics[width=1.75in]{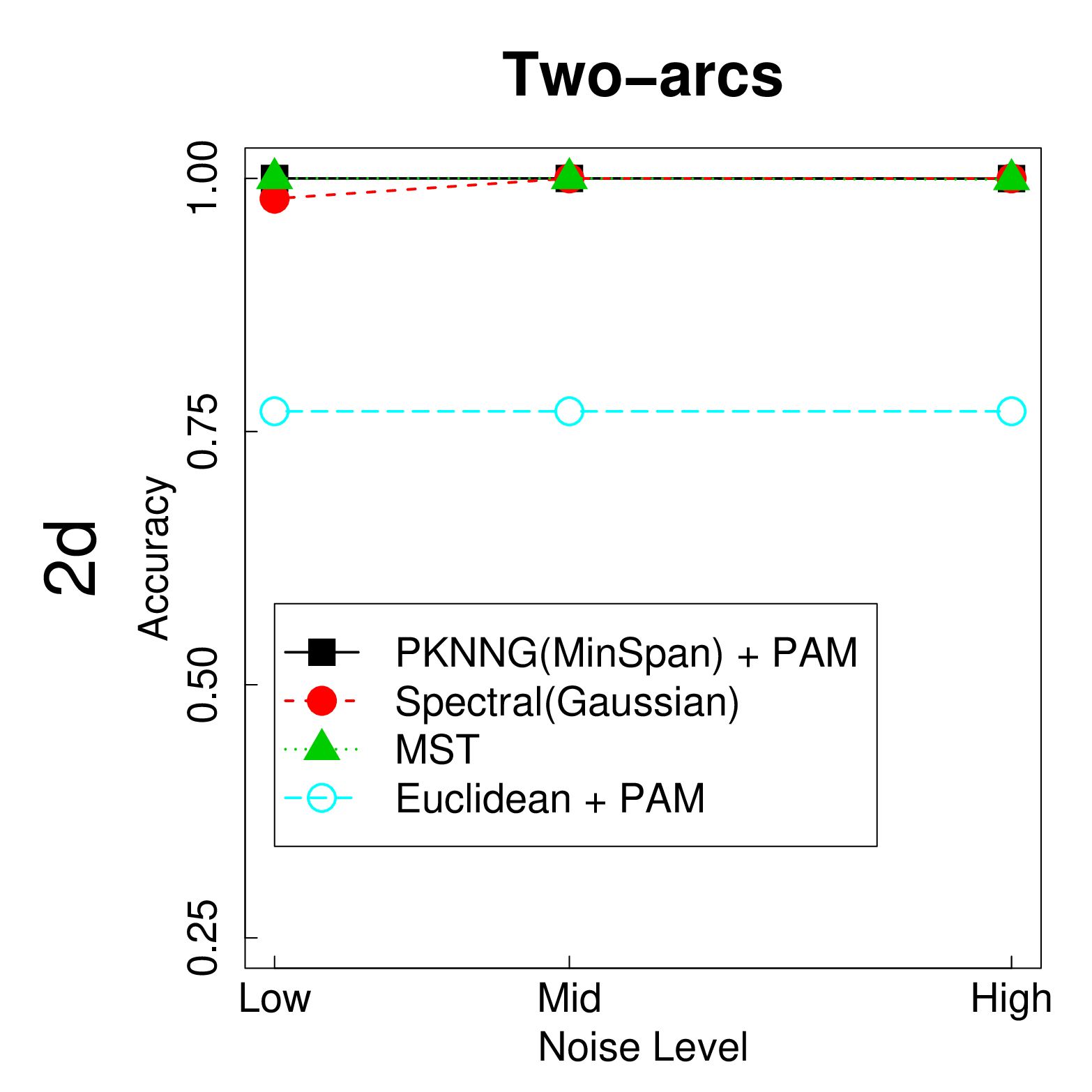}
}
\hspace{-0.3in}
\subfigure{
\includegraphics[width=1.75in]{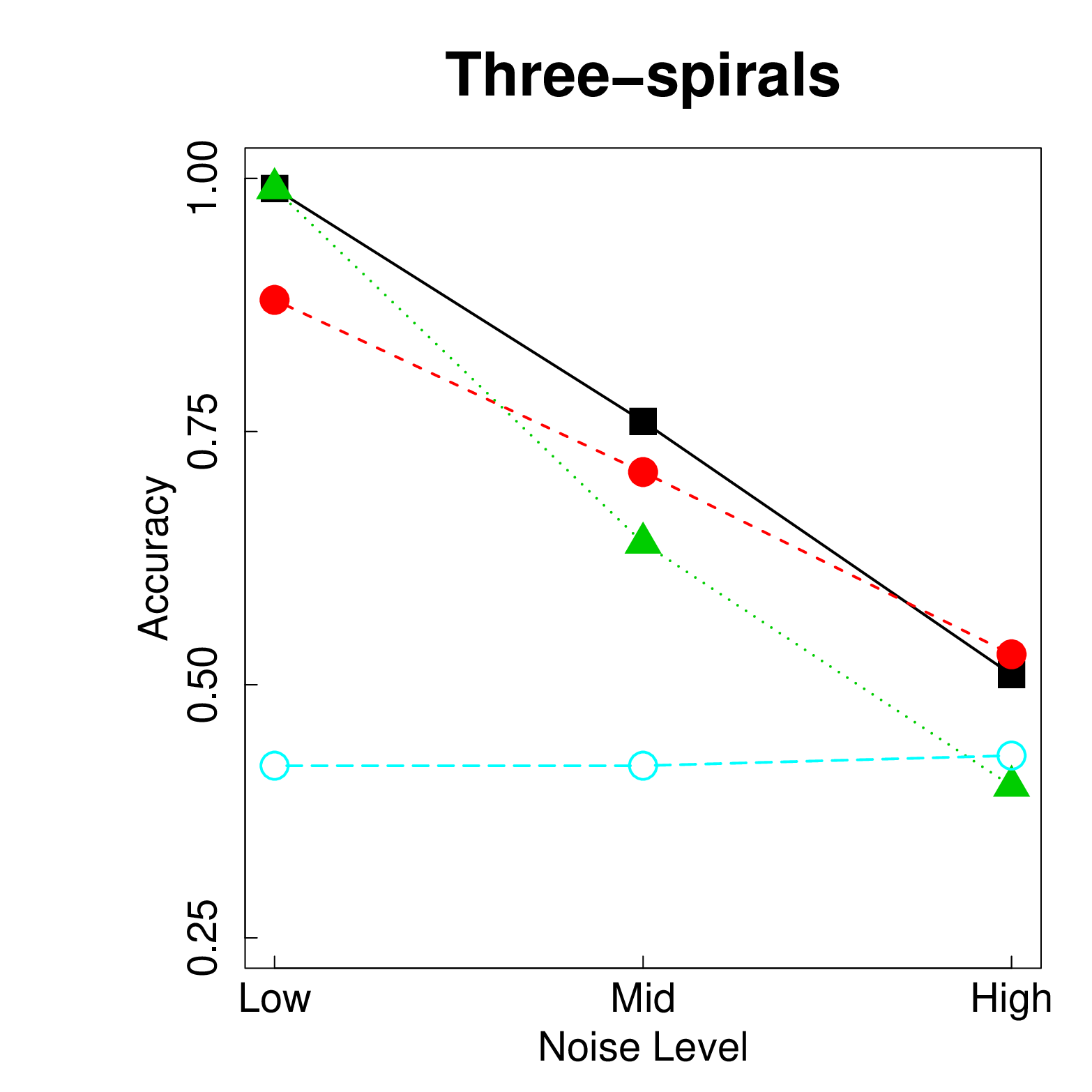}
}
\hspace{-0.3in}
\subfigure{
\includegraphics[width=1.75in]{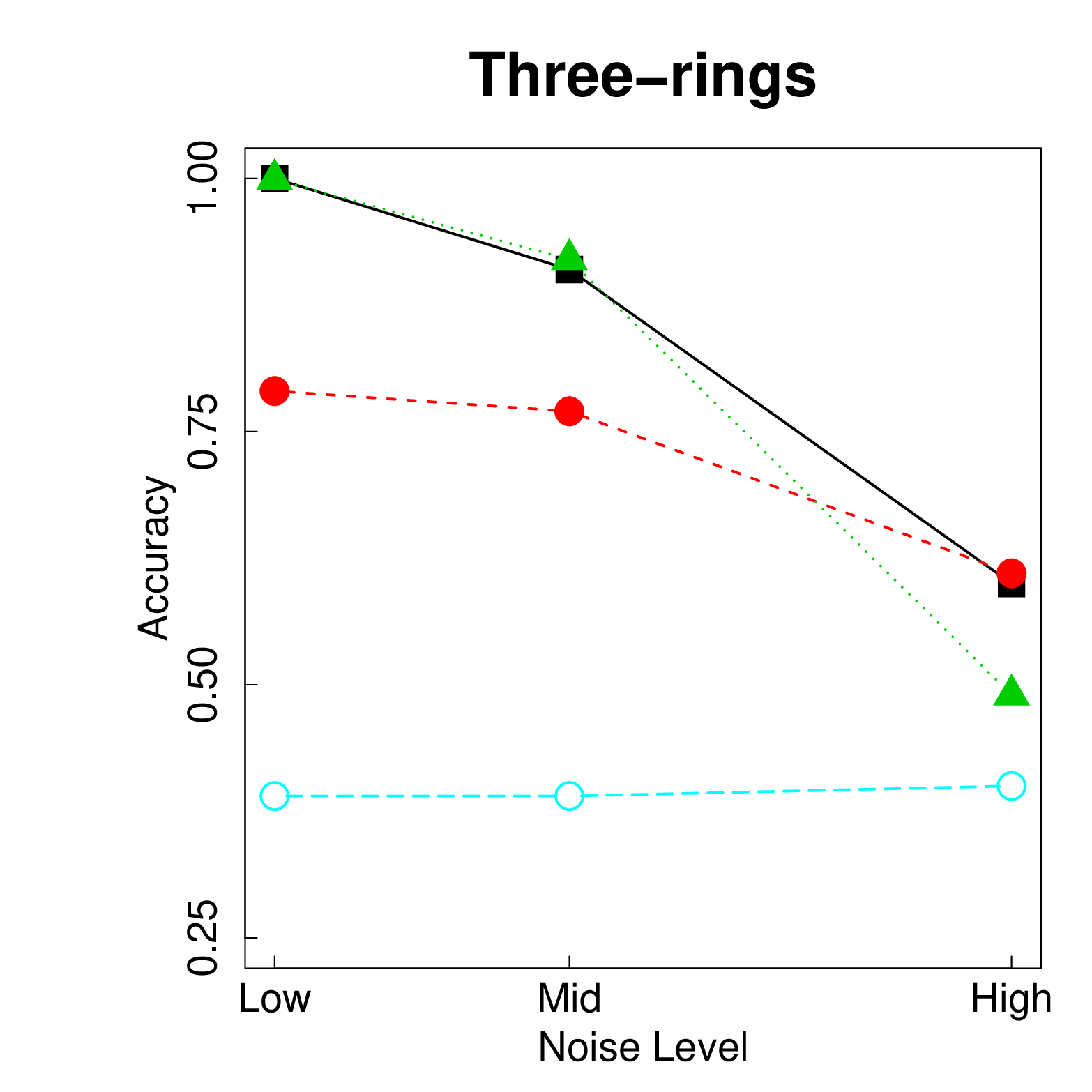}
} \\

\vspace{-.3in}

\subfigure{
\includegraphics[width=1.75in]{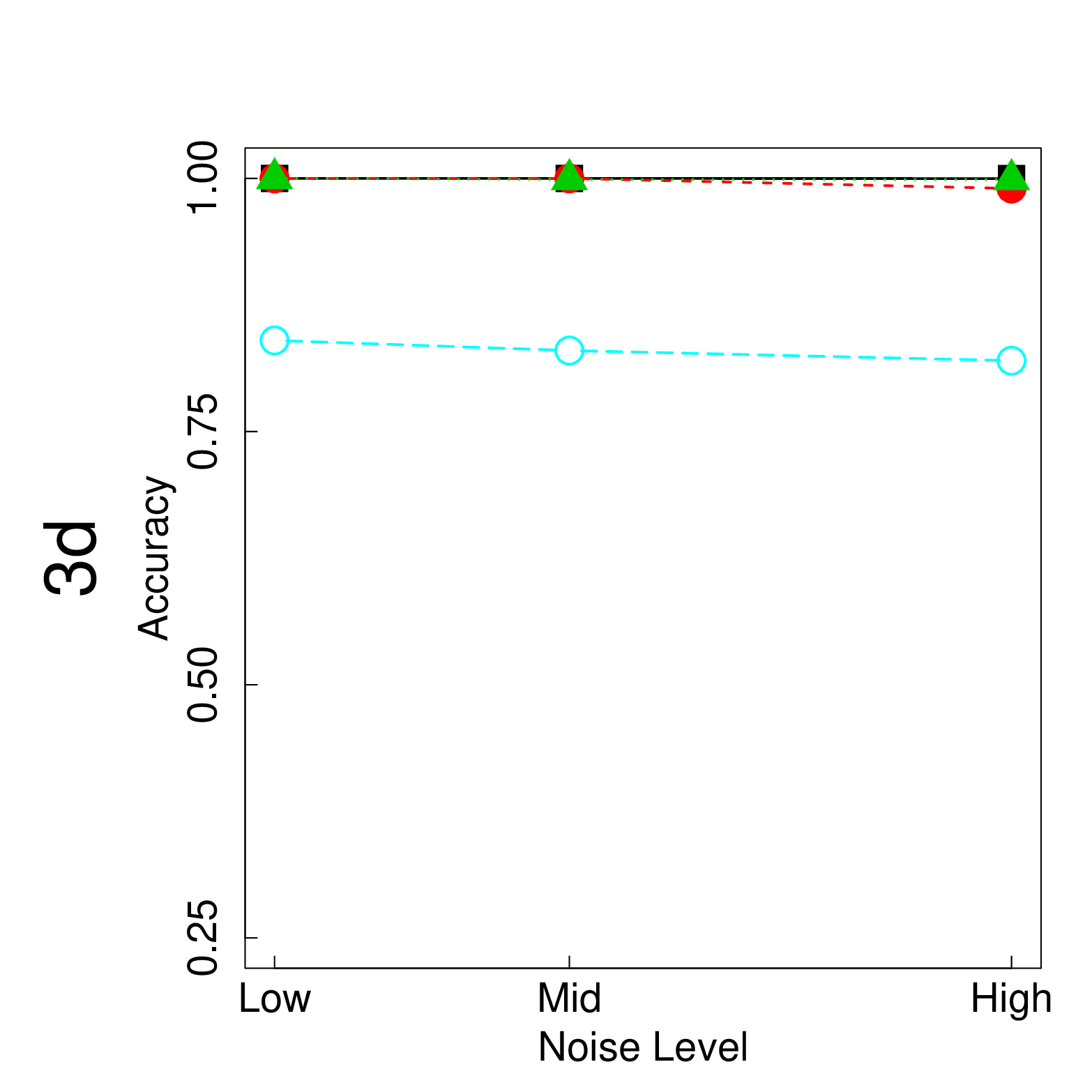}
}
\hspace{-0.3in}
\subfigure{
\includegraphics[width=1.75in]{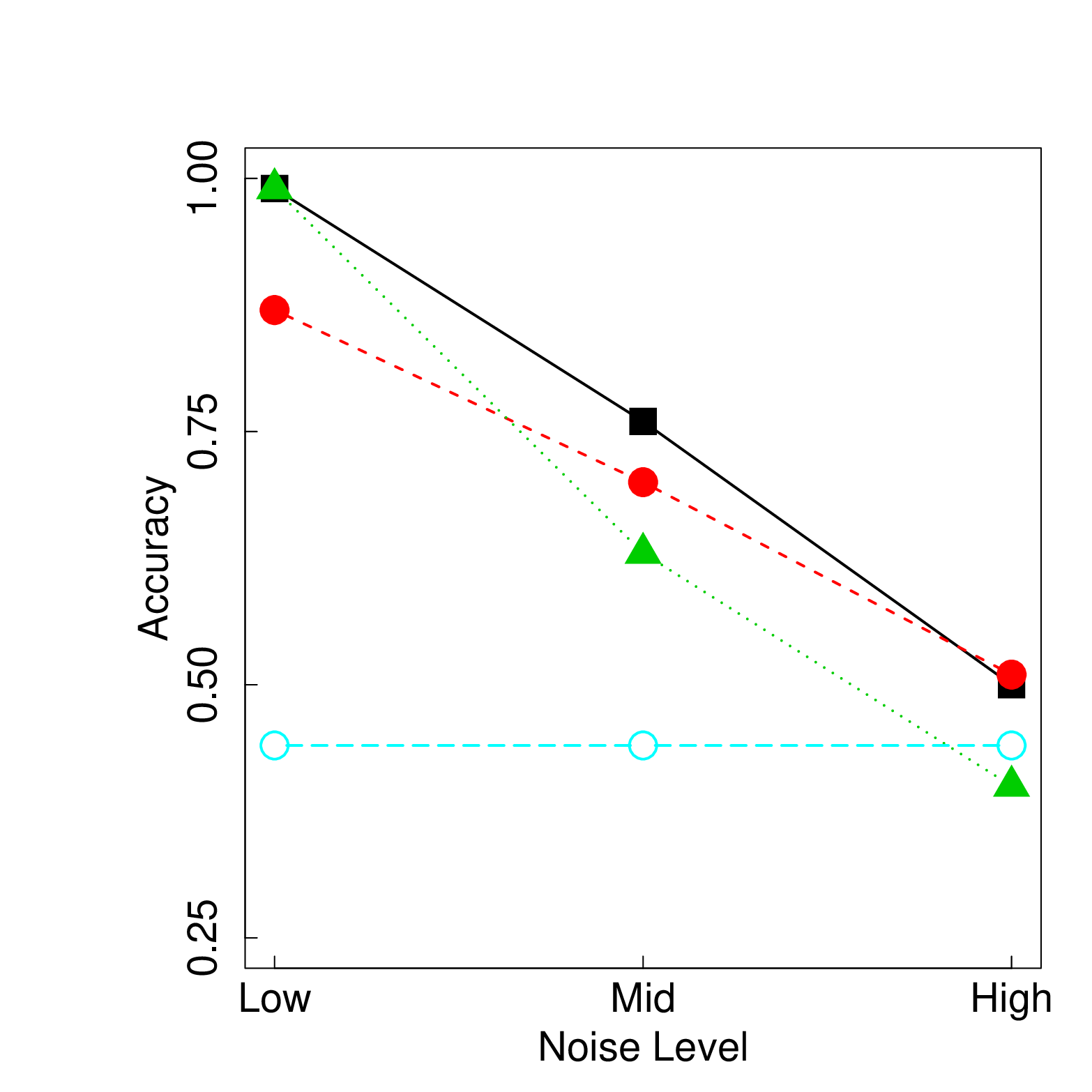}
}
\hspace{-0.3in}
\subfigure{
\includegraphics[width=1.75in]{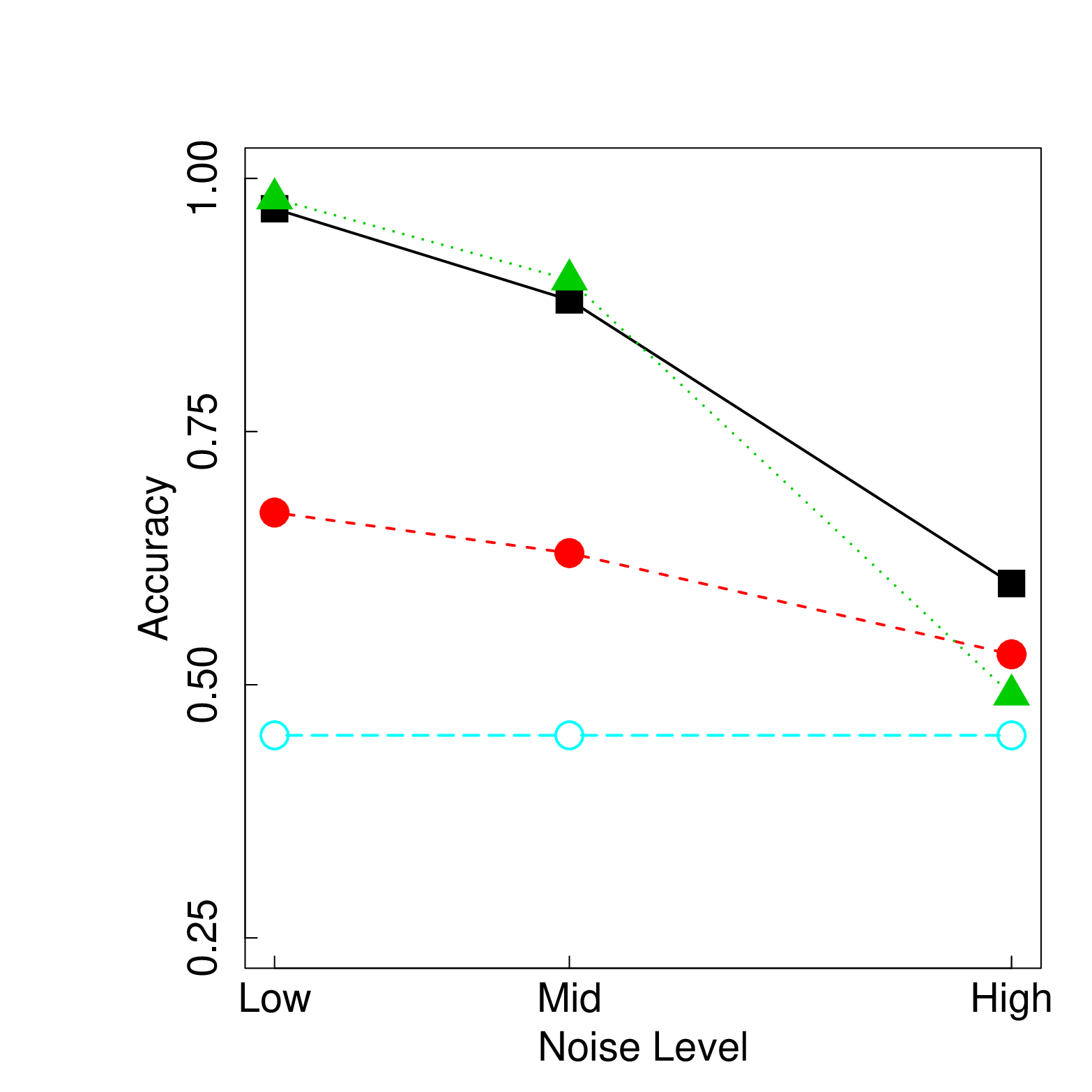}
} \\

\vspace{-.3in}

\subfigure{
\includegraphics[width=1.75in]{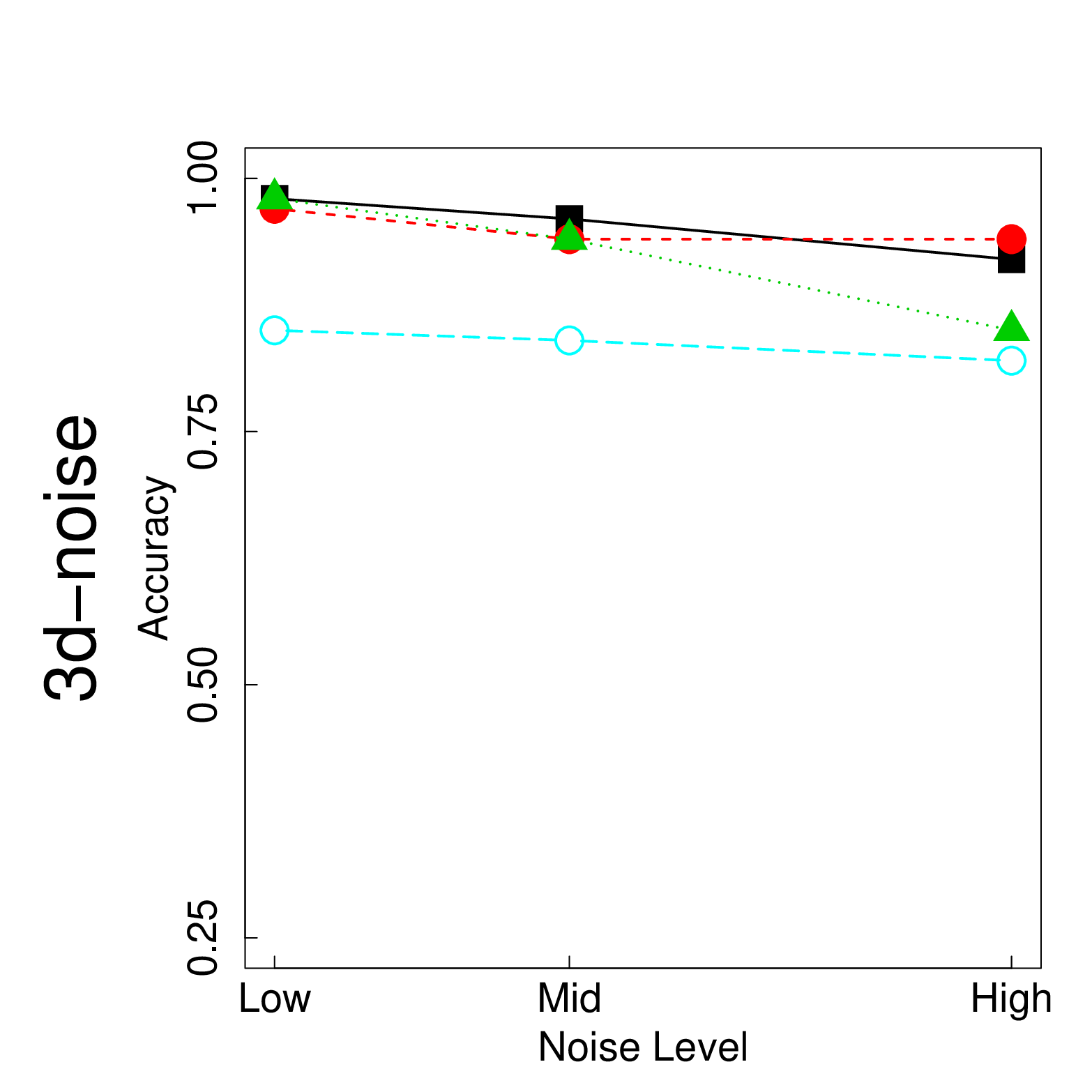}
}
\hspace{-0.3in}
\subfigure{
\includegraphics[width=1.75in]{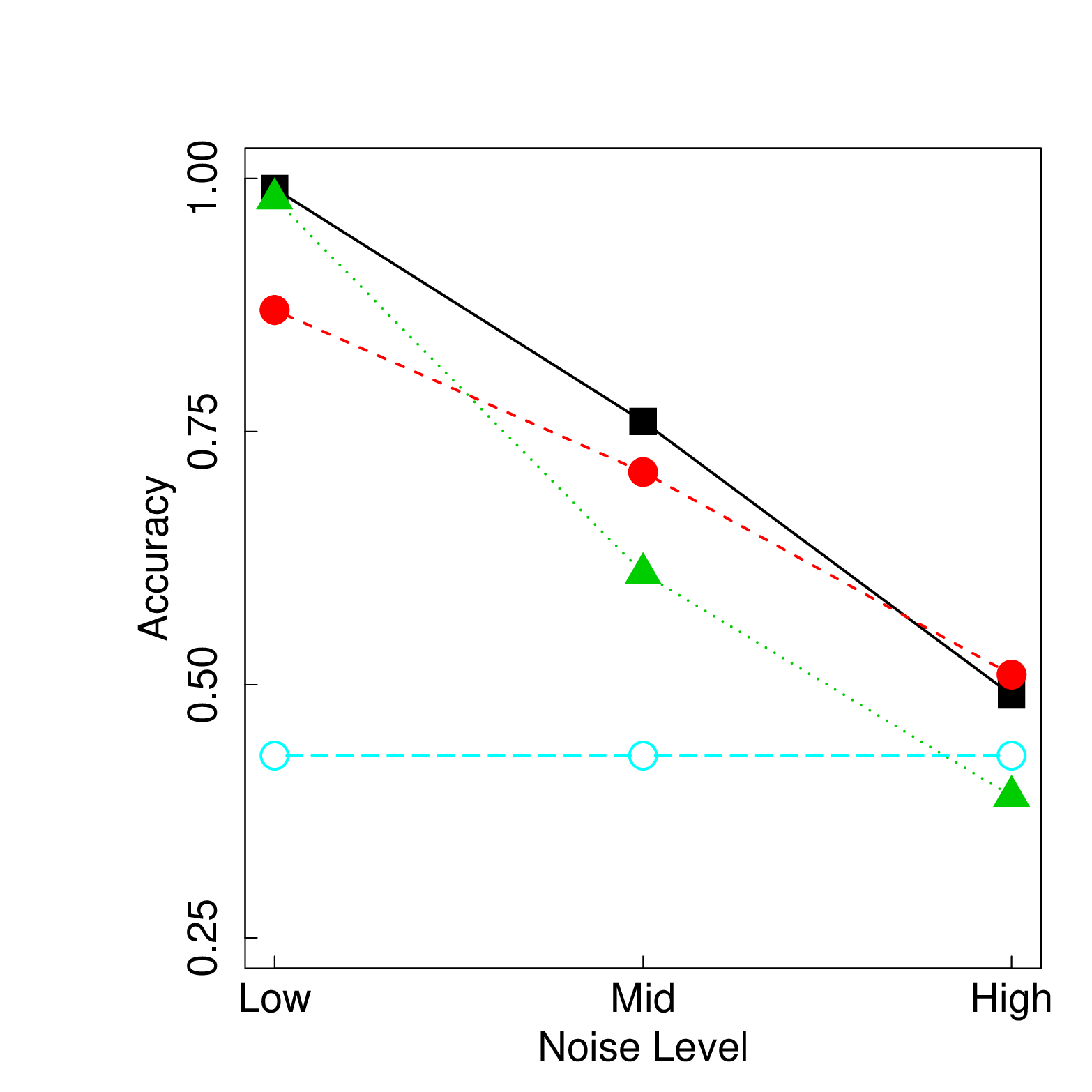}
}
\hspace{-0.3in}
\subfigure{
\includegraphics[width=1.75in]{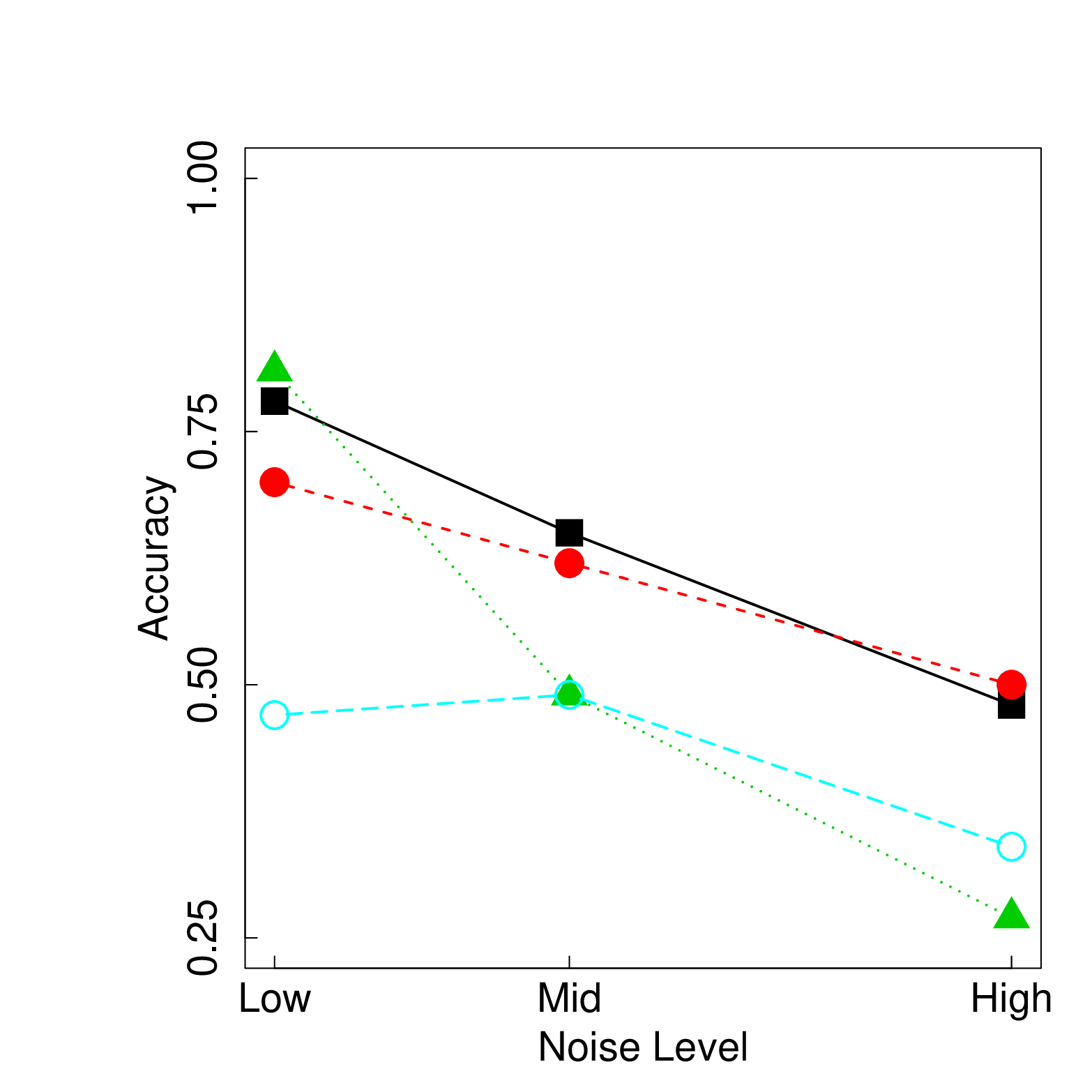}
} \\

\vspace{-.3in}

\subfigure{
\includegraphics[width=1.75in]{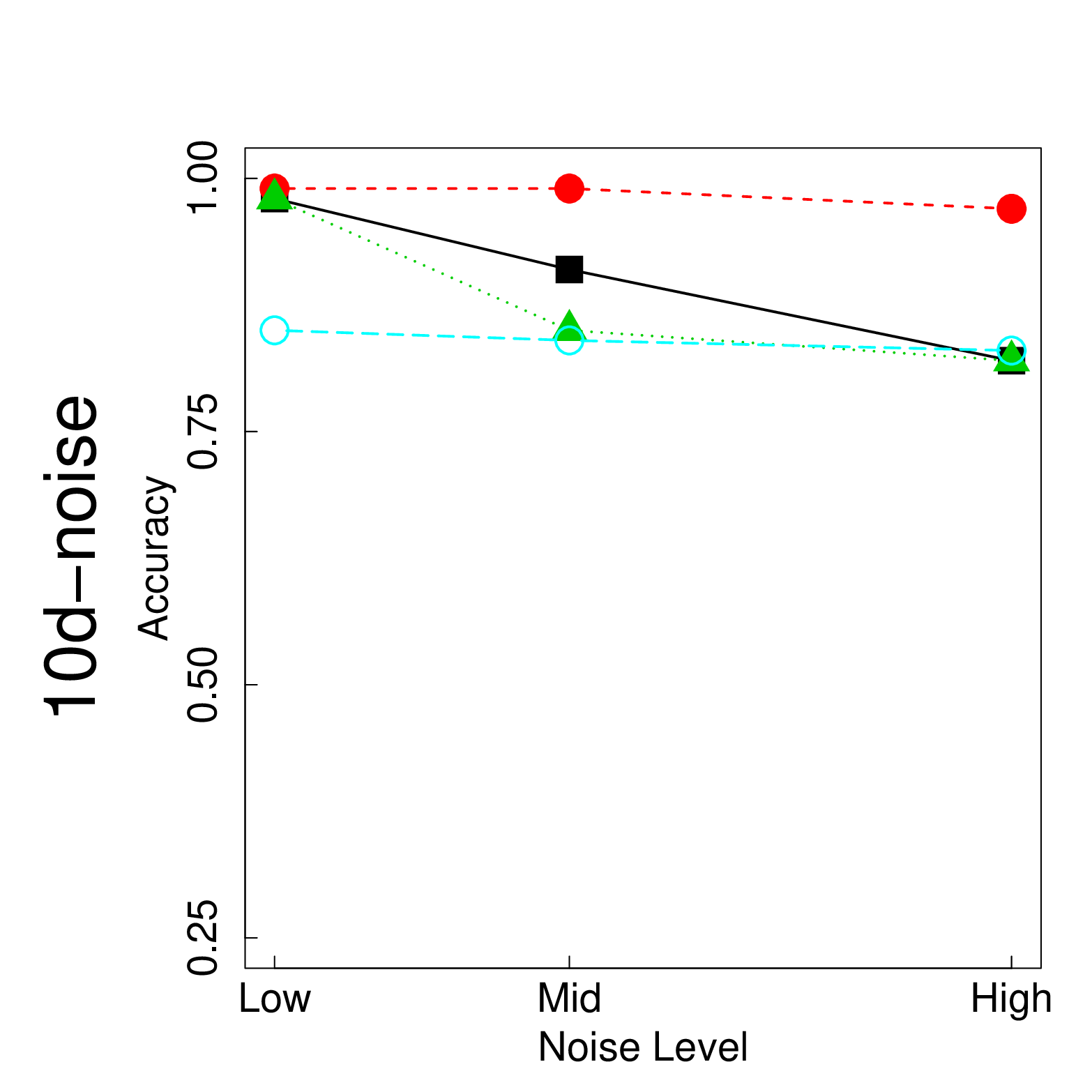}
}
\hspace{-0.3in}
\subfigure{
\includegraphics[width=1.75in]{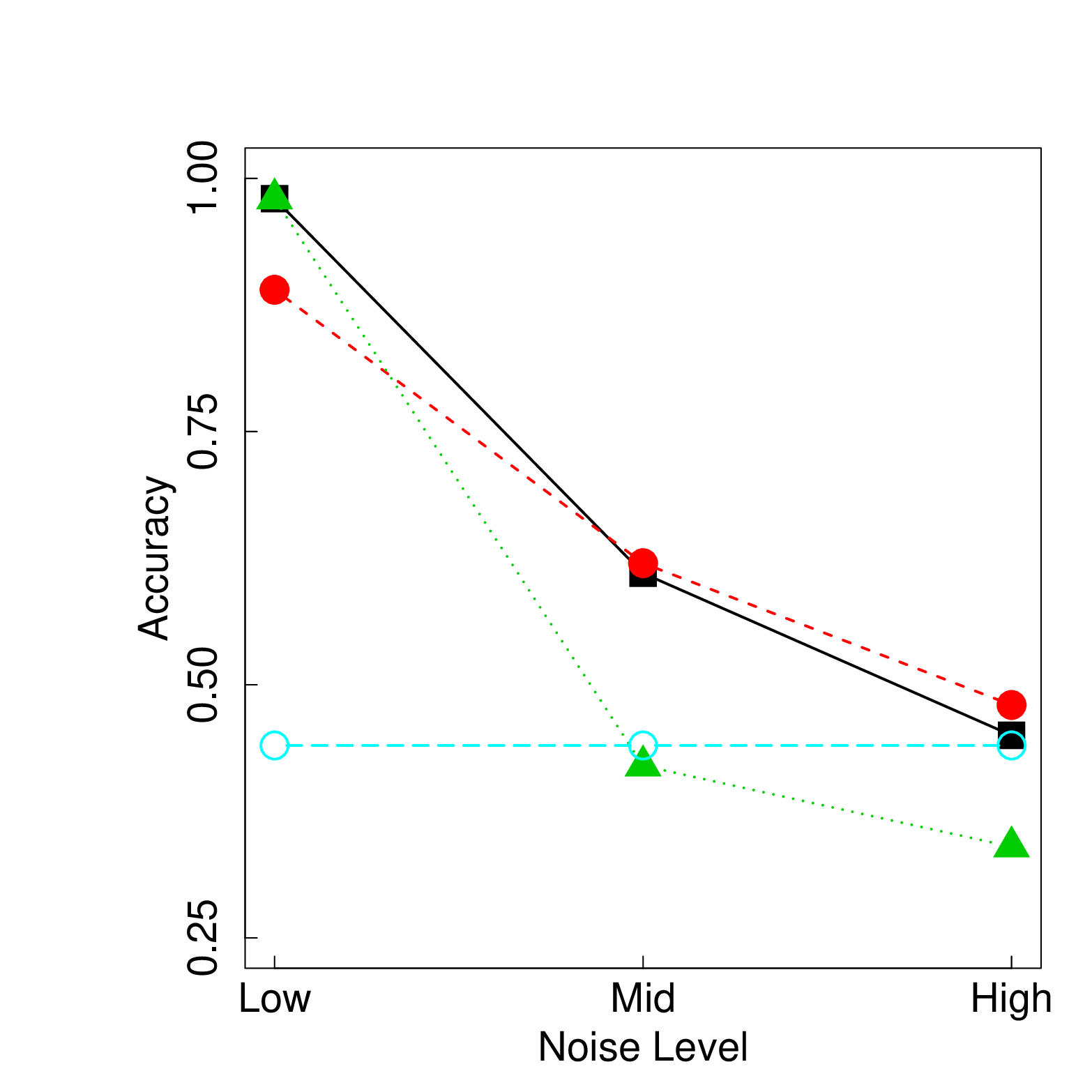}
}
\hspace{-0.3in}
\subfigure{
\includegraphics[width=1.75in]{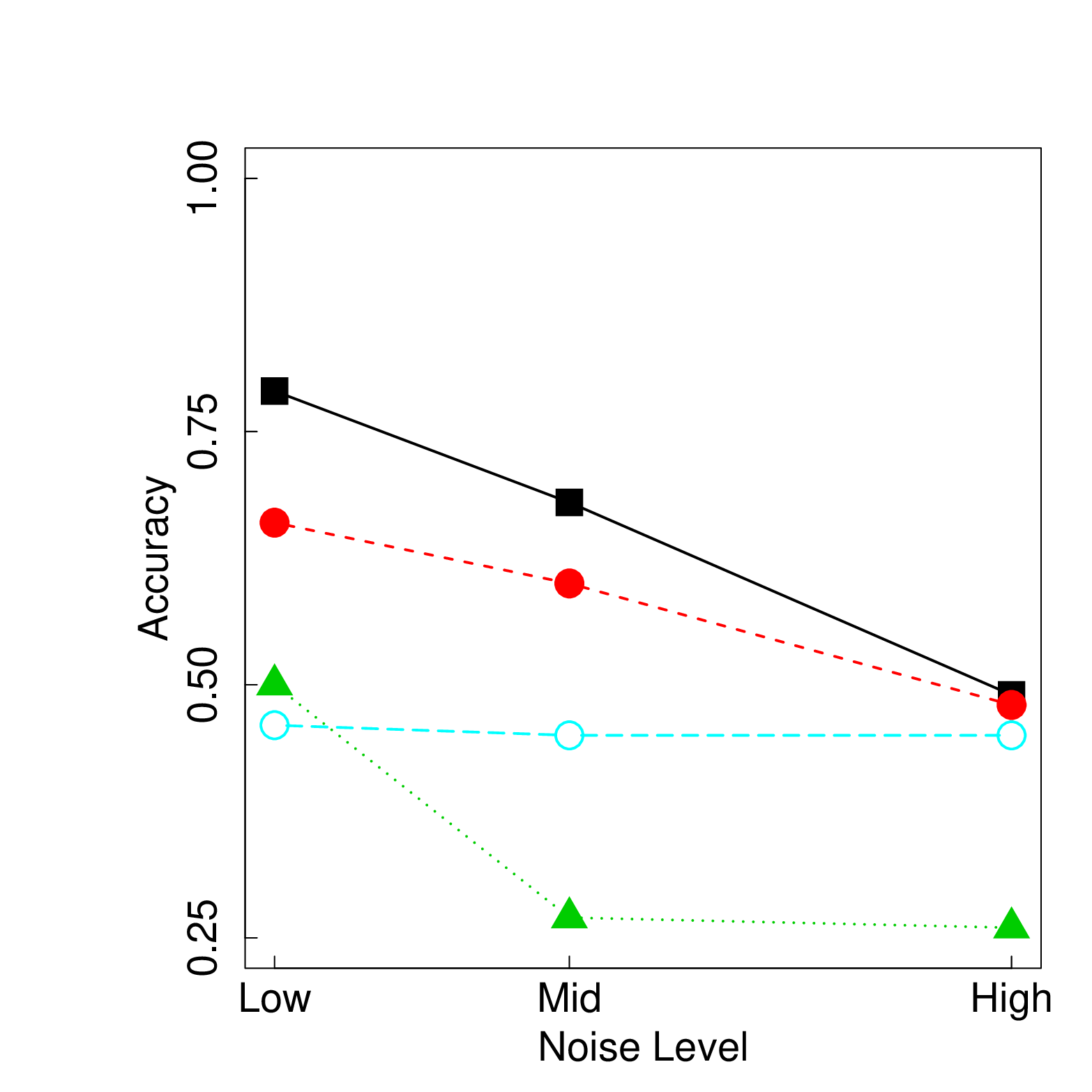}
} \\
\vspace{-.1in}
\end{center}
\caption{Comparison of the PKNNG-MinSpan metric + PAM clustering method with two other efficient methods (Spectral clustering and Minimum Spanning Tree (MST) clustering). The arrangement is similar to Figure \ref{schemes}.}
\label{comparacion}
\end{figure*}

In a first series of experiments we evaluated the four connection schemes previously discussed. In Figure \ref{schemes} we show the corresponding results using the PAM clustering algorithm\footnote{We repeated the full experiment using the HC and AP clustering algorithms, finding completely equivalent results (data not shown).}. The Figure includes the results for the three artificial datasets and the four embedding situations previously discussed. At each panel we also include the results of using the standard Euclidean metric as a baseline reference. 

The qualitative results are similar in all the situations under analysis. It is clear from the Figure that three connection schemes (MinSpan, AllSubGraphs and AllEdges) have very similar performances, clearly superior to the baseline methodology. The Medoids scheme, as expected, does not show a real improvement over the plain Euclidean metric. Comparing the three efficient schemes, MinSpan seems to be superior to AllSubGraphs and AllEdges in a few situations, in particular in the 3d--noise and 10d--noise embeddings.

In all panels the difference in accuracy between the baseline and the three efficient schemes increases when noise decreases. In these cases, when there is a bigger separation among the clusters, the new metric can easily find the curved clusters in all embedding situations. The edge in performance of the PKNNG metric decays smoothly when the clusters start to overlap, becoming equivalent (but never worse) to the plain Euclidean metric in high--noise situations. It is interesting to note that the PKNNG metric can deal efficiently with clusters of different densities (Three-rings datasets) or with densities changing within the same cluster (Three-spirals dataset).

\subsection{Clustering algorithms}

We compared in this second experiment the three clustering methods (PAM, HC and AP) that we described in the Introduction, in order to investigate whether there is any dependence of the performance of the PKNNG metric on the algorithm being used. In Figure \ref{clust_methods} we show the results of PAM, HC and AP applied to the similarities evaluated with the PKNNG-MinSpan metric\footnote{We repeated the experiment with the AllSubGraphs and AllEdges schemes, again finding similar results, not shown due to lack of space.}. We again include in each panel the results of the standard Euclidean metric plus PAM as a baseline reference. It is evident from the Figure that the good performance of the PKNNG metric is independent of the clustering algorithm in all situations.

\subsection{Penalization}

We performed a third experiment to assess the influence of the exponential penalization on the PKNNG metric. In this case we included in the comparison two connection schemes, namely MinSpan and AllEdges, under two different settings: i) in the (normal) exponentially penalized version (as used in all previous experiments), and ii) in a "plain" version in which we eliminated the penalization and used for the added connecting edges the standard Euclidean distance. Both settings used exactly the same graphs, the only difference being the exponential penalization of some weights in the first case. 
It is easy to see that the "plain" version of AllEdges is simply the standard Euclidean metric, our baseline in all experiments. For completeness, we also include in this case the results obtained measuring similarities with the knn-graph with the minimum k-value that produces a fully connected graph (we call this method min-k-connected-graph). In Figure \ref{pen_or_no} we show the results obtained with the PAM clustering algorithm applied to the five settings described in this paragraph\footnote{As in the previous experiments, we also applied the HC and AP algorithm obtaining similar results, not shown here.}. Analyzing the Figure, the two methods that use "plain", non-fully-connected Euclidean graphs (the MinSpan-Plain and the min-k-connected-graph) show in most cases a better performance than the standard Euclidean metric, but, more importantly, the two penalized metrics are in all cases clearly superior to the corresponding "plain" methods. These results suggest that the key factor of our new metric is the exponential penalization of the added edges.

\subsection{Other Clustering methods}

In this last experiment with the artificial datasets we selected a good representative of the new methods (the PKNNG-MinSpan metric plus the PAM clustering algorithm), and compared it with two well-known, state-of-the-art clustering methods. The first method under comparison is the Single Linkage Hierarchical Clustering algorithm\cite{sneath}, which is equivalent to constructing a Minimum Spanning Tree (MST) of the dataset. As we stated in the Introduction, this algorithm can identify curved/arbitrary clusters in some situations. The second method is the Spectral Clustering algorithm\cite{spectral}. This method has recently received increasing attention and is considered to be effective for arbitrary manifolds. We used a Gaussian kernel for which, at each run, we set the $\sigma$ parameter proportional to the mean distance between points in the datasets. Again, we included the PAM algorithm with the plain Euclidean metric as a baseline.

Figure \ref{comparacion} show the corresponding results. The three methods show equivalent performance in low noise situations. MST seems to be more affected by high noise levels and high dimensionality situations, giving worse results than the baseline method in some cases. Overall, PKNNG+PAM seems to be superior to Spectral Clustering on the Three--spirals and Three--rings datasets, and the opposite can be observed on the Two--arcs dataset. As usually happens with efficient methods, which one works better is highly problem--dependent. 

It is worth mentioning that we applied all three clustering algorithms as "off-the-shelf" methods, i.e., without any fine--tunning of internal parameters. This fact can explain the sometimes low performance of Spectral clustering, because in some cases the automatic setting of the $\sigma$ parameter can be inappropriate. Even taking this into account, we believe that this is the fairest setting for the comparison, because: i) the PKNNG metric also has a free parameter, $\mu$, which we choose to set automatically, and ii) in real world datasets, when there is no gold standard to compare against, there is no clear method for fine--tuning the free parameters.

\subsection{MNIST Digits datasets}

As a final, real-world example of the use of the PKNNG metric on a high-dimensional dataset, we applied it to clustering subsets of the MNIST handwritten digits dataset\footnote{Available at http://yann.lecun.com/exdb/mnist/} collected by LeCun and Cortez \cite{lecun98a}. The MNIST dataset is a subset of a larger set available from NIST. The 8-bit gray--scale images of the digits have been size-normalized and centered in 20x20 pixels images. We used only the 60000 images in the training set collected by LeCun, without any pre-processing at all. In a first experiment, we took 500 random samples of each digit and applied the four clustering procedures (as described in the previous subsection) to find 10 clusters. We then evaluated the clustering accuracy and repeated the experiment 10 times, taking care of using non--overlapping random samples. In the column labeled 1:10 of Table \ref{digits} we show the mean accuracy obtained with each method. In this problem the performance of the PKNNG+PAM method is $20\%$ better than Spectral Clustering, which is the second best. We repeated the experiment (500 non--overlapping random samples repeated 10 times) but reducing it to subsets of three digits that are usually highly confused (4-7-9 and 3-5-8). In the last two columns of Table \ref{digits} we show the corresponding results. Again, PKNNG+PAM has a clear edge over all other methods under comparison.

\begin{table}[tb]
\renewcommand{\arraystretch}{1.3}
\caption{Clustering results on the MNIST Digits dataset. Columns show the mean clustering accuracy ($\pm$ one standard deviation) over 10 random partitions of the different subsets.}
\label{digits}
\centering
\begin{tabular}{|c||c||c||c|}
\hline
\hline
Method & 1:10 & 4-7-9 & 3-5-8 \\
\hline
\hline
PKNNG + PAM & 0.60 $\pm$ 0.04 & 0.58 $\pm$ 0.05 & 0.69 $\pm$ 0.05 \\
Spectral & 0.48 $\pm$ 0.05 & 0.52 $\pm$ 0.04 & 0.62 $\pm$ 0.02 \\
Euclidean + PAM & 0.46 $\pm$ 0.02 & 0.48 $\pm$ 0.09 & 0.55 $\pm$ 0.08 \\
MST & 0.37 $\pm$ 0.04 & 0.43 $\pm$ 0.05 & 0.47 $\pm$ 0.05 \\
\hline
\hline
\end{tabular}
\end{table}

\section{Conclusions}

In this work we have introduced the Penalized k-Nearest-Neighbor-Graph based metric, and have shown that it is a useful tool for clustering arbitrary manifolds. The metric can be used in combination with any clustering algorithm, as we have demonstrated with several experiments using artificial datasets. Our new PKNNG metric is based on a two-step method: first we construct the k-Nearest-Neighbor-Graph of the dataset using a low k-value (from 3 to 7), and then we use exponentially penalized weights for connecting the sub-graphs produced by the first step. 

We have discussed several possible schemes for connecting the different sub-graphs, finding after long experiments with artificial datasets that any scheme that includes the MinSpan set shows equivalently good results. Overall, using strictly the MinSpan set seems to produce the best results according to our analysis.

The key innovation of our metric is the use of exponentially penalized weights to connect the sub-graphs. We have confirmed this fact by using three diverse artificial problems and four different (non-linear) embeddings. In that experiment most of the gain in accuracy over the Euclidean metric was lost when we eliminated the penalizing factor.

A comparison with two state-of-the-art algorithms for clustering manifold data, namely Spectral Clustering and Minimum-Spanning-Trees, confirms the validity of our results. As a real world application, we have clustered digit images from the MNIST dataset, finding that the PKNNG metric plus the PAM clustering algorithm was superior to other methods in this task.

Future work includes the application of the PKNNG metric to real genomic and proteomic data, face recognition and others.

\section*{Acknowledgements}

We acknowledge partial support for this project from ANPCyT (grant PICT 2003 11-15132). We thank Pablo F. Verdes for useful comments on this and on a previous manuscript, and Monica Larese and Lucas Uzal for a carefull reading of this manuscript.




\begin{thebibliography}{00}


\bibitem{tpami1} P. Franti, O. Virmajoki, and V. Hautamaki, Fast Agglomerative Clustering Using a k-Nearest Neighbor Graph, IEEE Trans. Pattern Analysis Machine Intelligence, vol. 28, no. 11, (2006), pp. 1875--1881.

\bibitem{tpami2} M. Yang and K. Wu, A Similarity-Based Robust Clustering Method, IEEE Trans. Pattern Analysis Machine Intelligence, vol. 26, no. 4, (2004), pp. 434--448.

\bibitem{tpami3} J. Yu, General C-Means Clustering Model, IEEE Trans. Pattern Analysis Machine Intelligence, vol. 27, no. 8, (2005), pp. 1197--1211.

\bibitem{benhur} A. Ben-Hur, A. Elisseeff, and I. Guyon, A Stability Based Method for Discovering Structure in Clustered Data, Proc. Pacific Symposium on Biocomputing, vol.7, (2002), pp. 6--17.

\bibitem{gap_stats} R. Tibshirani, G. Walther, and T. Hastie,  Estimating the Number of Clusters in a Dataset Via the Gap Statistic, J. of the Royal Statistical Soc. B, vol.63, no.2, (2001) , pp. 411--423.

\bibitem{review1} R. Xu, and D. Wunsch II, Survey of Clustering Algorithms, IEEE Trans. on Neural Networks, vol. 16, no. 3, (2005), pp. 645--678.

\bibitem{review2} A.K. Jain, M.N. Murti and P.J. Flynn, Data Clustering: A Review, ACM Computing Surveys, vol. 31, no. 3, (1999), pp. 264--323.

\bibitem{king} B. King, Step-wise clustering procedures, J. Am. Stat. Assoc., vol. 69, (1967), pp. 86--101.

\bibitem{kmeans} J. McQueen, Some methods for classification and analysis of multivariate observations, Proc. Fifth Berkeley Symp. on Math. Statistics and Probability, (1967), pp. 281--297.

\bibitem{pam} L. Kaufman, and P.J. Rousseeuw, Finding Groups in Data: An Introduction to Cluster Analysis, John Wiley \& Sons, 1990.

\bibitem{frey} B.J. Frey, and D. Dueck, Clustering by passing messages between data points, Science, vol. 315, (2007), pp. 972--976.

\bibitem{sneath}  P.H.A. Sneath, and R.R. Sokal, Numerical Taxonomy, Freeman, 1973.

\bibitem{chamaleon} G. Karypis, E..H. Han, and V. Kumar, Chameleon: Hierarchical clustering using dynamic modeling, Computer, vol. 32, no. 8, (1999), pp. 68--75.

\bibitem{kernelmethods} J. Shawe-Taylor, and N. Cristianini, Kernel Methods for Pattern Analysis, Cambridge University Press, 2004.

\bibitem{isomap} J. Tenenbaum, V. de Silva, and J. Langford, A global geometric framework for nonlinear dimensionality reduction, Science, vol. 290, (2000), pp. 2319--2323.

\bibitem{lle} S. Roweis, and L. Saul, Nonlinear dimensionality reduction by locally linear embedding, Science,  vol. 290, (2000), pp. 2323--2326.

\bibitem{laplacian} M. Belkin, and P. Niyogi, Laplacian eigenmaps and spectral techniques for embedding and clustering, Advances in Neural Information Processing Systems, vol. 14, (2002), pp. 585--591.

\bibitem{ham04} J. Ham, D.D. Lee, S. Mika, and B. Schölkopf, A Kernel View of the Dimensionality Reduction of Manifolds, Proc. twenty-first int. conf. on Machine learning, (2004), pp. 47--52.

\bibitem{asai07} A. Baya, and P.M. Granitto, ISOMAP based metrics for Clustering, Proc. 36th Int. Conf. of the Argentine Computer Science and Operational Research Society, 2007.

\bibitem{dijkstra} T.H. Cormen, C.E. Leiserson, R.L. Rivest, and C. Stein, Introduction to Algorithms, Second Edition, MIT Press and McGraw-Hill, 2001.

\bibitem{spectral} A. Ng, M. Jordan, and Y. Weiss, On spectral clustering: Analysis and an algorithm Advances in Neural Information Processing Systems, vol. 14, (2002),pp. 849--856.

\bibitem{lecun98a} Y. LeCun, L. Bottou, Y. Bengio and P. Haffner, Gradient-Based Learning Applied to Document Recognition, Proceedings of the IEEE, vol. 86, no. 11, (1998), pp.2278--2324.

\end{thebibliography}
\end{document}